\documentclass[preprint,12pt]{elsarticle}
\usepackage[margin=2.5cm]{geometry} 

\usepackage{verbatim} 
\usepackage[utf8]{inputenc} 

\usepackage{booktabs}
\usepackage{framed,multirow}
\usepackage{hyperref}
\hypersetup{
    colorlinks = true,
    linkcolor = {magenta},
    citecolor = {blue}
}
\usepackage{url}   

\usepackage{booktabs}
\usepackage{amsfonts}       
\usepackage{latexsym}
\usepackage{amsmath}
\usepackage{amssymb}
\usepackage{amsfonts}

\usepackage{cleveref}
\usepackage{nicefrac}       
\usepackage{microtype}      
\usepackage{lipsum}
\usepackage{adjustbox}
\usepackage{graphicx}
\usepackage{bm}

\usepackage{amsthm}
\newtheorem{theorem}{Theorem}
\newtheorem{lemma}{Lemma}

\usepackage[table,xcdraw]{xcolor}
\definecolor{newcolor}{rgb}{.8,.349,.1}
\usepackage[linesnumbered,ruled,vlined]{algorithm2e}
\usepackage{multirow}
\usepackage{diagbox}
\usepackage{subfig}
\usepackage{tikz}

\usetikzlibrary{external}

\usepackage{pgfplots}
\pgfplotsset{compat=1.7}
\usetikzlibrary{positioning}
\usepackage{etoolbox} 
\colorlet{updated_color}{green!80!red!90!}

\SetCommentSty{mycommfont}

\crefformat{section}{\S#2#1#3}
\crefformat{subsection}{\S#2#1#3}
\crefformat{subsubsection}{\S#2#1#3}
\crefrangeformat{section}{\S\S#3#1#4 to~#5#2#6}
\crefmultiformat{section}{\S\S#2#1#3}{ and~#2#1#3}{, #2#1#3}{ and~#2#1#3}

\journal{}

\begin{document}

\begin{frontmatter}

\title{Conditionally adaptive augmented Lagrangian method for physics-informed learning of forward and inverse problems}


\author[inst1]{\href{https://orcid.org/0009-0002-9465-7208}{\includegraphics[scale=0.08]
{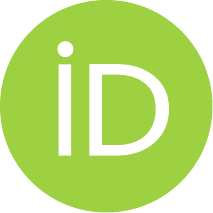}\hspace{1mm}Qifeng Hu}}
\ead{qih56@pitt.edu}

\author[inst1]{\href{https://orcid.org/0000-0002-1095-0881}{\includegraphics[scale=0.08]{orcid.pdf}\hspace{1mm}Shamsulhaq Basir}}
\ead{shb105@pitt.edu}

\author[inst1]{\href{https://orcid.org/0000-0003-1967-7583}{\includegraphics[scale=0.08]{orcid.pdf}\hspace{1mm}Inanc Senocak \corref{cor1}}}
\cortext[cor1]{corresponding author:~senocak@pitt.edu (Inanc Senocak)}

\address[inst1]{Department of Mechanical Engineering and Materials Science, University of Pittsburgh, Pittsburgh, PA 15261, USA}

\begin{abstract}
We present several key advances to the Physics and Equality Constrained Artificial Neural Networks (PECANN) framework \cite{PECANN_2022}, substantially improving its capacity to solve challenging partial differential equations (PDEs). Our enhancements broaden the framework's applicability and improve efficiency. First, we generalize the Augmented Lagrangian Method (ALM) to support multiple, independent penalty parameters for enforcing heterogeneous constraints. Second, we introduce a constraint aggregation technique to address inefficiencies associated with point-wise enforcement. Third, we incorporate a single Fourier feature mapping to capture highly oscillatory solutions with multi-scale features, where alternative methods often require multiple mappings or costlier architectures. Fourth, a novel time-windowing strategy enables seamless long-time evolution without relying on discrete time models. Fifth, and critically, we propose a conditionally adaptive penalty update (CAPU) strategy for ALM that accelerates the growth of Lagrange multipliers for constraints with larger violations, while enabling coordinated updates of multiple penalty parameters. We demonstrate the effectiveness of PECANN–CAPU across diverse problems, including the transonic rarefaction problem, reversible scalar advection by a vortex, high-wavenumber Helmholtz and Poisson's equations, and inverse heat source identification. The framework achieves competitive accuracy across all cases when compared with established methods and recent approaches based on Kolmogorov–Arnold networks. Collectively, these advances improve the robustness, computational efficiency, and applicability of PECANN to demanding problems in scientific computing.
\end{abstract}


\end{frontmatter}

\section{Introduction}\label{sec:intro}

The application of artificial neural networks (ANNs) to the solution of partial differential equations (PDEs) dates back to the early 1990s \cite{dissanayake1994neural,van1995neural}, following the theoretical establishment of ANNs as universal function approximators \cite{hornik1989multilayer}. With the advent of deep learning--driven by key developments such as back propagating errors in neural networks \cite{backpropagation1986}, deep belief networks \cite{Hinton2006}, rectified linear units \cite{relu_2010}, efficient training of neural networks \cite{batch_normalization2015}, dropout regularization \cite{dropout_2014}, and stochastic optimization \cite{kingma2015adam}--there has been a renewed interest in leveraging neural networks for solving PDEs \cite{E2018,Han2018,sirignano2018dgm, Zhu2019}. This resurgence was further catalyzed by the introduction of the physics-informed neural networks (PINNs) framework \cite{raissi2019pinn}.

In the baseline PINN approach \cite{dissanayake1994neural,van1995neural,raissi2019pinn}, governing PDEs are incorporated into training through a composite loss function derived from an unconstrained optimization perspective. This composite loss term combines contributions from residual forms of the PDE, boundary and initial conditions, and sample data, each weighted by tunable hyperparameters. Thus, in the baseline approach, solving a PDE is cast as an unconstrained, multi-objective optimization problem. While PINN performance depends on activation functions \cite{jagtap2020adaptive}, sampling strategies \cite{wu2023comprehensive}, and network architectures \cite{wang2021understanding,PECANN_2022}, the formulation and balance of loss terms are central to achieving accurate and stable predictions.

Because training a neural network relies on gradient-based optimizers, large gradients can dominate, leading to unstable behavior where some loss terms are prioritized over others. Balancing these the loss terms in a composite objective function requires manual or dynamic tuning of hyperparameters, which becomes impractical in the absence of validation data or exact solutions \cite{basir2022critical}. This challenge has driven extensive research into dynamic loss reweighting strategies \cite{wang2021understanding,van2022optimally,liu2021dual,mcclenny2020self,residual_based_attention_2024}, which are critical for improving PINN's accuracy and stability. A comprehensive review of these methods and emerging PINN variants is provided by \citet{Toscano2025_review}.

Principally speaking, in optimization problems, it is crucial to distinguish which terms should serve as objectives and which should be enforced as constraints \cite{martins2021engineering}. Departing from the multi-objective optimization paradigm in the baseline PINN approach, several works have adopted the constrained optimization perspective. In the self-adaptive PINN method \cite{mcclenny2020self}, the objective function was formulated as a PDE-constrained optimization problem using a penalty-based method with trainable penalty coefficients. Other works have adopted the exact enforcement of boundary conditions as hard constraints through auxiliary functions \cite{lagaris1998artificial, Berg2018, lu2021physics}. \citet{PECANN_2022} proposed the Physics and Equality Constrained Artificial Neural Networks (PECANN) framework to learn the solution of forward an inverse PDE problems.

In the PECANN framework, the augmented Lagrangian method (ALM) \cite{hestenes1969multiplier, powell1969method} is employed to recast the constrained optimization problem into an unconstrained one in a principled manner. ALM provides a formal framework to enforce the constraints and should not be confused with optimizers such as stochastic gradient descent used to minimize an objective function in unconstrained optimization. Notably, ALM integrates the strengths of both the penalty method and the method of Lagrange multipliers from the constrained optimization field. ALM aims for a balance between feasibility and optimality by iteratively updating a penalty parameter to regulate the influence of constraint violations \cite{Birgin2014,nocedal2006numerical, martins2021engineering}. 

With a few modifications, the PECANN framework has been successfully applied to a wide range of forward and inverse PDE problems from different domains \cite{Adaptive_ALM_2023b,Lan2024_pecann,hu_non-overlapping_2025,Su2025_pecann, Olson2026}. \citet{hu_non-overlapping_2025} demonstrated PECANN's versatile formulation by inferring interface transmission conditions in the parallel solution of two-dimensional Laplace and Helmholtz equations using a non-overlapping Schwarz-type domain decomposition method. In their approach, both the global solution and the interface transmission conditions--characterized by unknown parameters--were learned simultaneously, while enforcing vanishing PDE residuals and associated boundary conditions as constraints. \citet{Olson2026} developed one- and two-level domain decomposition methods to accelerate the PECANN method through parallel computing. Their work highlighted the critical role of differentiating boundary conditions from interface continuity requirements when formulating the loss functions. \citet{Lan2024_pecann} applied the PECANN framework to develop surrogate models for groundwater flow dynamics by learning unsaturated infiltration solutions from limited initial and boundary data. They showed that PECANN outperformed both PINNs and purely data-driven approaches, and achieved accuracy comparable to finite-difference methods while requiring significantly less data. Most recently, \citet{Su2025_pecann} applied the PECANN framework to learn unstable solutions of the so-called ``bad'' Jaulent–Miodek nonlinear equation, which models the motion of non-viscous shallow water wave packets in a flat-bottomed domain under shear forces. To accelerate the training process, \citeauthor{Su2025_pecann} first pre-trained the network using the PINN methodology and then fine-tuned it with PECANN to achieve higher accuracy. They also highlighted the limitations of both the original PINN and PECANN approaches in solving the Jaulent–Miodek equation over larger computational domains, underscoring the need for the advancements to the PECANN framework introduced in the present work.

Fully-connected feed-forward neural networks, commonly referred to as multi-layer perceptrons (MLPs), are widely employed in machine learning applications. However, recent developments have introduced alternative architectures that aim to enhance expressivity and performance. Among these, Kolmogorov-Arnold Networks (KANs) \cite{liu2025kan} and Fourier feature networks \cite{tancik_fourier_2020} represent two notable examples. KANs are grounded in the Kolmogorov-Arnold representation theorem, wherein the activation functions of neurons are learned as one-dimensional splines. Despite their theoretical advantages, KANs exhibit significantly slower training times compared to conventional MLPs. In contrast, Fourier feature networks augment the representational capacity of MLPs for high-frequency functions by applying a Fourier feature mapping to the input space.

\citet{wang_kolmogorovarnold-informed_2025} introduced the PIKAN framework, which integrates KANs with PINNs. Their study substantiates the superior expressive capacity of KANs, while also underscoring their computational inefficiency and limitations in solving PDEs characterized by complex geometries or extensive computational grids. Furthermore, \citet{wang_eigenvector_2021} demonstrated that employing multiple Fourier feature mappings--each initialized with distinct variances--in conjunction with a fully-connected MLP architecture significantly enhances the model's ability to approximate solutions to benchmark PDEs exhibiting high-frequency behavior--scenarios in which conventional PINN models tend to underperform.

In the following, we present several key enhancements to the original PECANN framework \cite{PECANN_2022}. These advancements collectively enable the new framework to learn solutions to challenging problems frequently employed in the development and benchmarking of numerical methods for PDEs. 

\section{Technical Formulation}\label{sec:tech_back}
Let us consider a generic equality-constrained optimization problem:
\begin{equation}
\begin{aligned}
    \min_{\theta} \mathcal{J}(\theta), ~\quad \text{subject to } ~\quad \mathcal{C}_i(\theta) =0, \quad \forall i \in \mathcal{E}, \label{eq:constrained_problem}
\end{aligned}
\end{equation}
where $\mathcal{J}(\theta)$ is the objective function, $\mathcal{C}_i(\theta)$ are constraint functions, and all functions are continuous and real-valued over the primal variable $\theta \in \mathbb{R}^n$. The set $\mathcal{E}$ indexes a finite number of equality constraints. We can express all constraints collectively as a vector $\bm{\mathcal{C}}(\theta)$ and reformulate the problem using the augmented Lagrangian method \cite{hestenes1969multiplier, powell1969method}. This leads to the following unconstrained minimax formulation:
\begin{equation}
    \min_{\theta} \max_{\bm{\lambda}} \mathcal{L}(\theta; \bm{\lambda},\mu) =  \mathcal{J}(\theta) + \bm{\lambda}^T \bm{\mathcal{C}}(\theta)  + \frac{1}{2} \mu \| \bm{\mathcal{C}}(\theta) \|^2_2 , \label{eq:old_unconstrained_problem}
\end{equation}
where $\mathcal{L}$ is the augmented unconstrained loss, $\bm{\lambda}$ denotes the vector of Lagrange multipliers (also known as dual variables) associated with the constraint vector $\bm{\mathcal{C}}$, $\mu$ is a scalar penalty parameter.
The relationship between the primal and dual formulations is captured by the minimax inequality (a form of weak duality), which ensures that:
\begin{align}
   \max_{\bm{\lambda}} \min_\theta \mathcal{L}(\theta;\bm{\lambda})  \le \min_\theta \max_{\bm{\lambda}} \mathcal{L}(\theta;\bm{\lambda}).
\end{align}
The maximization over $\bm{\lambda}$ at optimization step $e$ can be performed iteratively using the following update rule, as proposed by \citet{hestenes1969multiplier}:
\begin{equation}
\begin{aligned}
\bm{\lambda}^{e} &\leftarrow \bm{\lambda}^{e-1} + \mu^{e-1} ~ \bm{\mathcal{C}}(\theta^{e}),
\label{eq:dual_ascent_aug}
\end{aligned}
\end{equation}
where $\mu^{e-1}$ is the penalty parameter from the previous step and $\bm{\mathcal{C}}(\theta^{e})$ is the constraint violation at the current iterate $\theta^{e}$.
We should note that $\mu$ can be viewed as a global learning rate for the Lagrange multipliers. ALM combines the merits of the penalty method and the Lagrange multiplier method by updating the penalty parameter in such a way that it balances the trade-off between feasibility and optimality, and ensures convergence. We provide a derivation of ALM in \ref{sec:appendixc}, emphasizing the role of Lagrange multipliers and penalty parameters in its formulation.

Algorithm \ref{alg:mpu_classic_training_algorithm} is the procedure that is commonly found in most textbooks presenting ALM. It is also the algorithm adopted in the original PECANN \cite{PECANN_2022} method. In this procedure, the penalty parameter $\mu$ is monotonically increased at each training iteration by an exponential factor $\beta$, until it reaches a predefined maximum threshold $\mu_{\max}$ for safeguarding. 
Bounding the penalty parameter with an upper limit is a common strategy utilized to prevent excessively large values that could otherwise induce numerical instability. We will refer to Algorithm \ref{alg:mpu_classic_training_algorithm} as ALM with monotonic exponential penalty update (MPU).

\IncMargin{1em}
\begin{algorithm}[!h]
\SetAlgoLined
\SetKw{KwInput}{Input:}
\SetKw{KwOutput}{Output:}
\SetKw{KwDefaults}{Defaults:}
\KwInput{$\theta^0$, $\beta$, $\mu_{\max}$, $E$}\\
$\bm{\lambda}^0 = \textbf{1}$ \hspace{14em}
\tcc{Initializing Lagrange multipliers}
$\mu^0 = 1 $ \hspace{14em}
\tcc{Initializing the penalty parameter}
\BlankLine
\For{$e = 1  ~ \KwTo ~E...$}{
    $\theta^e \gets \text{Update}(\theta^{e-1})$ via Eq.~\ref{eq:old_unconstrained_problem}\hspace{5em}
    \tcc{primal update}
    $\bm{\lambda}^e \xleftarrow{} \bm{\lambda}^{e-1} + \mu^{e-1} \bm{\mathcal{C}}(\theta^e)$ \hspace{7em}
        \tcc{dual update}
    $\mu^e \xleftarrow{} \min( \beta~\mu^{e-1}, \mu_{\max})$ \hspace{7em}
        \tcc{penalty parameter update}
}
\KwOutput{$\theta^E$}\\
\KwDefaults{$\beta = 2, ~\mu_{\max} = 1 \times 10^4$}\\
\caption{ALM with monotonic exponential penalty update (MPU) \cite{martins2021engineering}}
\label{alg:mpu_classic_training_algorithm}
\end{algorithm}

In the context of neural network training, Algorithm~\ref{alg:mpu_classic_training_algorithm} takes as input the initial model parameters $\theta^0$, a maximum penalty cap $\mu_{\max}$, a multiplicative growth factor $\beta$ for updating the penalty parameter $\mu$, and the total number of training epochs $E$. We should note that in Algorithm \ref{alg:mpu_classic_training_algorithm}, both the Lagrange multipliers and the penalty parameter are updated at each iteration \cite{martins2021engineering}. To prevent divergence, the maximum penalty parameter was set to $10^4$ in default. However, updating the penalty parameter every epoch may lead to overly aggressive growth, potentially disrupting the balance between the objective and constraint terms, as will be shown in a later section. Additionally, selecting an appropriate maximum penalty parameter is often challenging and problem-dependent.

Another strategy to update the penalty parameter $\mu$ is to update it only when the constraints have not decreased sufficiently at the current iteration \cite{bierlaire2015optimization,wright1999numerical}. Algorithm 
\ref{alg:cpu_standard_training_algorithm}, presents an ALM with a conditional penalty parameter update strategy (CPU). A similar strategy has been adopted in the work of \citet{dener2020training} to train an encoder-decoder neural network for approximating the Fokker-Planck-Landau collision operator. 

\IncMargin{1em}
\begin{algorithm}[!h]
\SetAlgoLined
\SetKw{KwInput}{Input:}
\SetKw{KwOutput}{Output:}
\SetKw{KwDefaults}{Defaults:}
\KwInput{$\theta^0$, $\beta$, $\mu_{\max}$, $E$}\\
$\bm{\lambda}^0 = \textbf{1}$ \hspace{12em}
\tcc{Initializing Lagrange multipliers}
$\mu^0 = 1 $ \hspace{12em}
\tcc{Initializing the penalty parameter}
$\mathcal{P}^0 = \infty$ \hspace{11em} 
\tcc{Initializing $l^2$ norm of constraints}
\BlankLine
\For{$e = 1  ~ \KwTo ~E...$}{
    $\theta^e \gets \text{Update}(\theta^{e-1})$ via Eq.~\ref{eq:old_unconstrained_problem} \hspace{1em}
    \tcc{primal update}
    \uIf{$\|\bm{\mathcal{C}}(\theta^e)\|_2 < ~ \mathcal{P}^{e-1}$}{
        $\bm{\lambda}^e \xleftarrow{} \bm{\lambda}^{e-1} + \mu^{e-1} \bm{\mathcal{C}}(\theta^e)$ 
        \hspace{2em}        \tcc{dual update}
        $\mu^e = \mu^{e-1}$ \hspace{8em}
        \tcc{penalty parameter unchanged}
        }
    \Else{
        $\mu^e \xleftarrow{} \min( \beta~\mu^{e-1}, \mu_{\max})$ \hspace{2em}
        \tcc{penalty parameter update}
        $\bm{\lambda}^e = \bm{\lambda}^{e-1}$ \hspace{8em}
        \tcc{dual unchanged}
      }
      $\mathcal{P}^e = \|\bm{\mathcal{C}}(\theta^e)\|_2$
}
\KwOutput{$\theta^E$}\\
\KwDefaults{$\beta = 2, ~\mu_{\max} = 1 \times 10^4$}\\
\caption{ALM with conditional penalty update (CPU) \cite{wright1999numerical}}
 \label{alg:cpu_standard_training_algorithm}
\end{algorithm}

Algorithm~\ref{alg:cpu_standard_training_algorithm} shares the same inputs as the previous method.
$\mathcal{P}$ is a placeholder for the previous $l^2$ norm of the constraints. As before, $\mu_{\max}$ serves as a safeguard to cap the penalty parameter, and $\beta$ controls its multiplicative growth. The penalty is again limited to $10^4$ to avoid numerical instability, though determining an appropriate upper bound remains a nontrivial task.

\section{Proposed Method: Conditionally Adaptive Augmented Lagrangian Method}
\label{sec:proposed_capu}

In ALM, when multiple constraints are present, a unique Lagrange multiplier is assigned to each constraint. However, as demonstrated in the MPU and CPU algorithms, a single penalty parameter, $\mu$, which is exponentially increased during training, is commonly employed. Our experience indicates that this approach is insufficient for learning solutions to complex PDEs involving multiple constraints with diverse characteristics--such as those arising from flux conditions, observational data, or governing physical laws. Therefore, we advocate assigning a distinct penalty parameter to each Lagrange multiplier and propose an adaptive update strategy, rather than exponentially increasing $\mu$. This individualized approach allows each penalty parameter to evolve in accordance with the behavior of its associated constraint, thereby improving both training stability and constraint enforcement.

Formally, the augmented Lagrangian loss with constraint‑specific penalties is
\begin{align}
    \max_{\bm{\lambda}} \min_{\theta} \mathcal{L}(\theta,\bm{\lambda};\mu) = \mathcal{J}(\theta) + \bm{\lambda}^T \bm{\mathcal{C}}(\theta) + \frac{1}{2} \bm{\mu}^T \big[\bm{\mathcal{C}}(\theta) \odot \bm{\mathcal{C}}(\theta)\big].
    \label{eq:proposed_max_min}
\end{align}
where $\odot$ denotes the element-wise (Hadamard) product, $\bm{\lambda}$ is a vector of Lagrange multipliers, and $\bm{\mu}$ represents a vector of positive penalty parameters, with $\mu_i$ associated to constraint $\mathcal{C}_i$. 
The corresponding dual update at training epoch $e$ is
\begin{equation}
\bm{\lambda}^{e} \leftarrow \bm{\lambda}^{e-1} + \bm{\mu}^{e-1} \odot \bm{\mathcal{C}}(\theta^{e}).
\label{eq:dual_update}
\end{equation}

The key challenge is to update these penalty parameters independently and adaptively. 
Notably, since the gradient of the augmented loss \eqref{eq:proposed_max_min} with respect to $\bm{\lambda}$ is precisely the set of constraints $\bm{\mathcal{C}}$, the dual update \eqref{eq:dual_update} in ALM shares a formal resemblance to a gradient ascent step on a concave objective. In this sense, designing adaptive updates for distinct penalty parameters can be loosely regarded as analogous to developing adaptive learning rates in unconstrained optimization algorithms. 

We draw inspiration from one of the adaptive algorithms for convex optimization, the RMSProp method proposed by Hinton \cite{RMSprop_hinton}. RMSProp adjusts individual learning rates based on the moving averages of squared gradients, building an adaptive path of step sizes responsive to gradient history. If we apply this adaptive strategy for the dual update stage of ALM \eqref{eq:dual_update}, the moving average of the squared $i$-th constraint is expressed as
\begin{equation}
\bar{v}_i \xleftarrow{} \zeta \bar{v}_i + (1 - \zeta)\mathcal{C}_i^2(\theta),
\label{eq:rms_history}
\end{equation}
where $\zeta$ is the smoothing coefficient (default $\zeta = 0.99$). The corresponding penalty parameter is then computed as
\begin{equation}
\mu_i^{(RMSprop)} \xleftarrow{} \frac{\eta_i}{\sqrt{\bar{v}_i + \epsilon}},
\label{eq:rmsprop_mu}
\end{equation}
where $\eta_i$ is a specific scalar that serves as the penalty scaling factor, and $\epsilon$ is a small constant (default $\epsilon = 10^{-16}$) for numerical stability.
However, this inverse dependence on the moving average of the \textit{i}-th constraint $\bar{v}_i$ creates a serious conflict with the fundamental principle of ALM, which is enforcing stronger penalties when constraint violations are large. Instead, with \eqref{eq:rms_history} and \eqref{eq:rmsprop_mu}, when a constraint value gradually increases for a short period during training (so that $\bar{v}_i$ rise), $\mu_i^{(RMSprop)}$ inadvertently decreases, which runs counter to the aforementioned principle. Equally important, the resulting update of the Lagrange multiplier is
\begin{equation}
\lambda_i \xleftarrow{} \lambda_i + \mu_i^{(RMSprop)}\mathcal{C}_i(\theta)
= \lambda_i + \eta_i\frac{\mathcal{C}_i(\theta)}{\sqrt{\bar{v}_i + \epsilon}},
\label{eq:rmsprop_lambda}
\end{equation}
which notably provides no direct control over the penalty parameter, $\mu$. In this rule (Eq. \ref{eq:rmsprop_lambda}), the growth of each $\lambda_i$ is governed solely by the constant $\eta_i$ and the ratio between the current constraint $\mathcal{C}_i(\theta)$ and its RMS-averaged history $\sqrt{\bar{v}_i}$. Since this ratio typically remains close to unity, $\lambda_i$ keeps nearly the same rate of growth, essentially decided by $\eta_i$, even when constraint violations increase. To illustrate these pathologies in practice, we present a specific failure case in~\ref{sec:appendixa}. Figure~\ref{fig:helm_multi_l4_pred_evol}(a) shows the resulting Lagrange multipliers and penalty parameter, obtained by applying the rule in Eq.~\ref{eq:rmsprop_lambda}, plotted as a function of epochs.

To address the inconsistency arising from directly adopting RMSProp’s adaptive strategy, we introduce a safeguard by applying a maximum operation that retains the larger of the previous penalty parameter and its updated value from RMSProp. 
\begin{equation}
\mu_i \xleftarrow{} \max(\mu_i, \frac{\eta_i}{\sqrt{\bar{v}_i + \epsilon}}).
\label{eq:capu_mu}
\end{equation}
Despite its simplicity, this modification drastically changed the evolution of both Lagrange multipliers and penalty parameters during training, proving highly effective, as demonstrated in Figure~\ref{fig:helm_multi_l4_pred_evol}(b). By safeguarding the penalty parameter, unintended reductions during training are prevented, while the corresponding Lagrange multiplier can be rapidly amplified via Eq. \eqref{eq:dual_update} in response to increased constraint violations. As a result, the optimization process is guided back toward feasibility precisely at the stages when constraint satisfaction is most critical. At the same time, the RMSProp contribution in \eqref{eq:capu_mu} allows $\mu_i$ to increase gradually when constraint violations diminish smoothly during optimization, thereby strengthening enforcement progressively as constraint satisfaction is approached.

In the literature on adaptive learning-rate optimizers, AMSGrad \cite{reddi2018amsgrad} employs a related mechanism to address the convergence deficiencies of Adam \cite{kingma2015adam} by replacing the exponentially weighted second-moment estimates with a non-decreasing sequence defined via a maximum operator. However, AMSGrad is not suitable for our ALM context. Preventing a drift toward infeasibility here would require an AMSGrad-like method to employ a minimum operator on the historical squared constraint values and set the first-moment estimate equal to the current constraint, a configuration that maintains consistency with the dual update rule in Eq.~\eqref{eq:dual_update}.

Another important consideration is whether it's necessary to update the dual variables $\bm{\lambda}$ at every epoch. Consistent with several studies \citep{Birgin2014, echebest2016ALMconvergence, Andreani2024ALM}, our viewpoint is that such frequent updates are unnecessary until the current unconstrained optimization subproblem for minimizing with respect to $\theta$ (see Eq. \eqref{eq:proposed_max_min}) has converged sufficiently, given the previous values $\bm{\lambda}^{e-1}$ and $\bm{\mu}^{e-1}$.
We define the rate of decrease of the augmented Lagrangian loss between successive epochs as
\begin{equation}
    \frac{\mathcal{L}^{e}}{\mathcal{L}^{e-1}}=\omega,
\end{equation}
such that we simply declare convergence when $\omega_t < \omega \leq 1.0$, where $\mathcal{L}^e$ and $\mathcal{L}^{e-1}$ denote the current and previous augmented Lagrangian loss, respectively, and $\omega_t \in (0,1)$ is a user-defined threshold. The default value here is $\omega_t = 0.999$.
Meanwhile, $\omega > 1$ indicates that the gradient-based optimizer may perform stochastic updates leading to greater constraint violations.
Once either condition is satisfied, i.e., $\omega \geq \omega_t$, the Lagrange multipliers and penalty parameters are updated accordingly, and the training proceeds to the next unconstrained optimization subproblem with updated $\bm{\lambda}^e$ and $\bm{\mu}^e$. 

\IncMargin{1em}
\begin{algorithm}[!h]
\SetAlgoLined
\SetKw{KwInput}{Input:}
\SetKw{KwOutput}{Output:}
\SetKw{KwDefaults}{Defaults:}
\KwDefaults{$\zeta = 0.99,~\omega_t = 0.999,~ \epsilon = 10^{-16}$}\\

\KwInput{$\theta^0$, $E$, $\bm{\eta}$}\\
$\bm{\lambda}^0 = \textbf{1}$ \hspace{7em}
\tcc{Initializing Lagrange multipliers}
$\bm{\mu}^0 = \textbf{1}$ \hspace{7em}
\tcc{Initializing penalty parameters}
$\bar{\textbf{v}}^0 = \textbf{0}$ \hspace{7em}
\tcc{Initializing averaged square-gradients}
$\mathcal{L}^0 = \infty$ \hspace{6em} 
\tcc{Initializing augmented Lagrangian loss}
\BlankLine
\For{$e = 1  ~ \KwTo ~E...$}{
    $\theta^e \gets \text{Update}(\theta^{e-1})$\hspace{10em}
    \tcc{primal update}
    $\mathcal{L}^e \gets$  Eq.~\ref{eq:proposed_max_min} with $\theta^e, \bm{\lambda}^{e-1}, \bm{\mu}^{e-1}, \cdots$\hspace{3em}
    \tcc{augmented loss update}
    $\bar{\textbf{v}}^e  \xleftarrow{} \zeta ~\bar{\textbf{v}}^{e-1} + (1 - \zeta)~
    \left(\bm{\mathcal{C}}(\theta^e) \odot \bm{\mathcal{C}}(\theta^e)\right)$\hspace{2em}
    \tcc{square-gradient update}
    \uIf{$\mathcal{L}^e/ \mathcal{L}^{e-1} \geq \omega_t$ }
        {
        $\bm{\lambda}^e \xleftarrow{} \bm{\lambda}^{e-1} + \bm{\mu}^{e-1} \odot \bm{\mathcal{C}}(\theta^e)$ \hspace{6em}
        \tcc{dual update}
        $\bm{\mu}^e \xleftarrow{} \max\left( \bm{\mu}^{e-1}, \dfrac{\bm{\eta}}{\sqrt{\bar{\textbf{v}}^e + \epsilon}}\right)$ \hspace{6em}
        \tcc{penalty update}
        }
    }
\KwOutput{$\theta^E$}\\
 \caption{ALM with conditionally adaptive penalty updates (CAPU)}
 \label{alg:capu_adaptive_training_algorithm}
\end{algorithm}

The complete training procedure is presented in Algorithm~\ref{alg:capu_adaptive_training_algorithm}. Here, the smoothing constant $\zeta$ and the stability parameter $\epsilon$ are set to their default values from RMSProp. The convergence threshold is specified as $\omega_t = 0.999$. The vector of penalty scaling factors $\bm{\eta} = \{\eta_i: i \in \mathcal{E}\}$ is provided as an input to the algorithm, along with the initialized neural network parameters $\theta^{0}$ and the total number of training epochs $E$. The Lagrange multiplier vector $\bm{\lambda}$ and the penalty parameter vector $\bm{\mu}$ are both initialized to 1.0, and the moving averages of squared gradients, $\bar{\mathbf{v}}$, are initialized to zero. The initial augmented loss $\mathcal{L}^0$ is set to infinity to serve as a placeholder for the prior epoch's loss.

In~\ref{sec:appendixc}, we follow existing proofs of optimality and feasibility of the well-established augmented Lagrangian method, and present a proof demonstrating the boundedness of the penalty parameters and the convergence property of the Lagrange multipliers associated with the CAPU algorithm.

The vector $\bm{\eta}$ plays a key role in controlling the growth rate of the Lagrange multipliers $\bm{\lambda}$, as previously discussed. The optimization framework in our setting, as defined in \eqref{eq:proposed_max_min}, includes both primal ($\theta$) and dual ($\bm{\lambda}$) updates. If the dual update overwhelms the other, rapid growth of $\bm{\lambda}$ can hinder convergence by dominating the objective, similar to issues encountered in MPU and CPU strategies. Conversely, if $\bm{\lambda}$ grows too slowly, the multipliers may take longer to reach effective magnitudes, resulting in slow progress toward feasibility.
In RMSProp for gradient descent, a typical recommended value for the learning rate is 0.01.
Since RMSProp is a first-order optimizer, we recommend using $\eta_i = 0.01$ as well, when the primal update employs a first-order method like Adam \cite{kingma2015adam}. In contrast, for quasi-Newton optimizers such as L-BFGS \citep{nocedal1980updating}, which converge faster, a larger value such as $\eta_i = 1$ is advisable to balance constraint enforcement and objective minimization effectively. A sensitivity study of external input $\eta_i$ and internal hyperparameters $\zeta$, $\omega_t$ is presented and compared against other popular physics-informed approaches in~\ref{sec:appendixb}.

\subsection{Constrained optimization formulation for solving forward and inverse problems}
We begin by formulating the general constrained optimization problem for learning a generic PDE solution subject to boundary and initial conditions. This formulation is designed to accommodate the proposed CAPU algorithm and incorporate a constraint aggregation technique.

Consider a generic PDE problem for a scalar function \(u(\boldsymbol{x},t): \mathbb{R}^{d+1} \rightarrow \mathbb{R}\) on the domain \(\Omega \subset \mathbb{R}^d\) with its boundary \(\partial \Omega\):
\begin{equation}
    \begin{aligned}
            \mathcal{F}(\boldsymbol{x},t;\frac{\partial u}{\partial t}, \frac{\partial^2 u}{\partial t^2},\cdots,\frac{\partial u}{\partial \boldsymbol{x}}, \frac{\partial^2 u}{\partial \boldsymbol{x}^2},\cdots,\boldsymbol{\nu}) &= 0,\quad\forall (\boldsymbol{x},t) \in \mathcal{U},\\
    \mathcal{B}(\boldsymbol{x},t,g;u, \frac{\partial u}{\partial \boldsymbol{x}},\cdots) &= 0, \quad\forall (\boldsymbol{x},t) \in \partial \mathcal{U},\\
    \mathcal{I}(\boldsymbol{x},t,h;u,\frac{\partial u}{\partial t},\cdots) &= 0, \quad \forall (\boldsymbol{x},t) \in \Gamma,
    \end{aligned}
    \label{eq:general_pde_problem}
\end{equation}
where $\mathcal{F}$ is the PDE residual involving differential operators and parameter vector $\boldsymbol{\nu}$; $\mathcal{B}$ and $\mathcal{I}$ are the residuals of the boundary condition (BC) and initial condition (IC), incorporating source functions $g(\boldsymbol{x},t)$ and $h(\boldsymbol{x},t)$, respectively. The domain is defined as $\mathcal{U} = \{(\boldsymbol{x},t) ~|~\boldsymbol{x} \in \Omega, t = [0,T] \}$, with boundary $\partial \mathcal{U} = \{(\boldsymbol{x},t) ~|~\boldsymbol{x} \in \partial \Omega, t = [0,T]\}$, and initial surface $\Gamma = \{(\boldsymbol{x},t) ~|~\boldsymbol{x} \in \partial \Omega, t = 0\}$.

\subsubsection{Original PECANN formulation with point-wise constraints}
\label{sec:constrained_optimization_formulation_pointwise_constraint}
Considering the generic PDE problem \eqref{eq:general_pde_problem}, \citet{PECANN_2022} formulated the PECANN method as a constrained optimization problem where boundary and initial conditions at each collocation point were treated as a constraint: 
\begin{equation}
    \begin{aligned}
        \min_{\theta} ~& \sum_{i=1}^{N_{F}}\|\mathcal{F}(\boldsymbol{x}^{(i)},t^{(i)};\boldsymbol{\nu},\theta)\|_2^2, \\
        \text{subject to } & \phi(\mathcal{B}(\boldsymbol{x}^{(i)},t^{(i)},g^{(i)};\theta)) = 0,~ \forall (\boldsymbol{x}^{(i)},t^{(i)},g^{(i)}) \in \mathcal{\partial U},~ i = 1,\cdots, N_{B}, \\
            & \phi(\mathcal{I}(\boldsymbol{x}^{(i)},t^{(i)},h^{(i)};\theta)) = 0,~ \forall (\boldsymbol{x}^{(i)},t^{(i)},h^{(i)}) \in \Gamma,~ i = 1,\cdots, N_{I},
    \end{aligned}
    \label{eq:pntw_formulation}
\end{equation}
where $N_{F}$, $N_{B}$, $N_{I}$ are the number of collocation points in $\mathcal{U}$, $\partial \mathcal{U}$ and $\Gamma$ respectively. $\phi$ is a convex distance function, which we discuss in more detail in section \ref{sec:phi_&_sa_pinn}. We should note that number of Lagrange multipliers scale with the number of constraints $N_{B}$, $N_{I}$. For PDE problems, the number of collocation points typically grows with the complexity of the solution, leading to thousands of constraints that must be satisfied (\(N_{B}\) and \(N_{I}\) are large). Consequently, the efficiency of the optimization process degrades. Moreover, point-wise constraint enforcement hinders the adoption of standard deep learning techniques such as mini-batch training. To address these limitations, we propose a new formulation based on constraint aggregation. 

\subsubsection{Revised PECANN formulation with constraint aggregation}
\label{sec:proposed_formulation}
In large-scale design or topology optimization, failure criteria are often expressed as maximum or minimum functions, which are inherently non-differentiable. To address this, constraint aggregation techniques have been developed to approximate these criteria with continuously differentiable functions, enabling sensitivity analysis via the adjoint method \cite{Poon2006_constraint_aggregation, Kennedy2015_constraint_aggregation, Zhang2018_constraint_aggregation}.

As illustrated in the example problem in Section~\ref{sec:composite_heat}, when the number of point-wise constraints becomes large, Lagrange multipliers associated with a specific type of constraint tend to follow a probabilistic distribution with a well-defined expected value. Motivated by this observation, we aggregate point-wise constraints on each boundary using the mean squared residuals (MSR) metric, ensuring smoothness and differentiability. This aggregation reduces the number of Lagrange multipliers from thousands to just a few, making them easier to manage and enabling mini-batch training when full-batch training becomes impractical due to memory limitations.

Considering the generic PDE problem defined in Eq.~\eqref{eq:general_pde_problem}, we now express our new formulation with constraint aggregation as follows:
\begin{equation}
    \begin{aligned}
        \min_{\theta} ~& \mathcal{J}(\theta) = \frac{1}{N_{F}}\sum_{i=1}^{N_{F}}\|\mathcal{F}(\boldsymbol{x}^{(i)},t^{(i)};\boldsymbol{\nu},\theta)\|_2^2, \\
        \text{subject to } & \mathcal{C}_B = \frac{1}{N_{B}} \sum_{i=1}^{N_{B}} \phi( \mathcal{B}(\boldsymbol{x}^{(i)},t^{(i)},g^{(i)};\theta)) :=0, \\
            & \mathcal{C}_I = \frac{1}{N_{I}} \sum_{i=1}^{N_{I}}\phi( \mathcal{I}(\boldsymbol{x}^{(i)},t^{(i)},h^{(i)};\theta)) :=0.
    \end{aligned}
    \label{eq:avg_formulation}
\end{equation}
where $\phi$ is a quadratic function, but other options are possible \cite{Kennedy2015_constraint_aggregation}.

In \ref{sec:appendixc_comp}, we present a derivation establishing a connection between the point-wise constraint formulation and the aggregated constraint formulation. Interestingly, an extra penalty term dependent on the variance of the constraints appears with point-wise enforcement, a term that can be large for complex problems with challenging constraints. The benefits of using constraint aggregation on accuracy levels are demonstrated in Section \ref{sec:composite_heat}.


%
\subsubsection{Extension to PDE-constrained inverse problems}\label{subsec:inverse_formulation}
In the original PECANN approach for inverse problems \cite{PECANN_2022}, the loss function was minimized in a composite manner, incorporating both the governing PDE and noisy observational data into a single objective, while high-fidelity data were enforced as constraints. However, discrepancies in physical scales between PDE terms and noisy measurements can hinder the inference process. To address this challenge in inverse problems with multi-fidelity data, we propose a reformulation that prioritizes minimizing the loss associated with noisy observations, while treating the governing PDE and any available high-fidelity data as constraints. Notably, this reformulation is made possible by the design of our CAPU algorithm, which assigns independent penalty parameters to each constraint and updates them adaptively. In the absence of noisy data, the PDE residual naturally serves as the objective, with high-fidelity data enforced as constraints.

To formulate this revision, consider a generic inverse PDE problem with unknown parameters $\boldsymbol{\nu}$, and a set of noisy measurements $\{(\boldsymbol{x}^{(i)},t^{(i)}), \hat{u}^{(i)} \}_{i=1}^{N_{M}}$. The problem can be formulated as follows:
\begin{equation}
    \begin{aligned}
        \min_{\theta,\boldsymbol{\nu}} ~ & J(\theta) = \frac{1}{N_{M}}\sum_{i=1}^{N_{M}}\|u(\boldsymbol{x}^{(i)},t^{(i)};\theta) - \hat{u}(\boldsymbol{x}^{(i)},t^{(i)}) \|_2^2, \\
        \text{subject to } & \mathcal{C}_F = \frac{1}{N_{F}}\sum_{i=1}^{N_{F}}\phi (\mathcal{F}(\boldsymbol{x}^{(i)},t^{(i)};\boldsymbol{\nu},\theta)) :=0,\\
            & \mathcal{C}_B := 0; \quad  \mathcal{C}_I := 0; \quad \cdots
    \end{aligned}
    \label{eq:inv_formulation}
\end{equation}
If high-fidelity data is available, we can incorporate it as an additional constraint into the above formulation along with other constraints. In the context of CAPU, the choice of penalty scaling factors $\bm{\eta}$ should reflect the relative priority of constraints. Specifically, the scaling factor for the PDE constraint $\eta_F$ is recommended to be smaller than those for high-fidelity data sources ($\eta_B$ \& $\eta_I$), such as boundary and initial conditions. This prioritization encourages the optimization process to satisfy boundary and initial conditions before focusing on reducing the PDE residual. 

\subsubsection{Role of convex distance functions in constraint enforcement and aggregation}\label{sec:phi_&_sa_pinn}
In the original PECANN formulation with point-wise constraints, a quadratic distance function $\phi$ is adopted \cite{PECANN_2022}. In the present revised formulation (Eq. \ref{eq:avg_formulation}), we adopt a constraint aggregation technique based on the mean squared residual (MSR) metric, which inherits the quadratic distance function $\phi$ implicitly. Consequently, the associated penalty terms in ALM are proportional to the fourth power of the operator residuals in both formulations. Note that the augmented Lagrangian loss has separate contributions from the penalty term and the Lagrange multiplier (see Eq. \ref{eq:old_unconstrained_problem}). For the $i$th boundary constraint,
\begin{equation}
    \phi = |\cdot|^2 \Rightarrow \mathcal{C}_B^{(i)} \propto |\mathcal{B}^{(i)}|^2, \quad \frac{1}{2} \mu_B \mathcal{C}_B^{(i)2} \propto |\mathcal{B}^{(i)}|^4.
\end{equation}
When $\phi$ is an absolute-value function, the aggregated constraint reduces to the mean absolute error (MAE), leading to penalty terms proportional to the square of the residuals of the boundary condition operator
\begin{equation}
    \phi = |\cdot| \Rightarrow \mathcal{C}_B^{(i)} \propto |\mathcal{B}^{(i)}|, \quad \frac{1}{2} \mu_B \mathcal{C}_B^{(i)2} \propto |\mathcal{B}^{(i)}|^2.
\end{equation}\label{eq:distance_abs}
\citet{PECANN_2022} tested both options and found the quadratic distance function to be superior due to its smoothness and differentiability.

The choice of the distance function \(\phi \) provides a framework for interpreting alternative strategies within PINNs from the perspective of constrained optimization.
A notable example is the self-adaptive PINN (SA-PINN) work \cite{mcclenny2020self}. Upon close examination of the SA-PINN formulation, we can see that the self-adaptive behavior is realized through a minimax formulation in the inverse form (i.e., weak duality):
\begin{equation}
    \max_{\bm{\omega}} \min_{\theta} \mathcal{J}(\theta) + m(\bm{\omega})^T [\bm{\mathcal{C}}(\theta) \odot \bm{\mathcal{C}}(\theta)],
    \label{eq:sapinn}
\end{equation}
where the objective $\mathcal{J}(\theta)$ is calculated of the MSR of the boundary conditions, and the constraint vector includes point-wise residuals of the PDE in the entire domain ($|\mathcal{F}(\bm{x}^{(i)}, t^{(i)})|$) and initial conditions ($|\mathcal{I}(\bm{x}^{(i)}, 0)|$). Eq.~\eqref{eq:sapinn} closely resembles the penalty terms given in Eq.~\eqref{eq:distance_abs}, specifically those that arise when $\phi$ is defined as an absolute-value function with point-wise enforcement of constraints.

A key element of the SA-PINN method is the mask function
\begin{equation}
m(\bm{\omega}) = c\bm{\omega}^{q},
\label{eq:mask}
\end{equation}
where $\bm{\omega}$ represents the self-adaptive weights and a polynomial function is assumed for the mask function. Note that in the SA-PINN paper \cite{mcclenny2020self}, self-adaptive weights are denoted by the symbol $\lambda$. Here, we use $\omega$ to distinguish it from the Lagrange multipliers, $\lambda$, used in ALM.

Comparing Eq.~\eqref{eq:sapinn} with the augmented Lagrangian in Eq.~\eqref{eq:proposed_max_min}, we observe that Eq.~\eqref{eq:sapinn} lacks the constraint term with the Lagrange multipliers ($\bm{\lambda}$), which are key elements of ALM for enforcing feasibility through optimality conditions and ensuring boundedness alongside the penalty parameter, $\bm{\mu}$ \cite{Birgin2014}. This distinction is subtle, yet it is critical for differentiating the PECANN methodology from the SA-PINN method.


Thus, the SA-PINN approach can be viewed as an adaptive penalty method distinct from the augmented Lagrangian method in PECANN. In conventional penalty methods, achieving exact feasibility theoretically requires the penalty parameter to approach infinity, driving the solution to satisfy all constraints. But, in practice, such a large penalty parameter increase the condition number of the Hessian matrix, leading to ill-conditioning in the optimizer. The examples presented in \cite{mcclenny2020self} show that the self-adaptive weights $\bm{\omega}$ typically remain at a modest level below 20. It would therefore be a valuable exercise to systematically study SA-PINN performance when the mask functions are amplified toward much larger penalization values.

\subsection{Integration of Fourier feature mappings into the PECANN framework}


A Fourier feature mapping, $\gamma$, projects input coordinates $\mathbf{x} \in \mathbb{R}^d$ into a higher-dimensional space via sinusoidal transformations, thereby enhancing the expressive capacity of MLPs \cite{tancik_fourier_2020, wang_eigenvector_2021}. This technique is theoretically grounded in Bochner’s theorem and can be interpreted as a Fourier approximation of a stationary kernel function \cite{tancik_fourier_2020}, enabling the network to more effectively capture high-frequency components in the input space.

A single Fourier feature mapping is defined as:
\begin{equation}
\gamma(\mathbf{x}) =
\begin{bmatrix}
\cos(2\pi \mathbf{B} \mathbf{x}) \\
\sin(2\pi \mathbf{B} \mathbf{x})
\end{bmatrix}
\end{equation}
where $\mathbf{B} \in \mathbb{R}^{m \times d}$ consists of entries sampled from a Gaussian distribution $\mathcal{N}(0, \sigma)$, which remain fixed during training. The resulting transformed input is then fed into a standard fully connected neural network, with the following layer-wise operations:
\begin{align}
\mathbf{H}_1 &=  \gamma(\mathbf{x}), \\ 
\mathbf{H}_\ell &= \psi(\mathbf{W}_\ell \cdot \mathbf{H}_{\ell-1} + \mathbf{b}_\ell), \quad \ell = 2, \ldots, H, \\
f_\theta(\mathbf{x}) &= \mathbf{W}_{L+1} \cdot \mathbf{H}_L + \mathbf{b}_{L+1},
\end{align}
where $\psi$ represents the activation function.

According to \citet{wang_eigenvector_2021}, both the number of Fourier feature mappings and the choice of the scaling parameter $\sigma$ are problem-dependent. Typical values such as $\sigma = 1, 20, 50, 100$ enable the network to capture different frequency scales in the input function. Our experience has been different. In Section \ref{sec:1d_poisson}, we test $\sigma \in [0.25, 16]$ and demonstrate through numerical experiments that a single layer of Fourier feature mappings with the default setting $\sigma = 1$ is sufficient when combined with ALM and the CAPU strategy under the constrained optimization formulation of PECANN. Fourier feature mapping is only adopted for multi-scale problems with highly oscillatory solutions presented in Sections \ref{sec:1d_poisson} and \ref{sec:helmholtz_challenging}.

\section{Application to forward and inverse PDE problems}\label{sec:results}
We introduce several key enhancements to the original PECANN method~\cite{PECANN_2022} that improve its accuracy, stability, and generalization across diverse PDE problems. We construct a suite of benchmark problems, each specifically designed to isolate and evaluate a different aspect of the new framework. Before presenting detailed results, we summarize the key outcomes of each example problem.
\begin{itemize}
\item \S \ref{sec:composite_heat} (Unsteady heat transfer in a composite medium): This section analyzes the original PECANN formulation with point-wise constraint enforcement and demonstrates that the new formulation with constraint aggregation effectively satisfies boundary and flux continuity. Given the superior outcome of this example, the constraint aggregation formulation is adopted in all remaining examples.

\item \S \ref{sec:trans_rare} (Transonic rarefaction in the inviscid Burgers’ equation): This example provides a quantitative comparison of penalty update strategies within the ALM, demonstrating that PECANN-CAPU effectively controls network expressiveness in discontinuous and sub-differentiable regions without modifying the governing physics or the underlying method.

\item \S \ref{sec:1d_poisson} (1D Poisson's equation with multi-scale solutions): This example demonstrates that standard MLPs, when paired with either sufficiently dense collocation points or a single Fourier feature mapping ($\sigma =1$), can effectively learn complex solution structures within the PECANN-CAPU framework. The results challenge the prevailing assumption that a neural network architecture incapable of solving a regression task will necessarily fail when applied to PDE learning, thereby underscoring the critical role of the loss function formulation and the optimization strategy employed.

\item \S \ref{sec:2d_helm} (2D Helmholtz equation with high-frequency and multi-scale solutions): This example demonstrates that algorithmic innovation can rival architectural complexity. We compare our proposed PECANN-CAPU framework (using MLPs) against the physics-informed Chebyshev Kolmogorov-Arnold Networks (cPIKAN)~\cite{SHUKLA2024117290}, highlighting PECANN-CAPU’s ability to achieve competitive accuracy. Furthermore, we show that integrating MLPs with a single Fourier feature mapping yields additional gains in solution fidelity.

\item \S \ref{sec:reversible_adv} (Reversible advection of a passive scalar by a single vortex): This example highlights the PECANN-CAPU framework’s ability to perform stable, long-time integration of PDEs. By employing a non-overlapping time-windowing strategy, PECANN-CAPU achieves accuracy that is competitive with or superior to specialized computational methods designed for two-phase flow simulations.

\item \S \ref{sec:heat_source_reconstruction} (Inference of space-wise heat source in a transient heat conduction problem): This example evaluates the PECANN-CAPU framework with constraint aggregation for inverse PDE problems, demonstrating smooth and accurate reconstructions even when utilizing sparse and noisy data.

\item \S \ref{sec:appendixa} compares two constrained optimization methods, PECANN-CAPU and PECANN with direct application of RMSProp, under the constraint-aggregation formulation~\eqref{eq:avg_formulation}. The results demonstrate that PECANN-CAPU consistently drives the optimization toward feasibility with the fastest convergence rate.

\item \S \ref{sec:appendixb} examines the sensitivity of the PECANN-CAPU results to its internal hyperparameters and one external input. The accuracy of these results is assessed relative to other published works.

\item \S \ref{sec:appendixc} presents the theoretical underpinnings of the ALM, the mathematical relationship between the point-wise and aggregation formulations for constraints, and an analysis of the current CAPU algorithm regarding the boundedness of its penalty parameters and the convergence behavior of its Lagrange multipliers.

\end{itemize}

We define two metrics to assess the accuracy of our predictions. Given an $n$-dimensional vector of predictions $\boldsymbol{\hat{u}} \in \mathbf{R}^n$ and an $n$-dimensional vector of exact values $\boldsymbol{u} \in \mathbf{R}^n$, the relative Euclidean or $\mathit{l^2}$ norm and infinity norm $\mathit{l^\infty}$ norm respectively
\begin{align}
     \mathcal{E}_r(\hat{u},u) = \frac{\|\hat{\boldsymbol{u}} - \boldsymbol{u}\|_2}{\|\boldsymbol{u}\|_2}, 
    \label{eq:relative_L2_Error} \quad
    \mathcal{E}_{\infty}(\hat{u},u) = \| \boldsymbol{\hat{u}} - \boldsymbol{u}\|_{\infty}, \quad
\end{align}
where $\|\cdot \|_2$ denotes the Euclidean norm, and $\| \cdot \|_{\infty}$ denotes the maximum norm. 

Unless otherwise stated, we use standard MLPs parameterized by $\theta$, with hyperbolic tangent activation functions and Xavier initialization scheme \cite{glorot2010understanding}. When selected, L-BFGS optimizer available in the Pytorch package \cite{paszke2019pytorch} is employed with \emph{strong Wolfe} line search function. The penalty scaling factor $\bm{\eta}$ in CAPU strategy is set to 0.01 when using Adam for the primal update, and 1.0 when using L-BFGS. For evaluations involving statistical performance such as the relative $l^2$ norm, the mean prediction and its standard deviation are calculated based on five independent trials.

\subsection{Forward problem: unsteady heat transfer in a composite medium}\label{sec:composite_heat}
We consider the time-dependent heat transfer problem in a composite material where temperature and heat fluxes are matched across the interface \cite{baker1985heat}. This canonical problem has initial condition, boundary condition and flux constraints, which makes it a suitable example to demonstrate the effectiveness of PECANN-CAPU with constraint aggregation and contrast it against the original PECANN with point-wise enforcement of constraints. 

We formulate the time-dependent heat conduction in a composite medium problem as follows:
\begin{align}
   \frac{\partial u(x,t)}{\partial t} &= \frac{\partial }{\partial x}[\kappa(x,t) \frac{\partial u(x,t)}{\partial x}] +  s(x,t), \quad (x,t) \in \Omega \times [0,\tau],
    \label{eq:heat_pde}
\end{align}
where $u$ is the temperature, $s$ is heat source, $\kappa$ is the thermal conductivity.
The domain $\Omega = \Omega_1 \cup \Omega_2$ consists of two non-overlapping subdomains: $\Omega_1 = \{x | -1 \le x < 0 \}$ and $\Omega_2 = \{x | 0 < x \le 1 \}$. To assess the accuracy of the learned solutions, we prescribe the thermal conductivity and exact solution as:
\begin{equation}
[\kappa, u]^T = 
\begin{cases} 
      [1, t\sin(3 \pi  x)]^T, & (x,t) \in \Omega_1 \times [0,2], \\
      [3 \pi, t  x]^T, & (x,t) \in \Omega_2 \times [0,2].
   \end{cases}
   \label{eq:exact_heat_solution}
\end{equation}  

Let us introduce an auxiliary flux parameter $q(x,t) = \kappa(x,t) \frac{\partial u}{\partial x}$ to reduce the original second order PDE into the following system of first-order PDEs 
\begin{subequations}
\begin{align}
\mathcal{F}(x,t) &:= \frac{\partial u(x,t)}{\partial t} -  \frac{\partial q(x,t) }{\partial x} -  s(x,t),\\
\mathcal{Q}(x,t) &:=  q(x,t) - \kappa(x,t) \frac{\partial u(x,t)}{\partial x},\\
\mathcal{B}(x,t) &:=  u(x,t) - g(x,t),\\
\mathcal{I}(x,t) &:=  u(x,0) - h(x,0),
\end{align}
\label{eq:heat_operators}
\end{subequations}
where $\mathcal{Q}$ represents the residuals of the flux operators.
The corresponding source term $s(x,t)$ and boundary/initial functions $g(x,t)$ and $h(x,t)$ can be calculated exactly using Eqs. \eqref{eq:exact_heat_solution} and \eqref{eq:heat_operators}.  

We use a MLP with two inputs $(x,t)$ and two outputs $(u,q)$, consisting of six hidden layers with 60 neurons each. 
The network is trained for 500,000 epochs using the Adam optimizer and a \textit{ReduceLROnPlateau} scheduler from PyTorch \cite{paszke2019pytorch}, which reduces the learning rate by a factor of 0.98 after 100 stagnant epochs, eventually driving it to a near-zero value.
For the purpose of this problem, $N_{F} = N_{\mathcal{Q}} =  8192$ collocation points are generated once before training, together with $N_{B} = 2 \times 8192$ boundary points and $N_{I} = 8192$ initial-condition points. 

\begin{figure}
    \centering
    \subfloat[]{\includegraphics[width=0.35\textwidth]{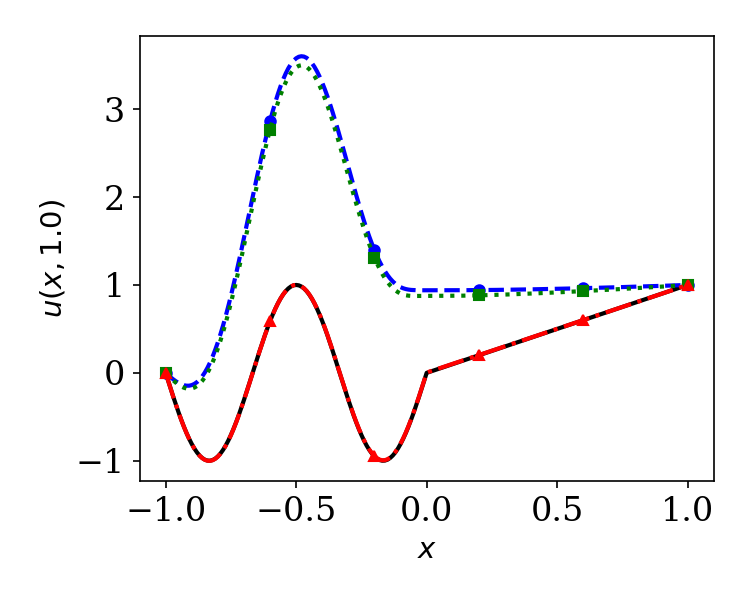}} \quad
    \subfloat[]{\includegraphics[width=0.35\textwidth]{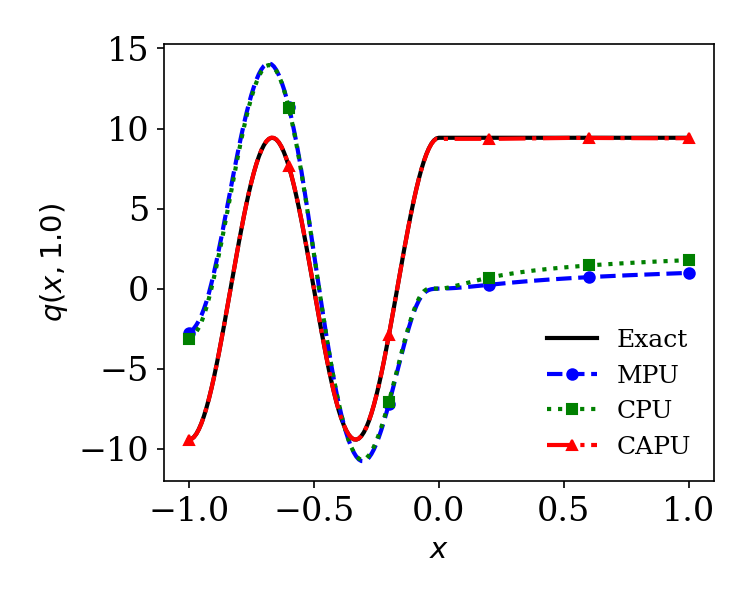}}
    \caption{Heat transfer in a composite medium: Profiles of (a) predicted temperature and (b) predicted heat flux at $t = 1$, obtained using different penalty update strategies with the point-wise constraint formulation from Section~\ref{sec:constrained_optimization_formulation_pointwise_constraint}.}
    \label{fig:existing_constrained_optimization_with_alm}
\end{figure}

We begin by solving the problem using the point-wise constraint formulation described in Section~\ref{sec:constrained_optimization_formulation_pointwise_constraint}. Results at $t=1$ are shown in Fig.~\ref{fig:existing_constrained_optimization_with_alm}, where panels (a) and (b) depict the temperature and heat flux distributions in the composite medium, respectively. Our proposed CAPU algorithm yields results in excellent agreement with the exact solution. In contrast, both the monotonic (MPU) and conditional (CPU) penalty update strategies produce solutions that deviate significantly from the reference, except at the boundary locations.

\begin{figure}[!h]
\centering
    \subfloat[]{\includegraphics[width=0.3\textwidth]{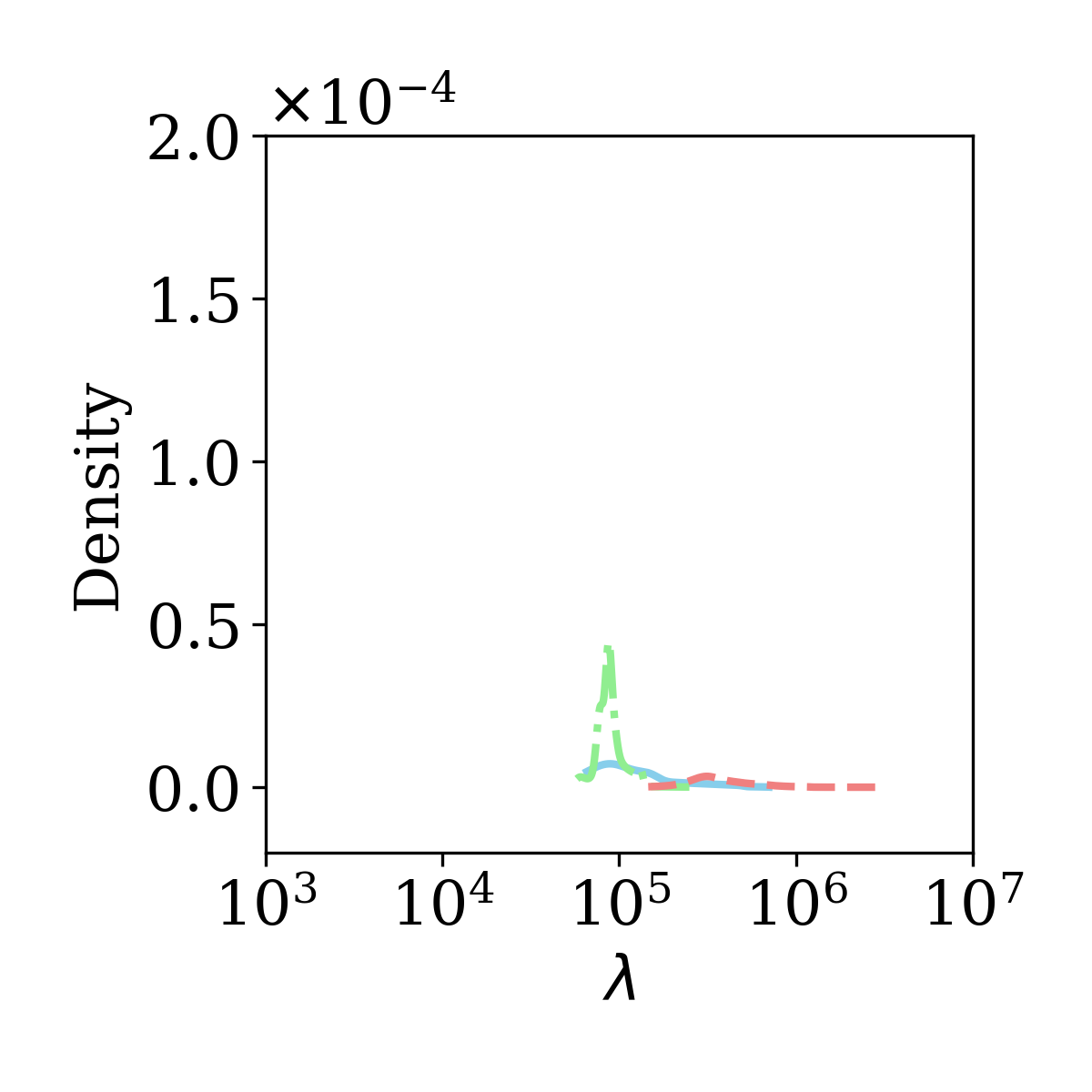}} \quad
    \subfloat[]{\includegraphics[width=0.3\textwidth]{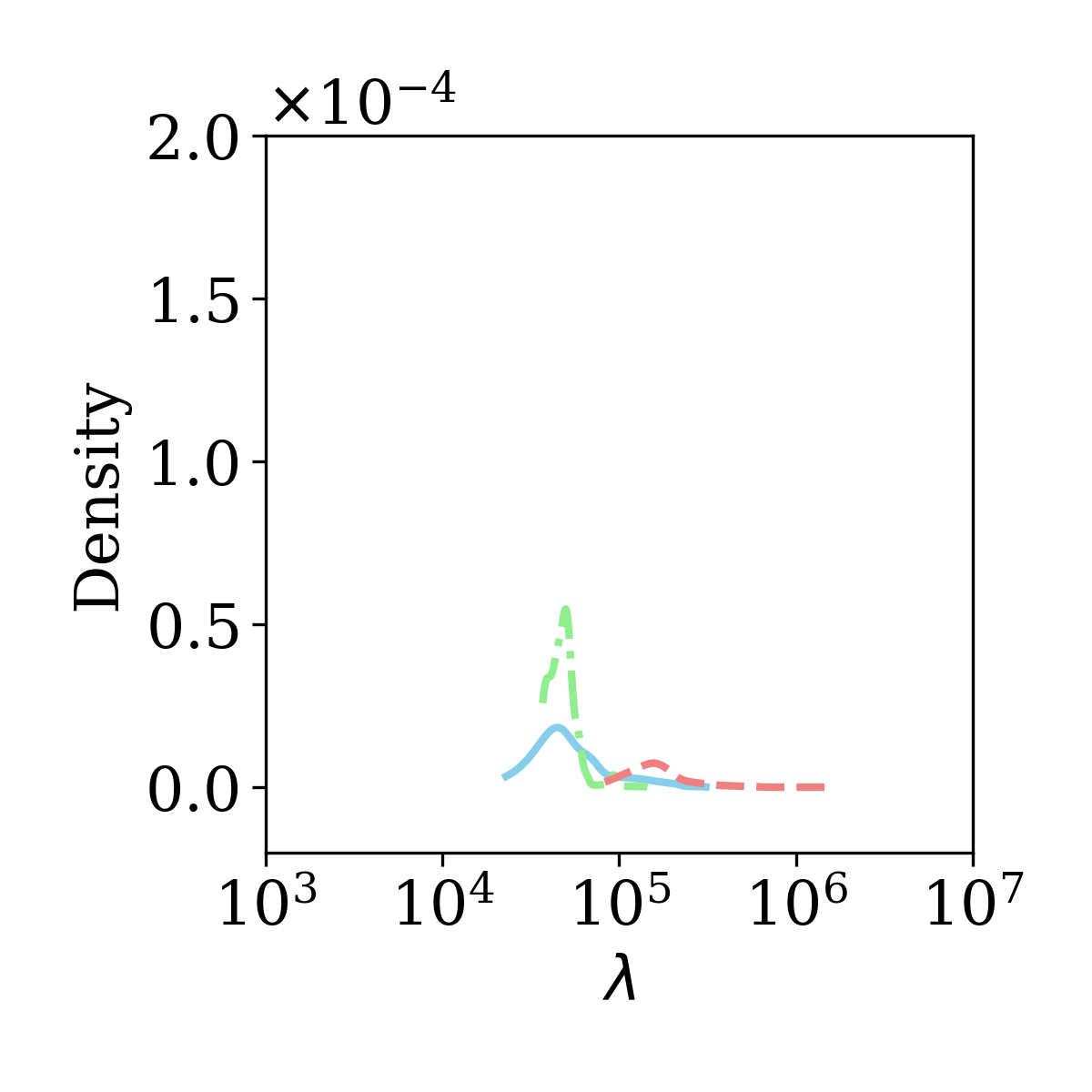}} \quad
     \subfloat[]{\includegraphics[width=0.3\textwidth]{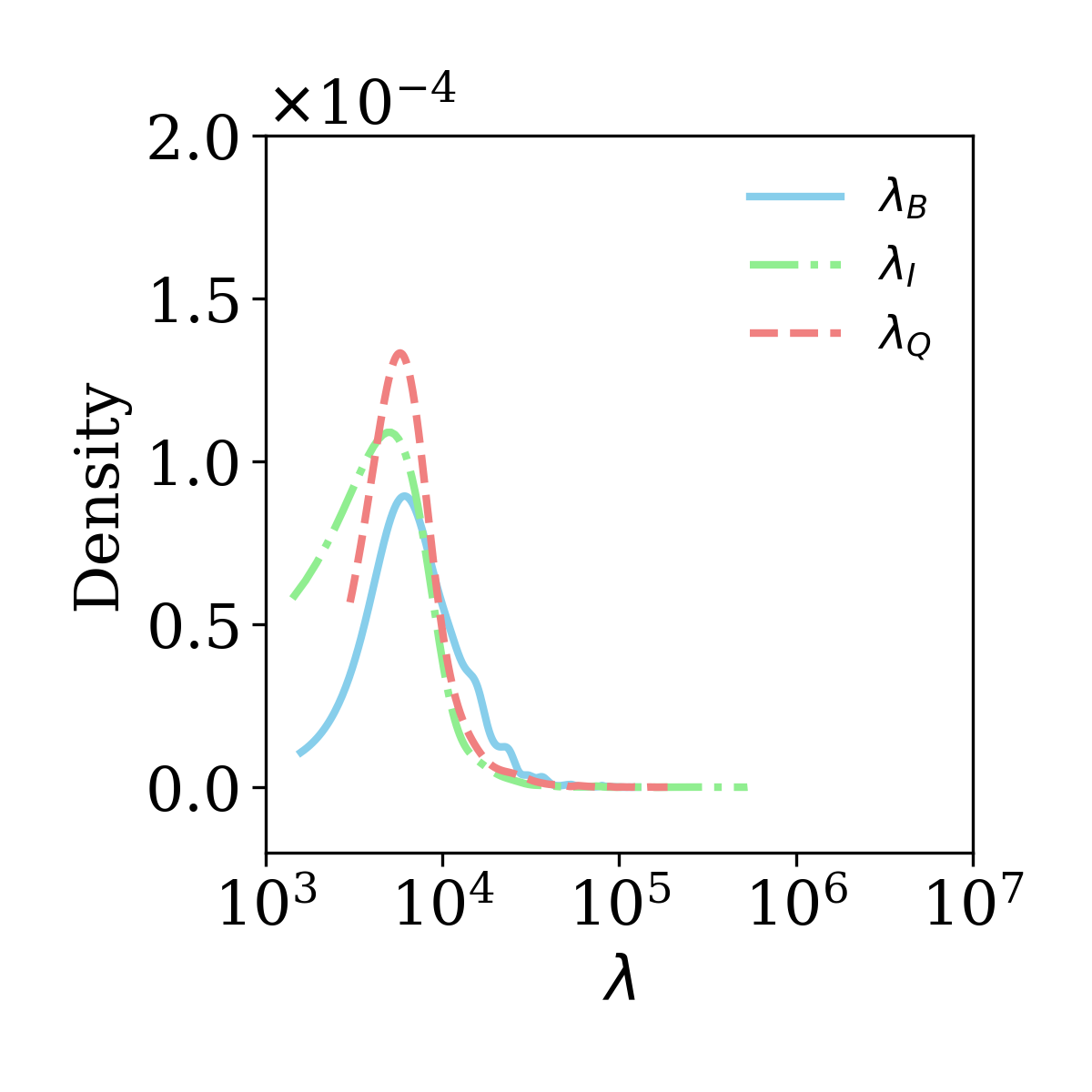}}
    \caption{Time-dependent heat conduction in a composite medium: Distribution of Lagrange multipliers obtained using the point-wise constraint formulation from Section~\ref{sec:constrained_optimization_formulation_pointwise_constraint}. (a) MPU (Algorithm 1); (b) CPU (Algorithm 2); (c) CAPU (Algorithm 3, proposed).}
\label{fig:distribution_of_lagrange_multipliers}
\end{figure}

Despite the strong performance of the CAPU strategy, constraints within the PECANN framework are enforced in a point-wise manner. 
In Fig.~\ref{fig:distribution_of_lagrange_multipliers}, we present the distribution of point-wise Lagrange multipliers at the final training epoch. 
Notably, the multipliers of each constraint type exhibit distinct and non-uniform distributions characterized by well-defined most probable values, particularly when the CAPU algorithm is employed, as shown in Fig.~\ref{fig:distribution_of_lagrange_multipliers}(c). This observation suggests that constraint enforcement strategies based on constraint aggregation--rather than point-wise enforcement at every collocation or observational data point--may be both viable and computationally advantageous.

\begin{figure}[!h]
\centering
    \subfloat{\includegraphics[width=1.\textwidth]{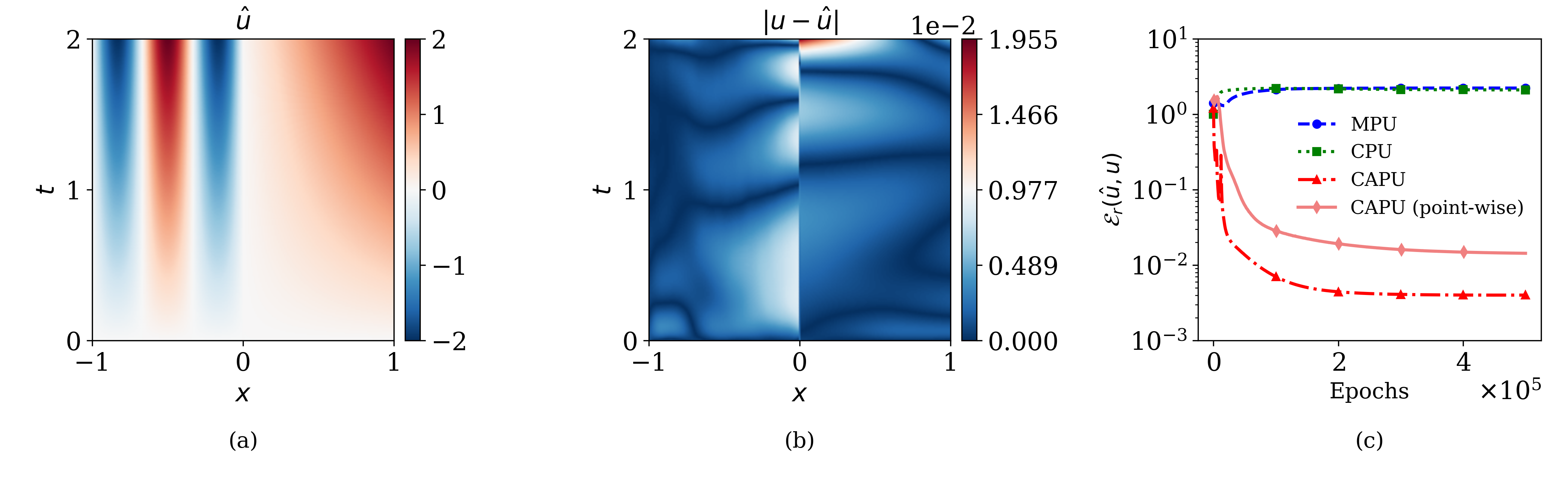}}
    \\
    \subfloat{\includegraphics[width=1.\textwidth]{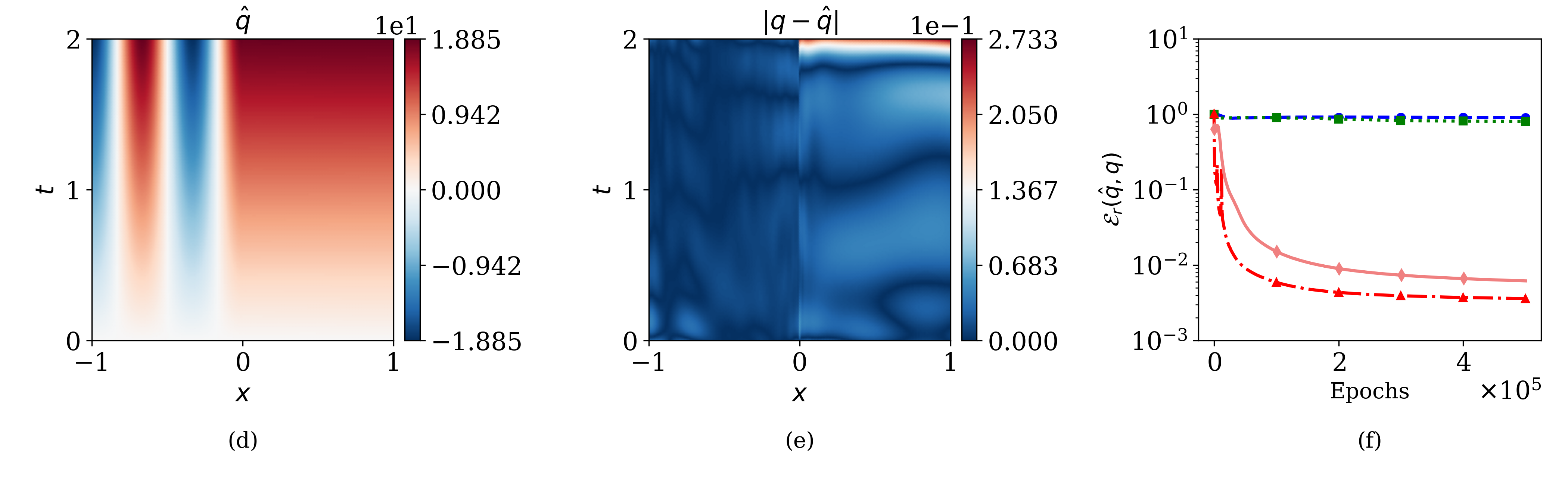}}
    \caption{Time-dependent heat conduction in a composite medium: Results are obtained using the constraint-aggregation formulation described in Section~\ref{sec:proposed_formulation}. The prediction using CAPU with point-wise constraints is also included. Panels (a)-(c) show the contours of predicted temperature, the absolute error obtained with CAPU, and the evolution of the relative $l^2$ error $\mathcal{E}_r(\hat{u}, u)$, respectively, with the latter including comparisons among MPU, CPU, and CAPU (point-wise); panels (d)–(f) present the corresponding heat flux results.}
    \label{fig:expected_global_constraint}
\end{figure}

Figure~\ref{fig:expected_global_constraint} first  illustrates the superior performance of the CAPU algorithm with the new aggregation formulation (Section~\ref{sec:proposed_formulation}), and then compares the evolution of evaluation metrics with that obtained using point-wise constraint enforcement.
Panels (a) and (b) show the space-time temperature solution and the absolute-error distribution using CAPU, while panels (d) and (e) display the corresponding heat flux and its error. These results demonstrate close agreement with the exact solutions.
Models trained with the MPU and CPU strategies failed to converge; therefore, their temperature and heat flux predictions are omitted. Their poor behavior of relative $l^2$ error $\mathcal{E}_r(\hat{u}, u)$ is nonetheless shown in panels (c) and (f) for comparison. 
These results demonstrate that, under the constraint-aggregation formulation~\eqref{eq:avg_formulation}, the CAPU algorithm consistently outperforms both MPU and CPU.

Compared with the result using point-wise enforcement in Fig.~\ref{fig:expected_global_constraint}(c) and (f), CAPU under constraint-aggregation formulation achieves faster convergence in the early stages of training and stabilizes at lower error norms.
The statistical performance of $\mathcal{E}_r(\hat{u}, u)$ further indicates that the new formulation with constraint aggregation achieves higher accuracy, with $5.42 \pm 2.41 \times 10^{-3}$ compared to $1.28\pm0.33 \times 10^{-2}$ for the point-wise formulation.

Overall constraint aggregation not only result in computational savings by reducing the number of Lagrange multipliers from thousands to a few, but it also improve accuracy levels. Based on these findings, we adopt PECANN with constraint aggregation for all remaining examples.

\subsection{Forward problem: transonic rarefaction in inviscid Burgers' equation}\label{sec:trans_rare}


Nonlinear hyperbolic conservation laws frequently encounter rarefaction wave and shock phenomena. Accurately capturing high gradients and discontinuities without introducing spurious oscillations is crucial and constitutes a challenging task for any numerical method. The baseline PINN approach is known to struggle learning the solution of PDEs with discontinuities. Several enhancements to the baseline PINN model have been proposed to improve upon this deficiency of the baseline PINN model \cite{ABBASI2025_challenges_shock}.
For instance, \citet{Liu2023} observed that training performance deteriorates due to transition points and introduced a local gradient-dependent weight into the governing equations.
Comparable strategies, including the addition of artificial viscosity~\cite{Coutinho2023avpinn}, help stabilize training by weakening the network’s representation capability near discontinuities.

To illustrate the versatility of our proposed PECANN-CAPU framework in handling problems characterized by discontinuities and sharp gradients, we examine the transonic rarefaction scenario governed by the inviscid Burgers' equation. This benchmark problem serves as a basis for quantitatively comparing three distinct penalty update strategies within ALM, corresponding to Algorithms 1–3. 

Specifically, the inviscid transonic rarefaction problem is considered within the spatiotemporal domain $\mathcal{U} = \{(x, t) \mid x \in [-0.5, 0.5], t \in [0, 1]\}$:
\begin{equation}
    \begin{aligned}
    \text{PDE: } & \frac{\partial u}{\partial t} + u \frac{\partial u}{\partial x} = 0, \\
    \text{IC: } & u(x,t=0) = 
        \begin{cases}
            -0.2, & x \leq 0, \\
            0.4, & x > 0,
        \end{cases} \\
    \text{BCs: } & 
        \begin{cases}
            u(x=-0.5,t) = -0.2, \\
            u(x=0.5,t) = 0.4.
        \end{cases}
    \end{aligned}
\end{equation}
The exact solution is given by:
\begin{equation}
u(x, t) =
\begin{cases} 
      -0.2, & \text{if } x \leq -0.2t, \\
      x/t, & \text{if } -0.2t \leq x \leq 0.4t, \\
      0.4, & \text{if } x \geq 0.4t.
\end{cases}
\end{equation}
The initial contact discontinuity evolves into a transonic rarefaction wave with a sign change in the characteristic speed, located at a sonic point ($u=0$) \cite{toro2009riemann}.  Accurately predicting this initial discontinuity is crucial, as any error will propagate and expand over time, creating an error-amplifying scenario.

In order to exclude the effects of random sampling, especially in high-gradient regions, uniform structured meshes are employed across the domain $\mathcal{U}$.
During training, the coarse mesh uses $66\times33$ points in space ($x$) and time ($t$), comprising $64 \times 32$ interior points, 64 initial condition points at $t = 0$, and 33 points per boundary. The fine mesh adopts a $130\times65$ resolution, and all results are evaluated on a $258 \times 129$ mesh.
The MLP employed comprises three hidden layers with 20 neurons per layer. We adopt Adam optimizer, with its default learning rate ($10^{-3}$) across 10,000 epochs. Note that we do not modify the governing physics in our approach.

\begin{figure}[!h]
\centering
    \subfloat[]{\includegraphics[scale=0.45]{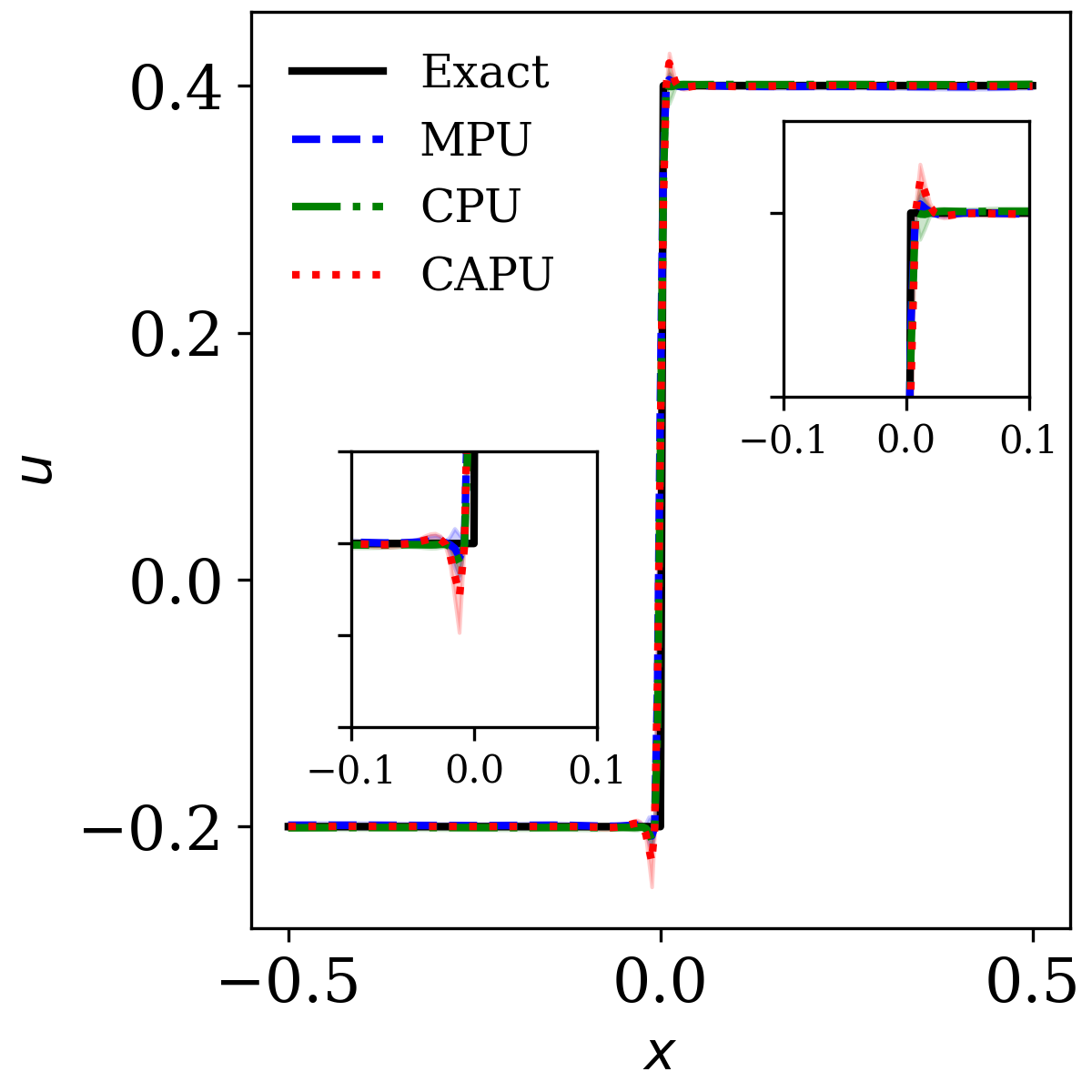}}\quad
    \subfloat[]{\includegraphics[scale=0.45]{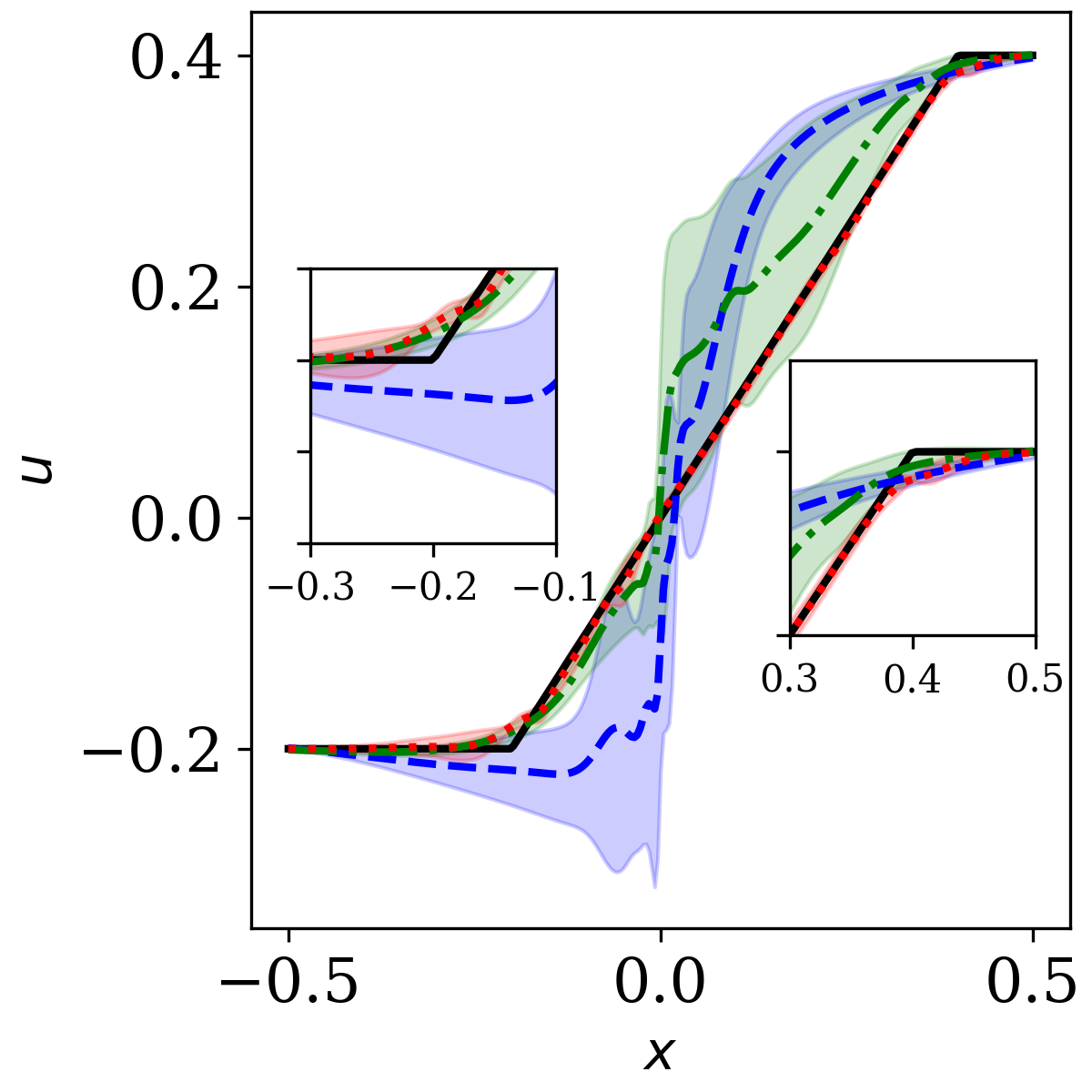}}\quad 
    \subfloat[]{\includegraphics[scale=0.45]{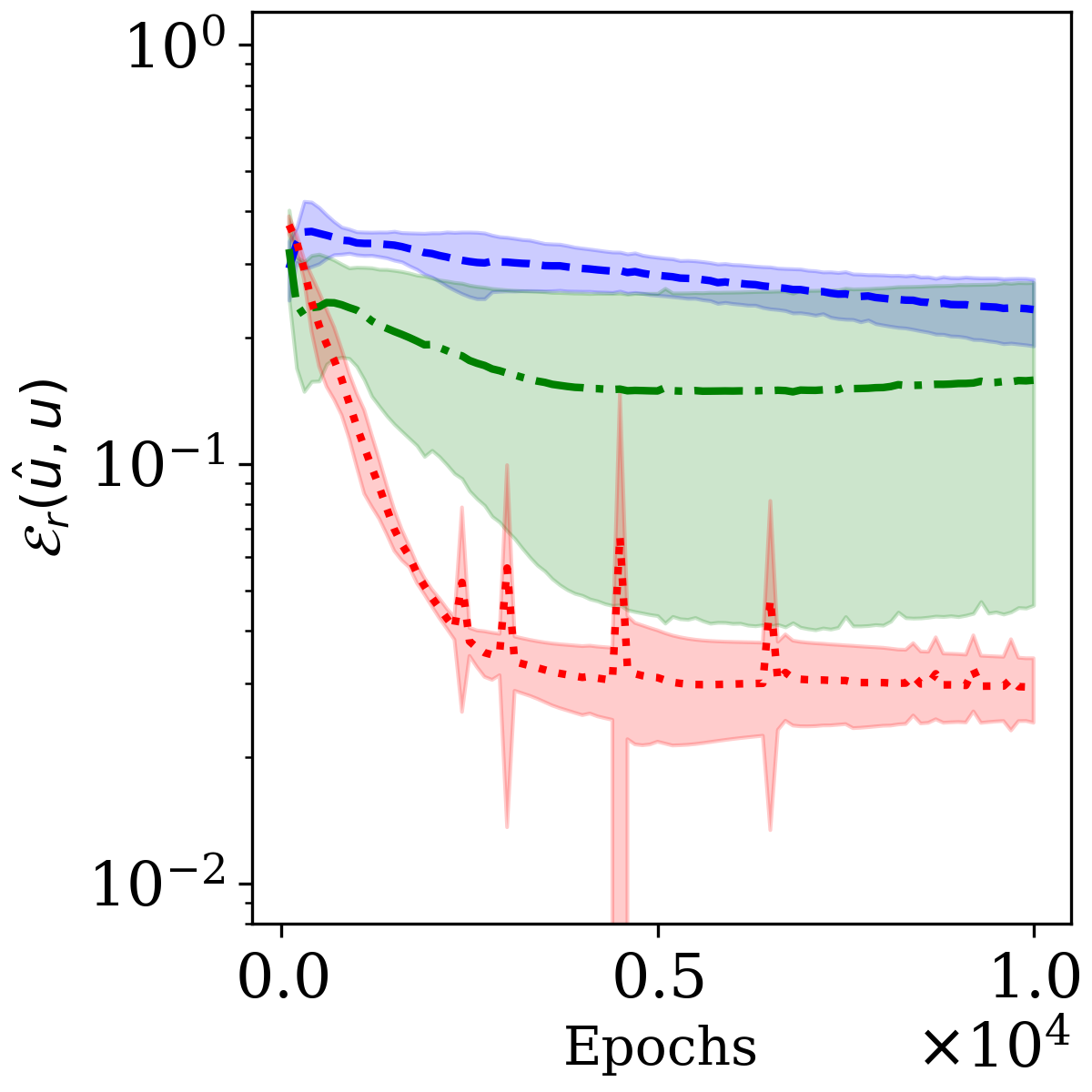}}\quad \\
    \subfloat[]{\includegraphics[scale=0.45]{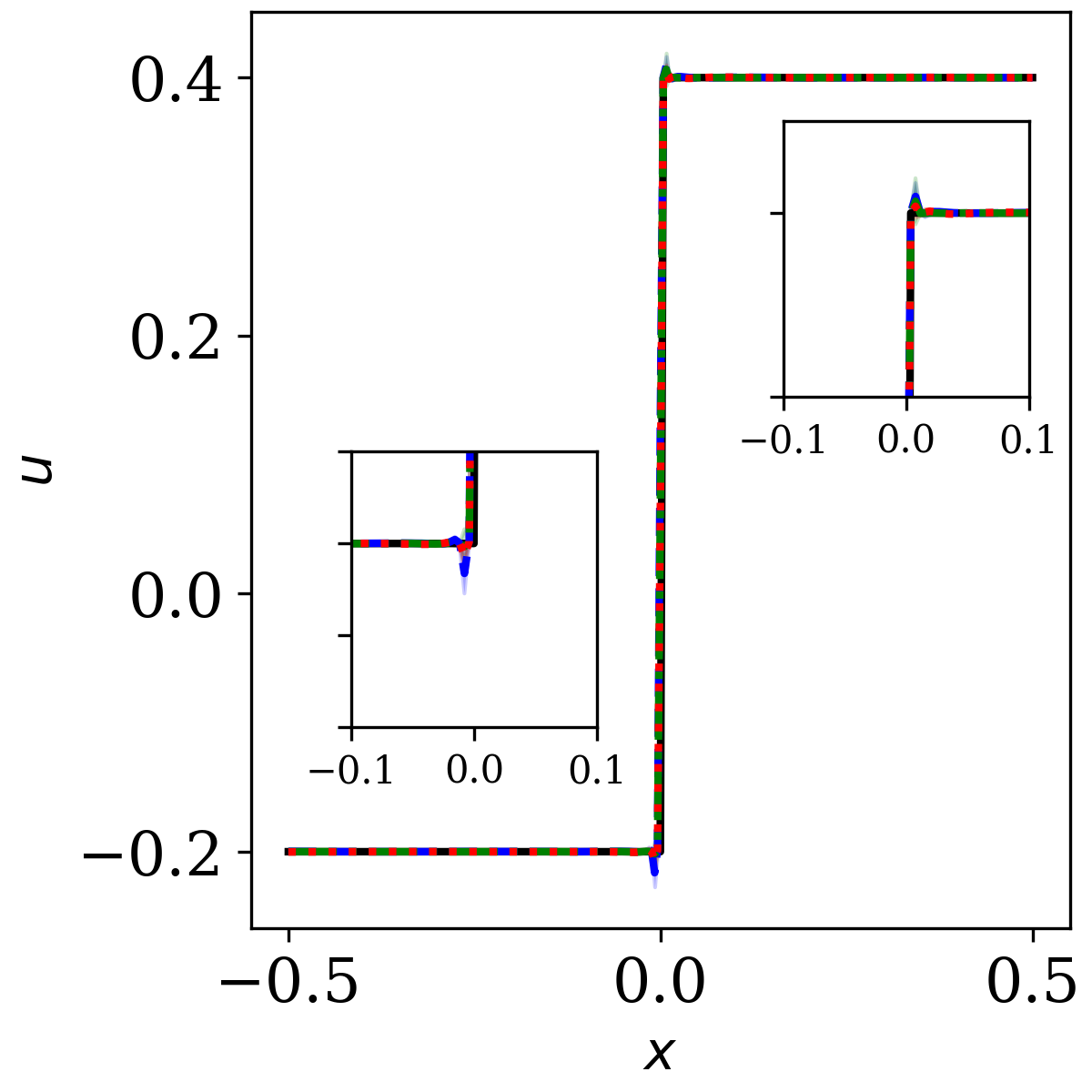}}\quad
    \subfloat[]{\includegraphics[scale=0.45]{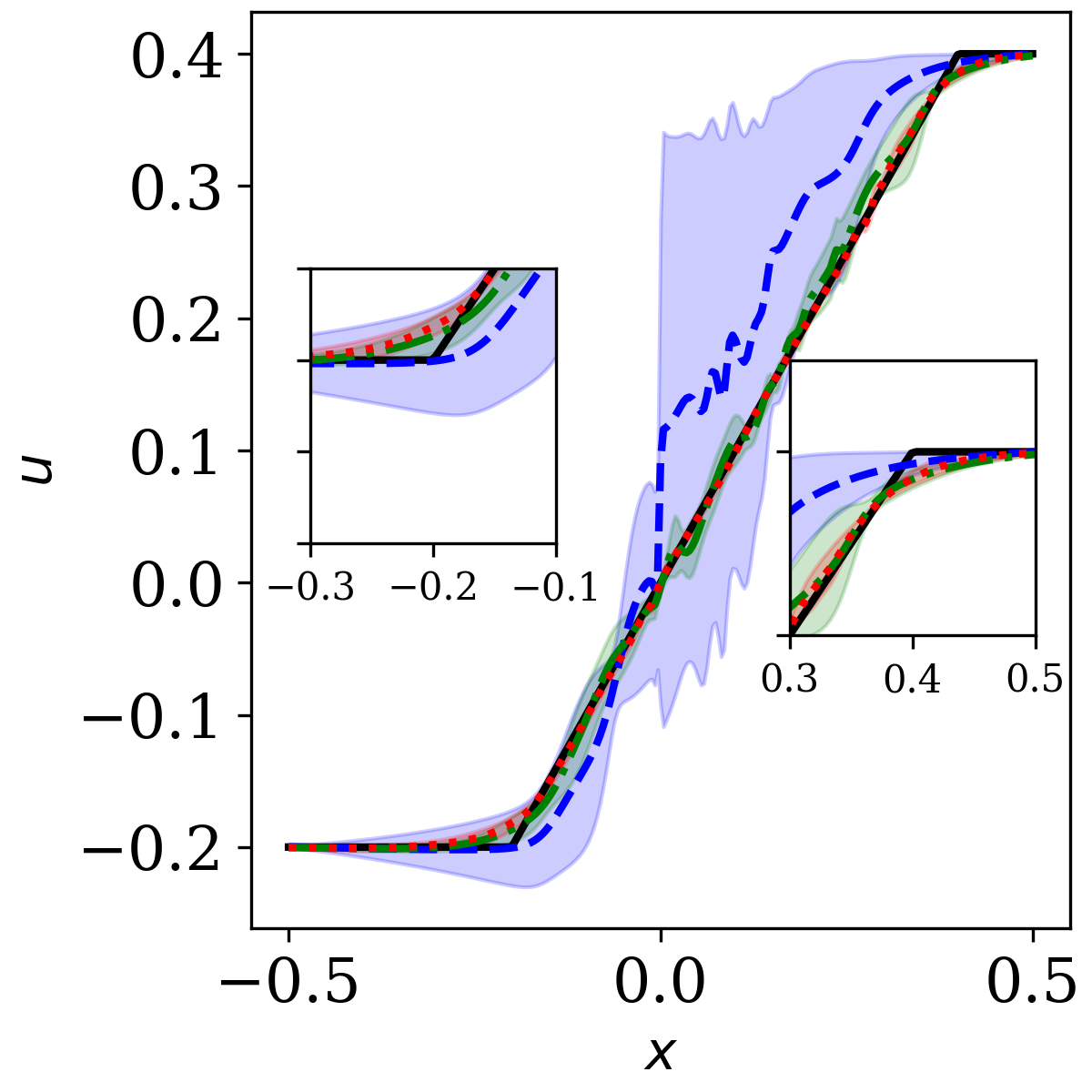}}\quad
    \subfloat[]{\includegraphics[scale=0.45]{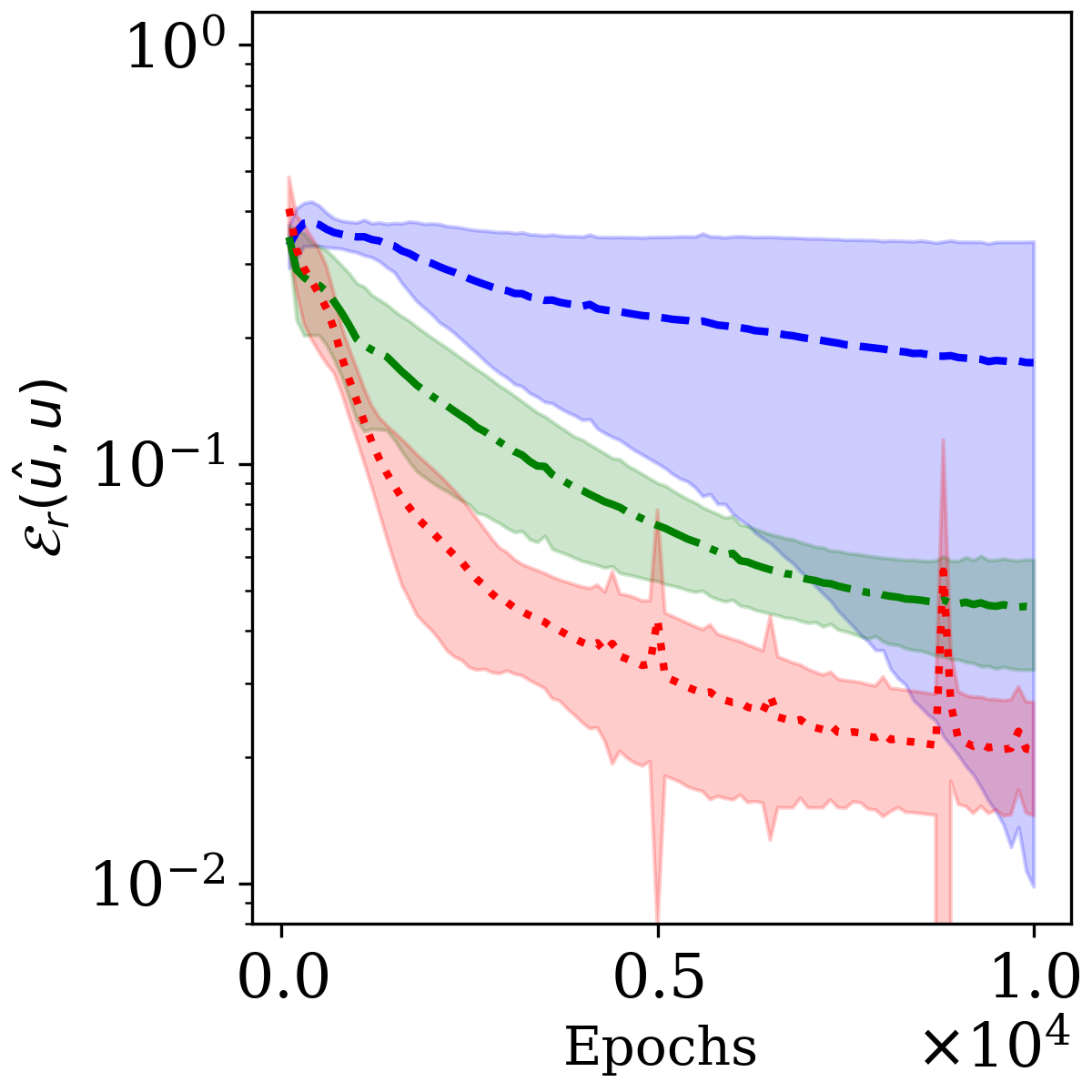}}\quad
    \caption{Transonic rarefaction: Panels (a)–(c) show coarse-mesh results; panels (d)–(f) show fine-mesh counterparts. Specifically: (a, d) predictions at $t=0$; (b, e) predictions at $t=1$; (c, f) evolution of relative $l^2$ error norms. Curves represent mean predictions; shaded areas indicate standard deviation bands.}
    \label{fig:invisid_burgers_profiles}
\end{figure}

Figure~\ref{fig:invisid_burgers_profiles} compares the predicted and exact solutions through their mean profiles and standard deviations at initial and final times with coarse and fine mesh, as well as the evolution of relative $\mathit{l^2}$ norms during training. 
All the predictions generally adhere to the prescribed initial data in panel (a), while minor overfitting is observed near the discontinuity, which mesh refinement can effectively mitigates as shown in panel (d).
In contrast, the final time results in panel (b) demonstrate that only the CAPU algorithm maintains excellent agreement with the exact solution, showing both high accuracy and consistency. The performance of MPU cannot be improved even with fine mesh in panel (e), due to the challenges in handling the transition point of rarefaction wave. Furthermore, panel (c) and (f) illustrate the superior convergence rate of CAPU in terms of the relative $\mathit{l^2}$-norm, despite occasional spikes in the progress.

\begin{table}[]
\centering
\footnotesize
\begin{tabular}{l|r|r}
\hline
Algorithm & {\centering Mesh $66\times33$ \quad} &    {\centering Mesh $130\times65$ \quad} \\ 
\hline
PECANN-MPU       & $2.00 \pm 1.27\times 10^{-1}$ & $1.66 \pm 1.23\times 10^{-1}$   \\
PECANN-CPU       & $2.61 \pm 1.25\times 10^{-1}$ & $9.89 \pm 4.24\times 10^{-2}$ \\
\textbf{PECANN-CAPU}       & $\boldsymbol{2.93 \pm 0.51\times 10^{-2}}$ & $\boldsymbol{1.93 \pm 0.52\times 10^{-2} }$ \\
\end{tabular}
\caption{Transonic Rarefaction: Relative $\mathit{l^2}$-norms over 5 trials using different penalty updating algorithms and two levels of uniform meshes.}
\label{tab:invisid_burgers_l2}
\end{table}

Table~\ref{tab:invisid_burgers_l2} summarizes the mean and standard deviation of the relative $\mathit{l^2}$-norms at the end of the training. 
The results demonstrate that CAPU outperforms both MPU and CPU across all metrics, exhibiting significantly lower mean errors and superior robustness (smaller standard deviations) on both coarse and fine meshes.
Mesh refinement generally enhances statistical performance, with additional residual points improving the capture of the residual loss particularly in high-gradient regions. 
Nevertheless, even with quadrupled sampling density, both the MPU and CPU consistently fail to achieve relative $\mathit{l^2}$ errors below $10^{-1}$.

\begin{figure}
    \centering
    \includegraphics[width=0.95\linewidth]{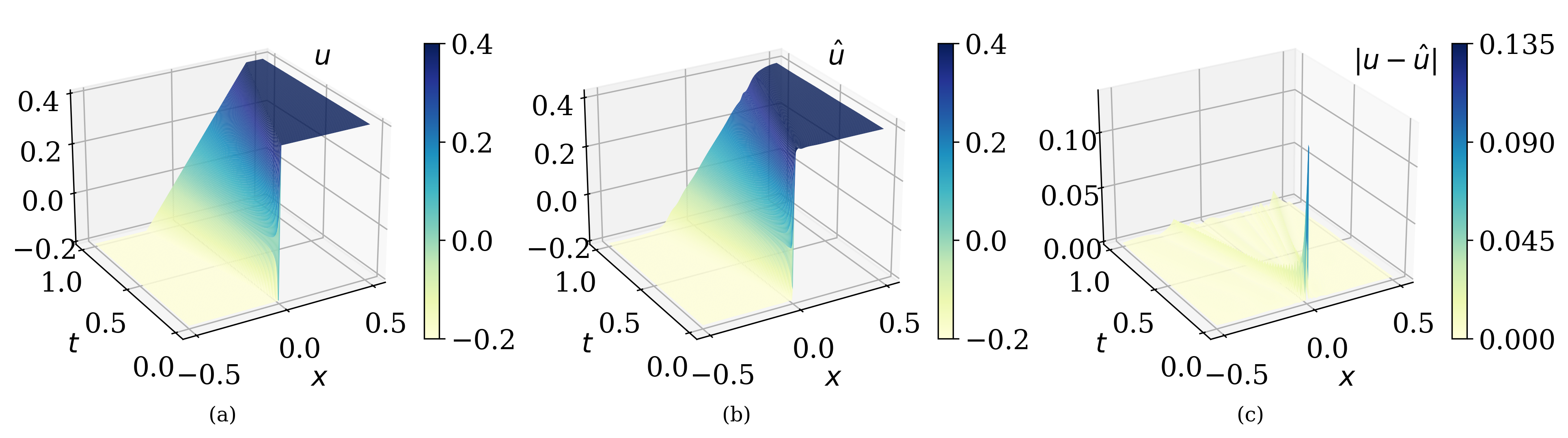}
    \caption{Transonic rarefaction: (a) exact solution, (b) prediction of the CAPU algorithm from the best fine-mesh trial, (c) the corresponding absolute point-wise error.}
    \label{fig:invisid_burgers_prediction}
\end{figure}

Fig.~\ref{fig:invisid_burgers_prediction} shows the contours of exact solution, the prediction and the absolute point-wise errors from the best trial of the CAPU algorithm with fine mesh. The main errors are concentrated around the discontinuity. This occurs because smooth function approximation can never replace the sharp discontinuity, leading to the maximum point-wise error keeping relatively large despite the overall low global errors. 


\begin{figure}[!h]
\centering
    \subfloat[]{\includegraphics[scale=0.45]{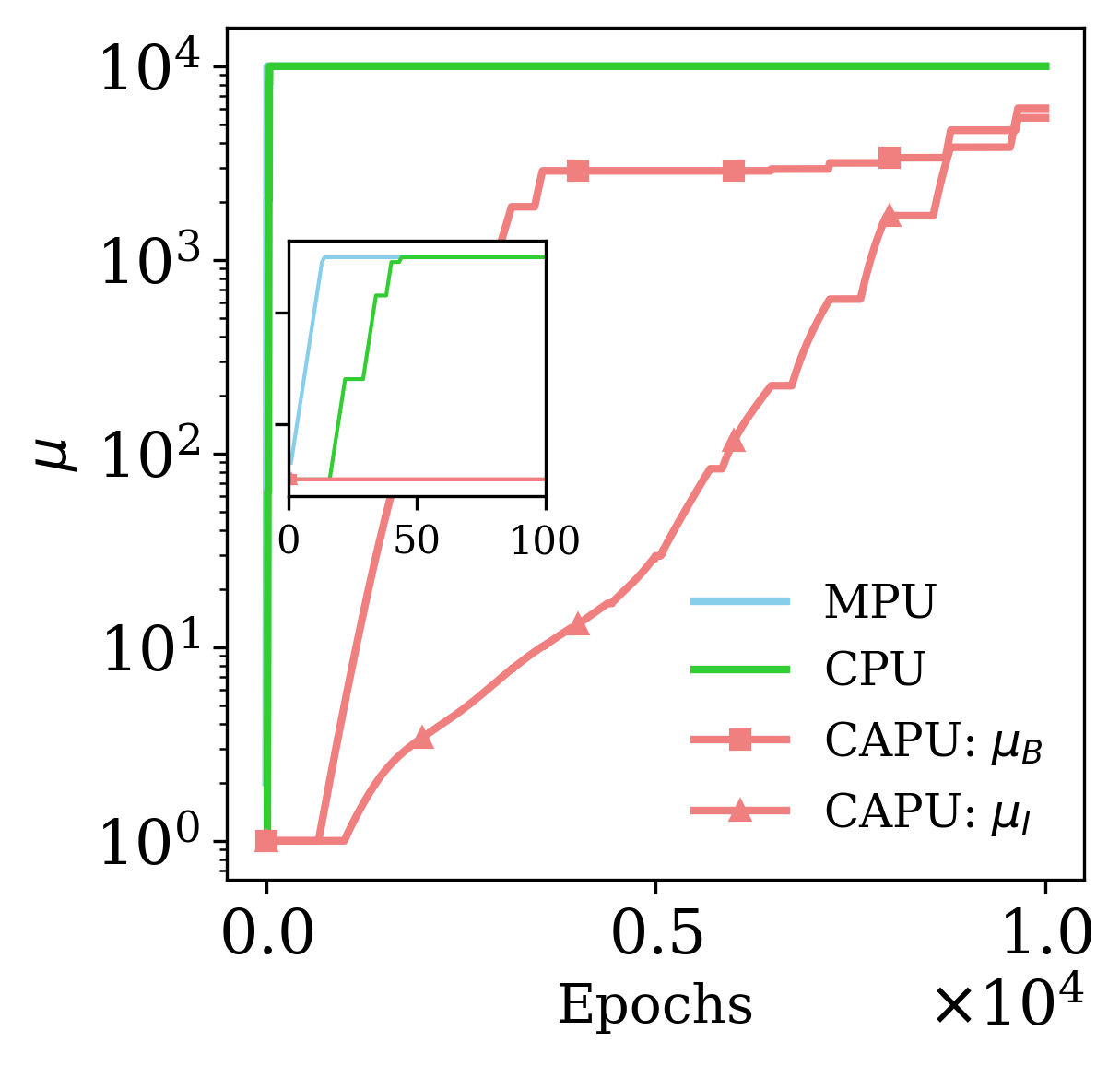}}\quad
    \subfloat[]{\includegraphics[scale=0.45]{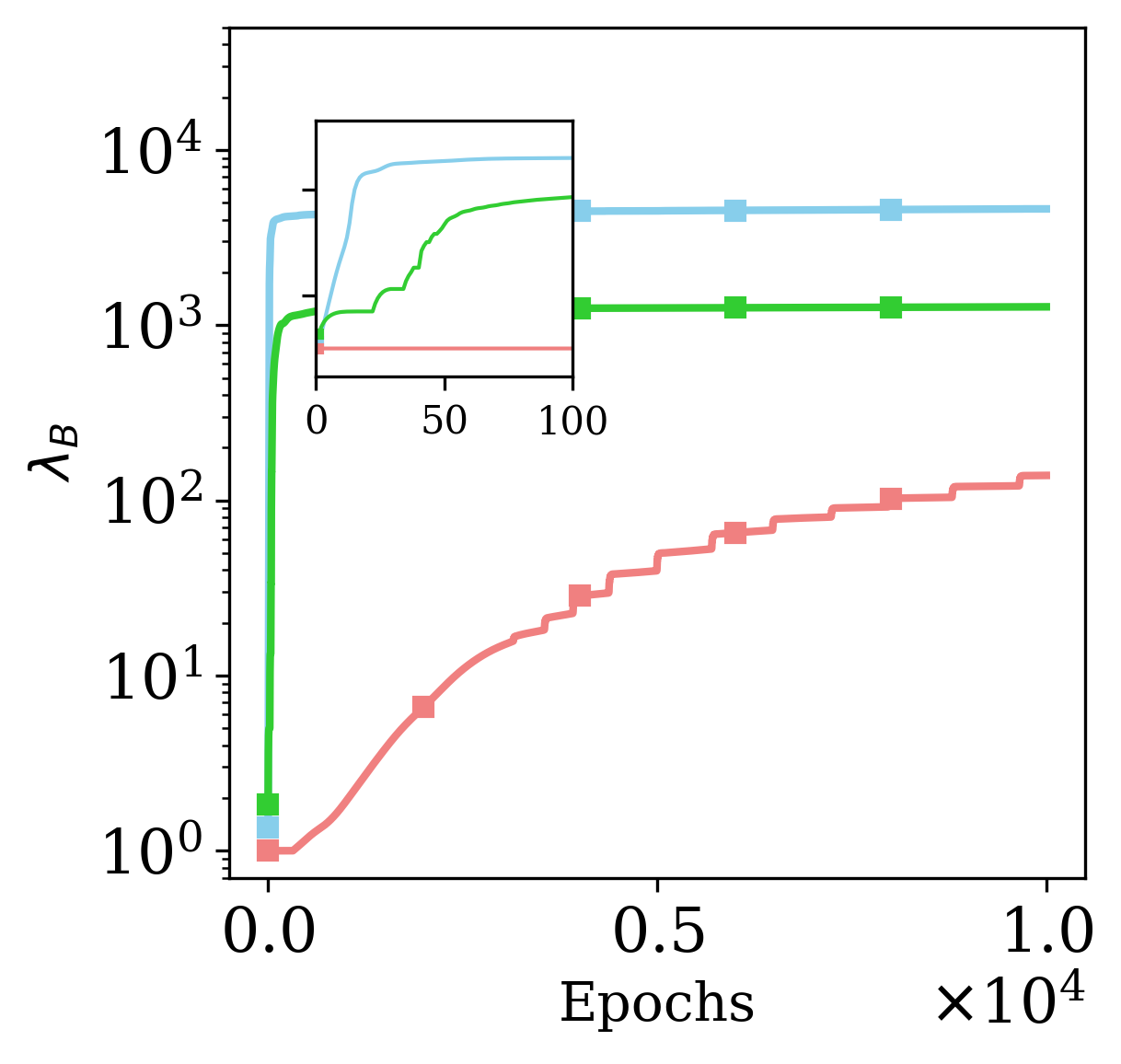}}\quad 
    \subfloat[]{\includegraphics[scale=0.45]{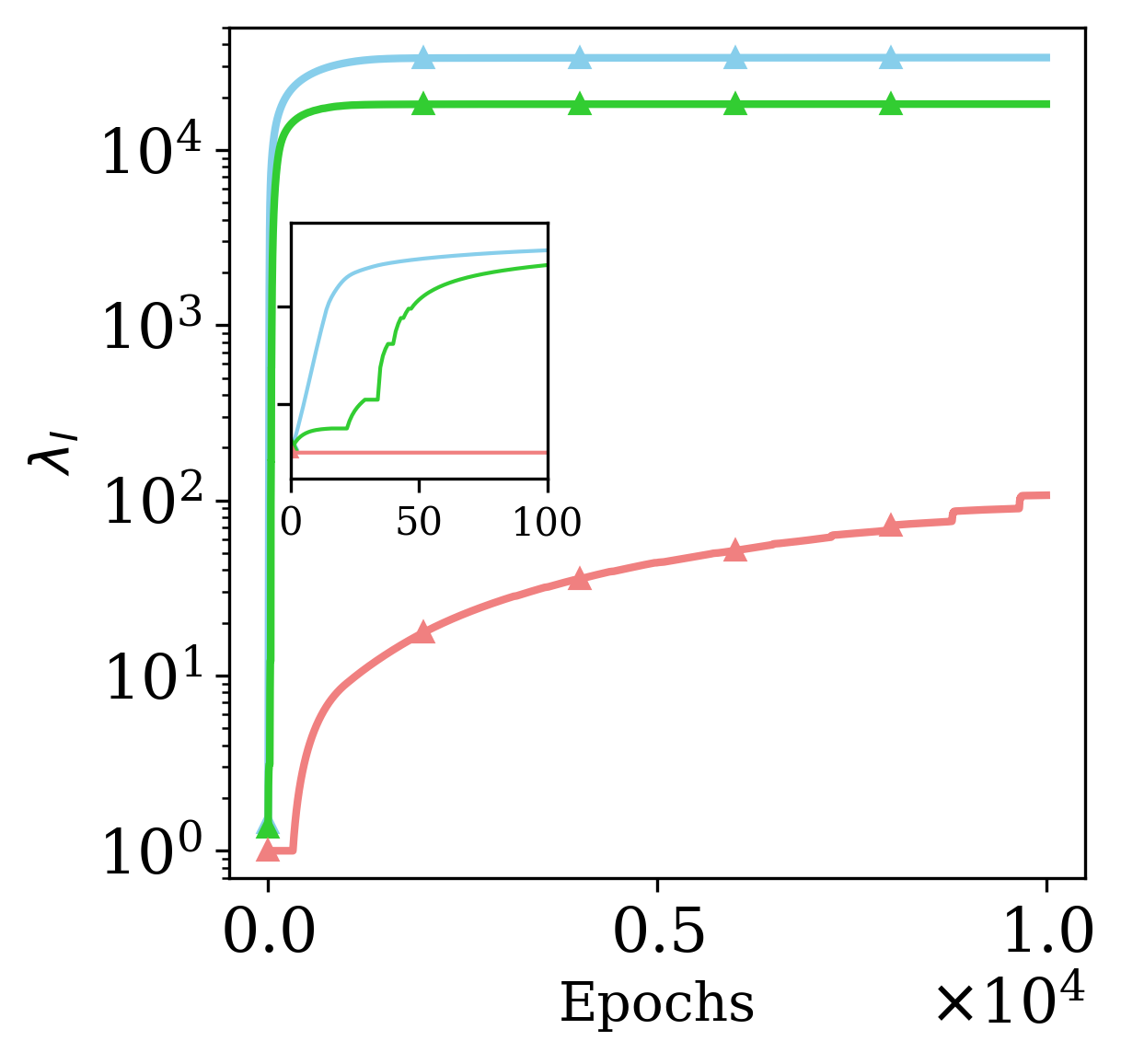}}\quad \\
    \subfloat[]{\includegraphics[scale=0.45]{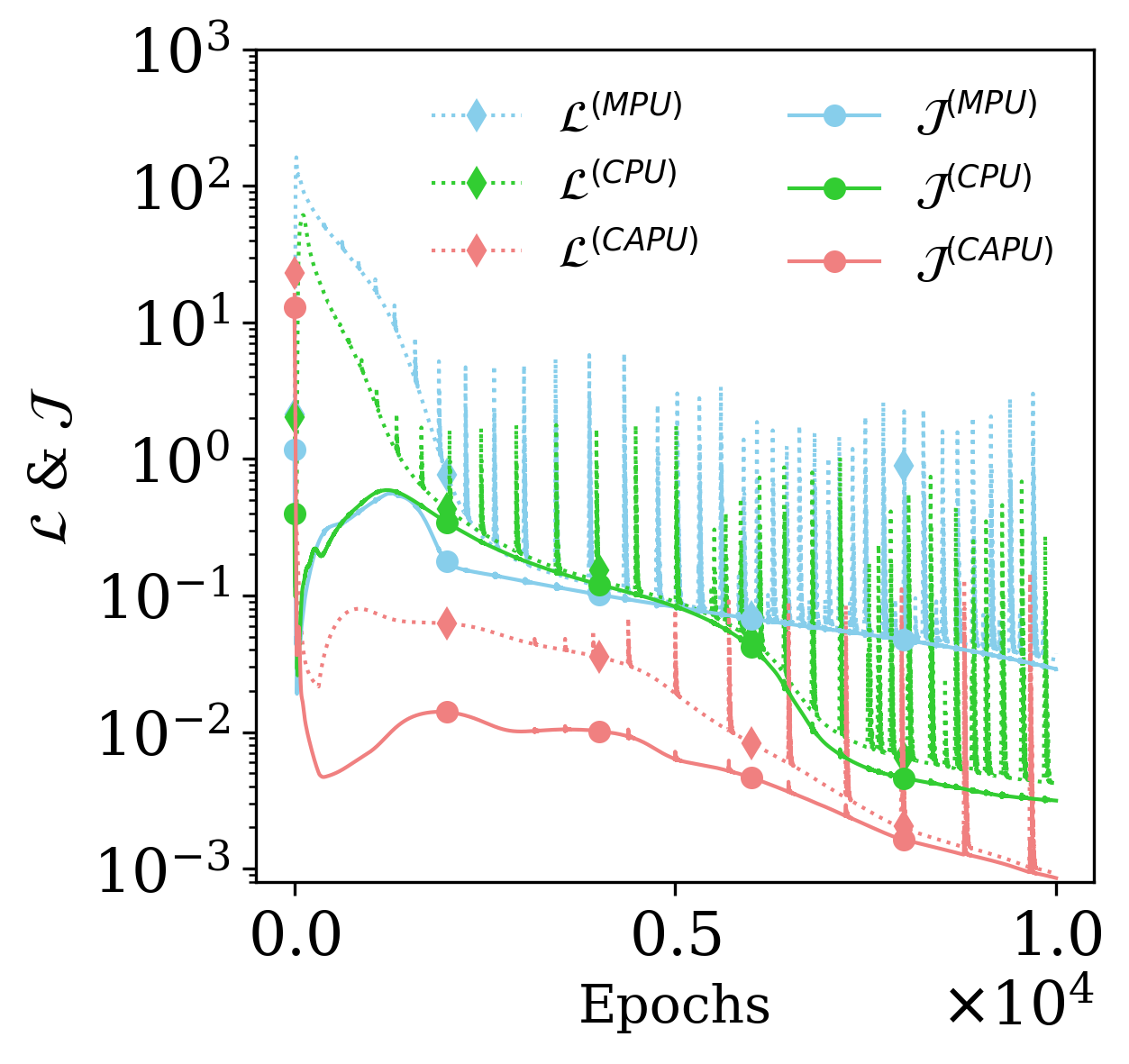}}\quad
    \subfloat[]{\includegraphics[scale=0.45]{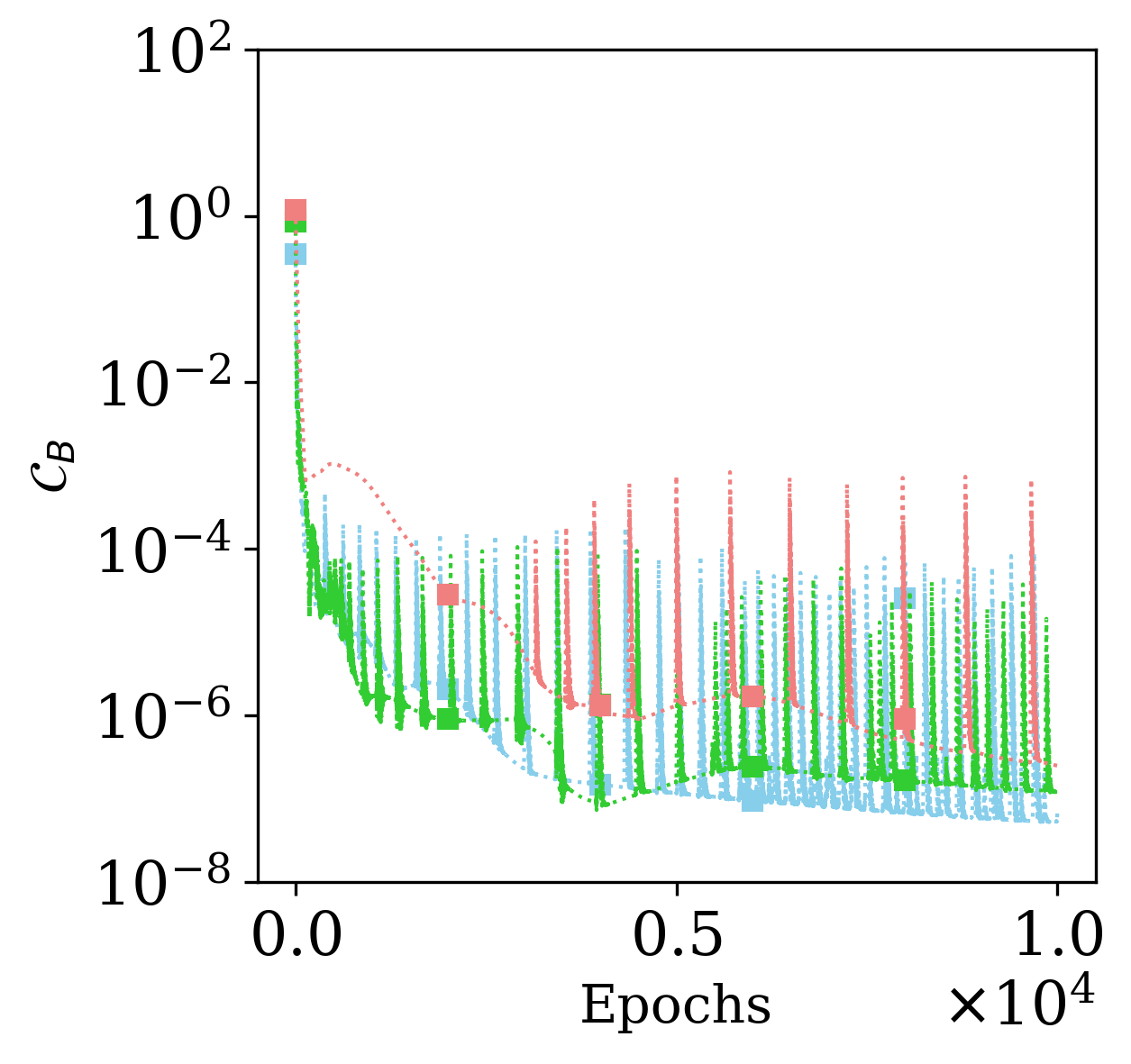}}\quad
    \subfloat[]{\includegraphics[scale=0.45]{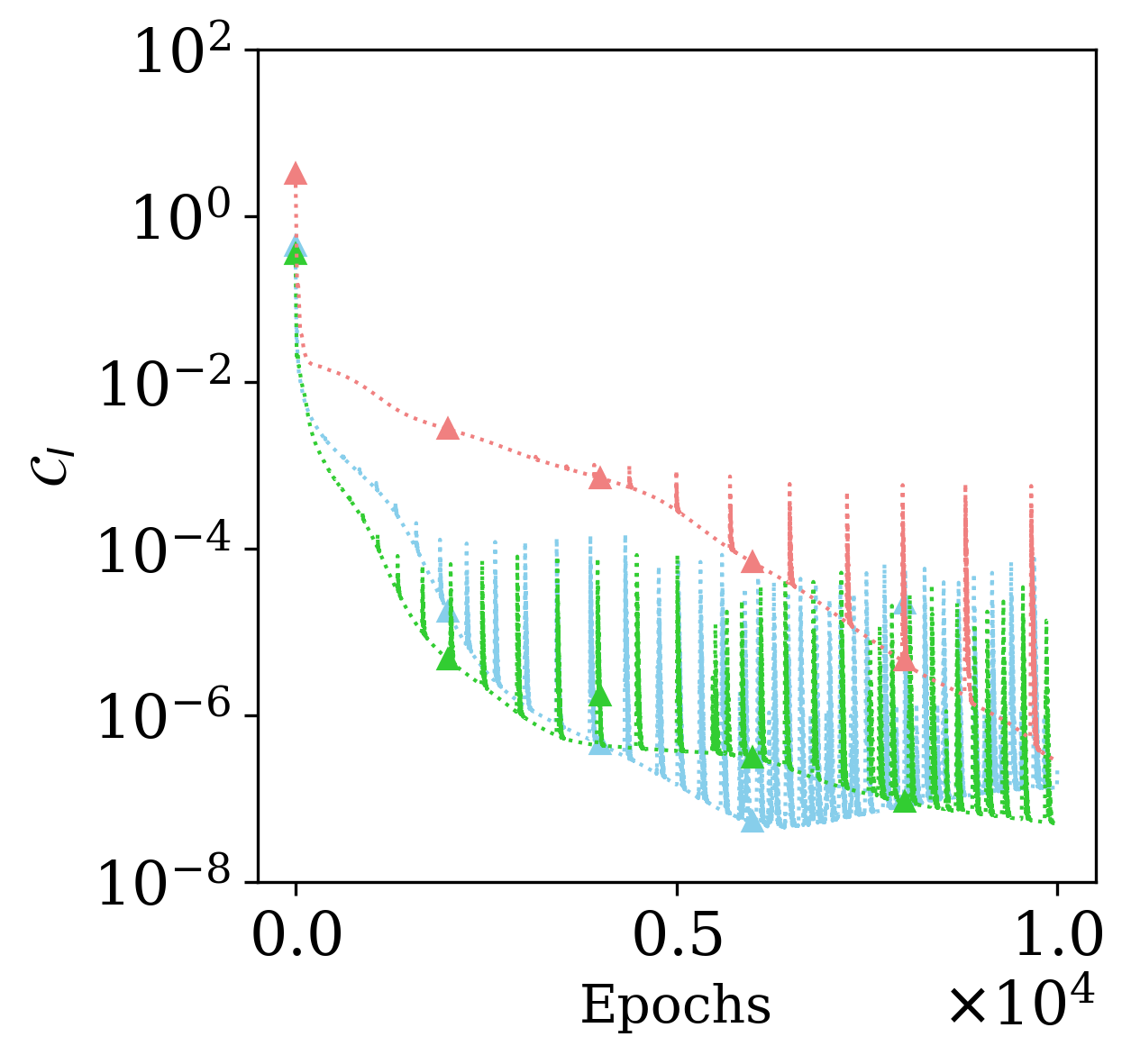}}\quad
    \caption{Transonic rarefaction problem: evolution of (a) penalty parameters, (b) Lagrange multipliers for boundary condition (BC), (c) Lagrange multipliers for initial condition (IC), (d) augmented losses and objectives, (e) BC constraints, (f) IC constraints from the worst trials of three strategies with fine mesh. }
    \label{fig:invisid_burgers_evolution}
\end{figure}

To better understand the inner-workings of handling discontinuities, Fig.~\ref{fig:invisid_burgers_evolution} compares the evolution of penalty parameters, Lagrange multipliers, and corresponding constraints, along with objective and augmented losses, using the worst-performing trials with fine mesh for each algorithm.
In panel (a), the penalty parameter of MPU grow exponentially and rapidly reach the maximum value $\mu_{\max} = 10^{4}$, while CPU reaches this limit more gradually by around epoch 50. 
In contrast, CAPU keeps the penalty parameters fixed at their initial values until the augmented loss satisfies a convergence criterion, after which each parameter increases independently. Notably, $\mu_{B}$ of CAPU consistently exceeds $\mu_{I}$ while remaining below MPU's prescribed upper bound.

Fig.~\ref{fig:invisid_burgers_evolution}(b) and (c) show the evolution of Lagrange multipliers $\lambda_{B}$ and $\lambda_{I}$. 
MPU demonstrates rapid growth ($\lambda_{B} > 10^3$, $\lambda_{I} > 10^4$ in 100 epochs) driven by its aggressive $\mu$ update, while CPU shows similar but dampened behavior.
In contrast, CAPU maintains initial $\lambda$ values before gradually converging toward the much lower values.
Based on the evaluation metrics, CAPU achieves the most accurate approximation of the near-optimal solution. This indicates that the Lagrange multipliers learned by CAPU are close to optimal, whereas the excessively rapid growth observed in MPU and CPU leads to deviations from the optimal values.
Panel (d) shows CAPU achieves lower objective (residual-averaged) and augmented losses than MPU/CPU, with a more stable ratio indicating that MPU/CPU's premature constraint over-prioritization leads to excessive Lagrange multipliers and optimization imbalance.
Augmented loss spikes correlate with sudden $\mathcal{C}_{B}$/$\mathcal{C}_{I}$ violations in panels (e) and (f).
The gradual reduction of $\mathcal{C}_{I}$ in the CAPU algorithm indicates that it progressively improves network expressiveness near discontinuities, unlike MPU and CPU, which prematurely overemphasize it.

\subsection{Forward problem: 1D Poisson's equation}\label{sec:1d_poisson}

\citet{wang_kolmogorovarnold-informed_2025} employed a one-dimensional Poisson's equation with a multi-scale solution—originally introduced by \citet{wang_eigenvector_2021}--as a regression task to illustrate the superior expressive capacity of KANs compared to standard MLPs. In support of their argument, they made the bold assertion that ``if function fitting fails, solving PDEs will certainly fail,'' and substantiated this claim by analyzing the convergence behavior of KANs and MLPs through the eigenvalue spectra of the Neural Tangent Kernel (NTK) matrix. For the same problem, \citet{wang_eigenvector_2021} demonstrated that PINNs, conventional or equipped with a single Fourier feature mapping, failed to accurately learn the solution. However, by employing two distinct Fourier feature mappings, they achieved a relative $l^2$ error norm of $1.36 \times 10^{-3}$.

To demonstrate the versatility of our PECANN-CAPU algorithm in solving multi-scale problems we apply our approach to the same 1D Poisson's equation considered on the domain $\Omega = { x \mid 0 \le x \le 1 }$.
\begin{equation}
\begin{aligned}
\nabla^2 u & = s, && \text{in} \quad \Omega, \\
u & = 0, && \text{on} \quad \partial \Omega.
\label{eq:1d_psn_eqn}
\end{aligned}
\end{equation}
An exact solution to Eq.\eqref{eq:1d_psn_eqn} is assumed in the following form:
\begin{equation}
u(x) = \sin(2\pi x) + 0.1 \sin(b \pi x), \quad \forall x \in \Omega,
\label{eq:1d_psn_sol}
\end{equation}
where $b$ is a user-defined wavenumber. The corresponding source function $s(x)$ is computed by substituting this solution into Eq.\eqref{eq:1d_psn_eqn}. \citet{wang_eigenvector_2021} set $b=50$ to highlight the severe limitations of conventional PINNs, as well as PINNs with a single Fourier feature mappings, in learning the correct solution. 


Following \citet{wang_kolmogorovarnold-informed_2025}, we adopt the same feed-forward neural network with four hidden layers of 100 neurons each. The model is trained for 100,000 epochs using the Adam optimizer and the same \textit{ReduceLROnPlateau} scheduler described in Section \ref{sec:composite_heat}. To avoid bias from point placement, a fixed set of uniformly distributed residual points and two boundary points are used as the batch during training.
\begin{table}[h!]
    \centering
    \footnotesize
    \begin{tabular}{c|c|r|r}
        \hline
        \multirow{2}{*}{$b$} & \multirow{2}{*}{Batch} & \multicolumn{2}{c}{Relative $l^2$} \\ \cline{3-4}
         &   & mean $\pm$ std & best \\
         \hline
         8 & 128 & $8.49\pm 6.27\times10^{-6}$ & $8.37\times10^{-7}$ \\ 
        16 & 128 & $1.51\pm 1.06\times10^{-5}$ & $5.13\times10^{-6}$ \\ 
        32 & 128 & $8.36\pm 2.72\times10^{-4}$ & $5.17\times10^{-4}$ \\ 
        \hline
        50 & 128 & $1.37\pm 2.15\times10^{0}$ & $9.15\times10^{-3}$ \\ 
        50 & 256 & $2.04\pm 3.96\times10^{-1}$ & $1.07\times10^{-3}$ \\
        50 & 512 & \boldsymbol{$3.69\pm 4.78\times10^{-2}$} & \boldsymbol{$1.08\times10^{-3}$} \\
    \end{tabular}
    \caption{Poisson's equation: Relative $l^2$ norms for different wavenumbers $b$ in the exact solution (Eq. \ref{eq:1d_psn_sol}) and varying batch (i.e. total number of collocation points) sizes, using PECANN-CAPU framework with standard MLPs.}
    \label{tab:1d_psn_l2}
\end{table}

Before introducing Fourier feature mappings, we first assess our PECANN-CAPU method with standard MLPs across a range of wavenumbers $b$ (8–50) and batch sizes (128–512). Table~\ref{tab:1d_psn_l2} presents the relative $l^2$ norm $\mathcal{E}_r$ of the predictions. As the solution complexity grows with $b$, the mean and standard deviation of $\mathcal{E}_r$ increase--rising from $10^{-6}$ to $10^0$ at a fixed batch size (i.e. number of collocation points) of 128. Despite this, the most challenging case ($b = 50$) still yields a low error of $\mathcal{E}_r = 9.153 \times 10^{-3}$ in the best trial. To better capture high-frequency features at $b = 50$, we test larger batch sizes (256 and 512), observing improved performance: the mean error drops to $3.690 \times 10^{-2}$, with the best trial converging to the $10^{-3}$ level.

\begin{figure}[!h]
\centering
    \subfloat[]{\includegraphics[scale=0.45]{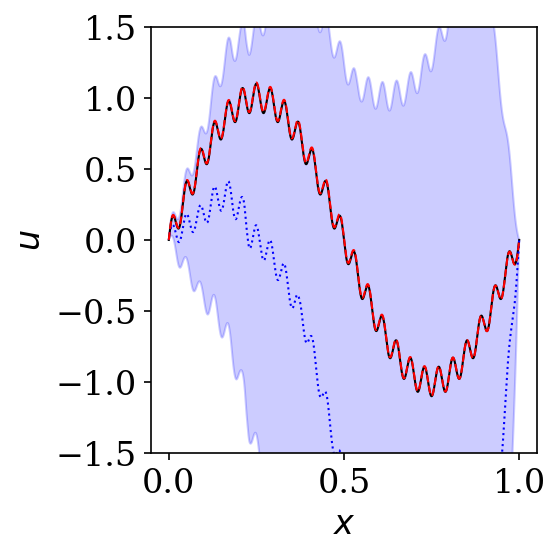}}\quad
    \subfloat[]{\includegraphics[scale=0.45]{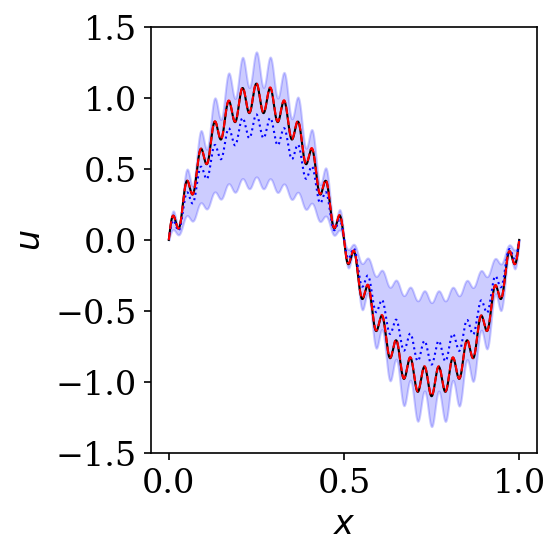}}\quad
    \subfloat[]{\includegraphics[scale=0.45]{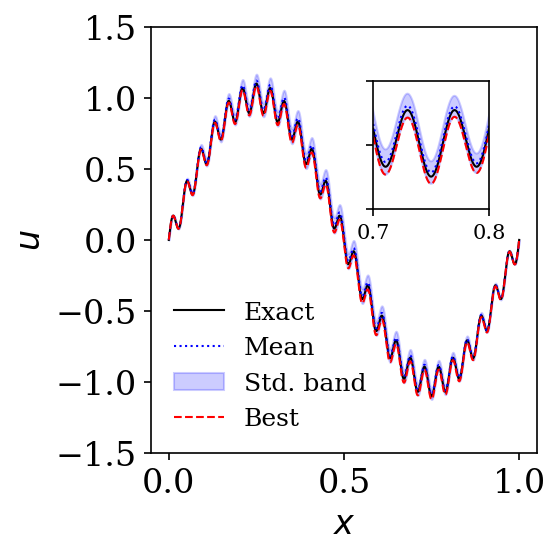}}\quad
    \caption{Poisson’s equation ($b = 50$ in Eq. \ref{eq:1d_psn_sol}): Mean and standard deviation of the predictions, along with the best-performing trial, for batch sizes of (a) 128, (b) 256, and (c) 512 using the PECANN-CAPU framework with standard MLPs.}
    \label{fig:1d_psn_pred}
\end{figure}

Figure~\ref{fig:1d_psn_pred} presents the mean and standard deviation of predictions, along with the best-performing trial, for different batch sizes using the CAPU algorithm. As the batch size increases from Fig.~\ref{fig:1d_psn_pred}(a) to (c), the mean predictions better align with the exact solution, and the uncertainty bands shrink. The inset in Fig.~\ref{fig:1d_psn_pred}(c) confirms this trend, though minor discrepancies remain. In all cases, the best-performing trials closely match the exact solution.

\begin{figure}[!h]
\centering
    \subfloat[]{\includegraphics[scale=0.45]{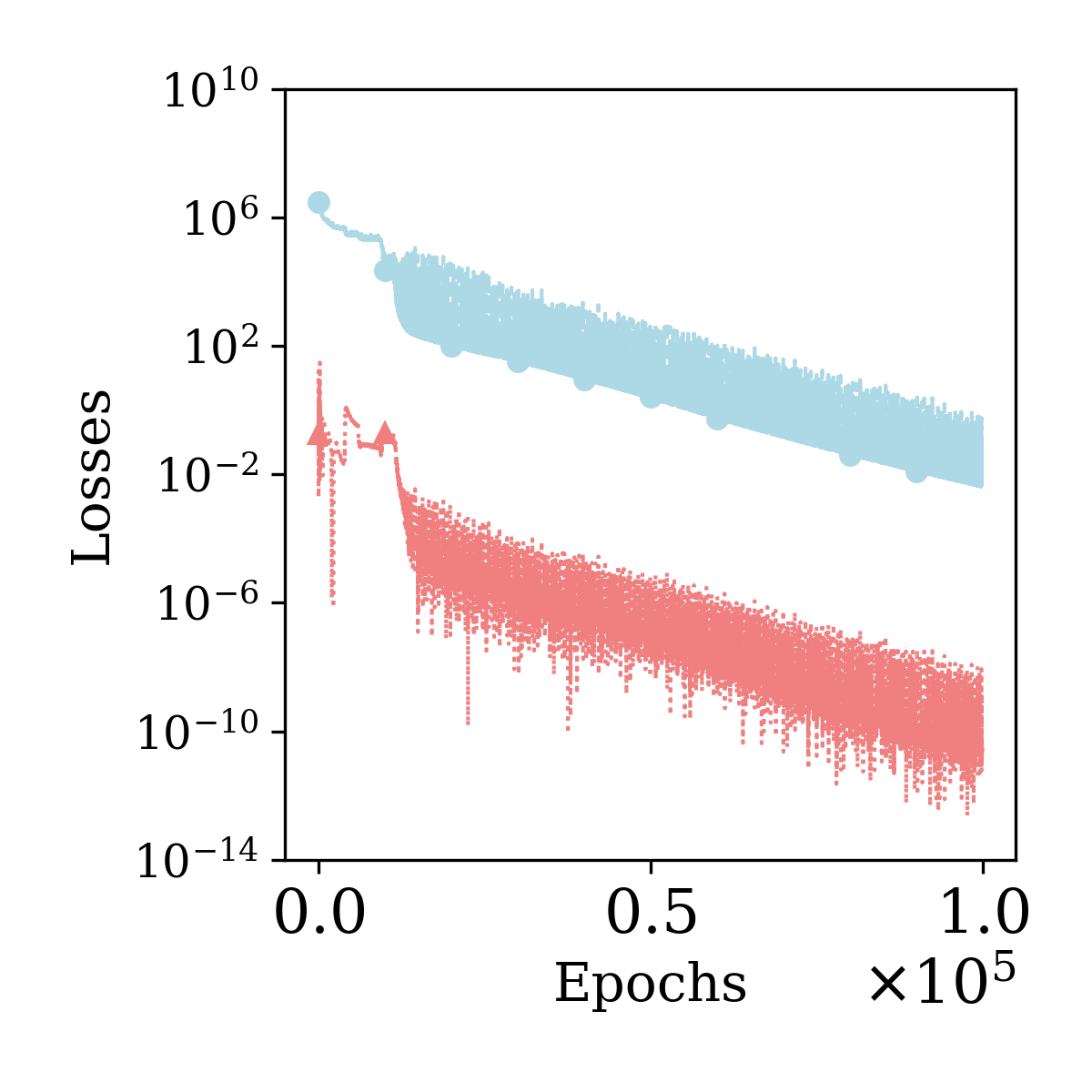}}\quad
    \subfloat[]{\includegraphics[scale=0.45]{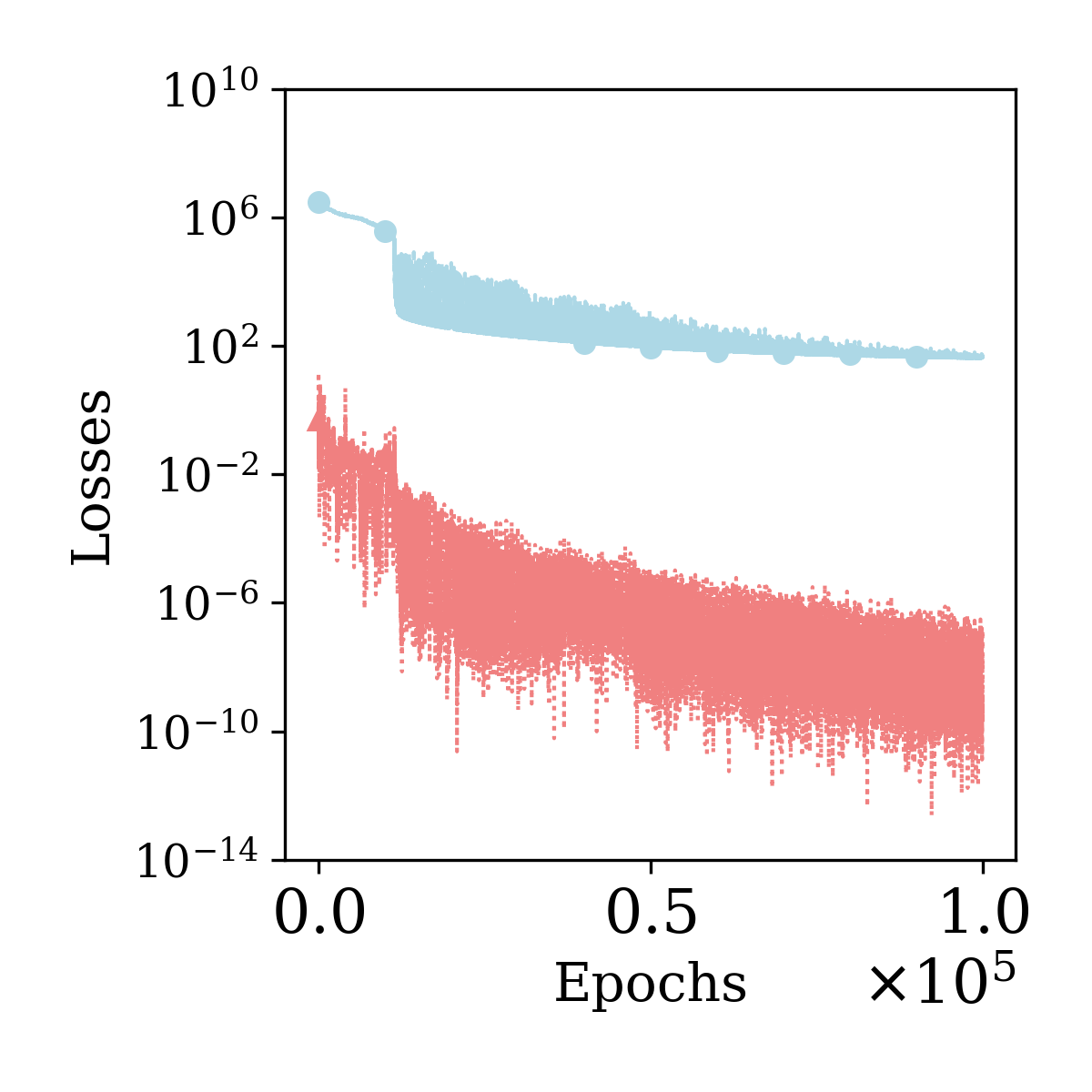}}\quad
    \subfloat[]{\includegraphics[scale=0.45]{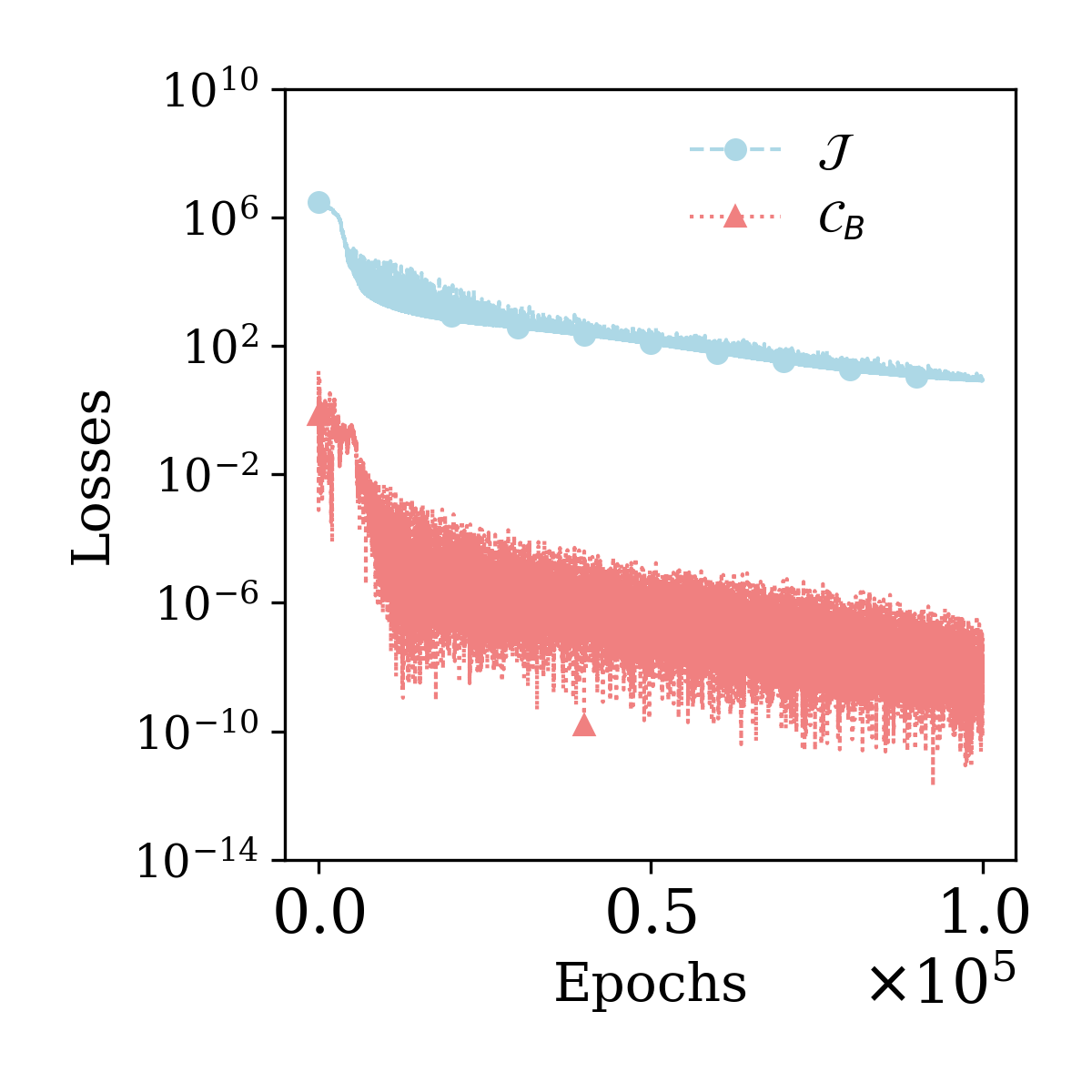}}\quad
    \caption{Poisson’s equation ($b = 50$ in Eq. \ref{eq:1d_psn_sol}): Evolution of the objective and boundary constraint of the best-performing trial for batch sizes of (a) 128, (b) 256, and (c) 512 using the CAPU algorithm.}
    \label{fig:1d_psn_evol_loss}
\end{figure}

Figure~\ref{fig:1d_psn_evol_loss} presents the evolution of the objective and boundary constraint for the best-performing trials. In all cases, the boundary constraint remains tightly satisfied around $10^{-10}$. However, with a batch size of 128 (panel a), the objective exhibits strong oscillations, unlike the more stable trends in panels (b) and (c). Despite reaching a lower final value (around $10^{-2}$), the case in (a) yields the highest relative $l^2$ error, suggesting overfitting due to insufficient residual sampling. The observed oscillations likely stem from unresolved high-frequency components, and the decaying learning rate fails to stabilize training. These results underscore the importance of adequate residual sampling for capturing complex solution features.

Beyond the observed overfitting, our results demonstrate that while MLPs are commonly perceived as limited in their capacity to model multi-scale solutions, they can nonetheless effectively solve this forward problem when employed within the PECANN framework and optimized using the CAPU algorithm. We therefore argue that it is premature to dismiss MLPs as inherently incapable of representing multi-scale phenomena. Crucially, when evaluating the expressive capacity of a neural network in physics-informed learning of PDEs, the formulation of the optimization problem is as important as the choice of network architecture.

\subsubsection{Enhancements with a single Fourier feature mappings}
We adopt the same wide and shallow neural network as in \citet{wang_eigenvector_2021}, with two hidden layers of 100 units each. The network is trained with the Adam optimizer for 40,000 epochs, using an exponential learning rate decay (factor 0.9 every 1000 iterations). Residual points are resampled randomly at each epoch. Instead of multiple, tuned Fourier feature mappings as adopted in \cite{wang_eigenvector_2021}, we apply the original method from \citet{tancik_fourier_2020}, replacing the first hidden layer with a single Fourier feature mapping matrix $\mathbf{B} \in \mathbb{R}^{1 \times 50}$ only. We note that \citet{wang_eigenvector_2021} did not obtain good predictions with a single layer of Fourier feature mappings.

\begin{figure}[!h]
\centering
    \includegraphics[scale=0.6]{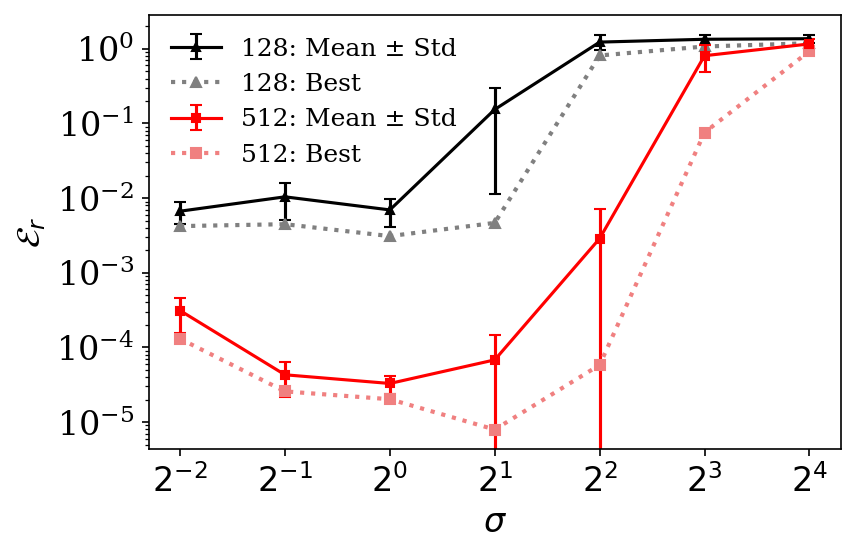}
    \caption{Variation in the statistical performance of relative \(l^{2}\) norms for Poisson's equation (\(b=50\) in Eq. \ref{eq:1d_psn_sol}), shown as a function of Gaussian deviation \(\sigma \in [0.25,16]\). A single Fourier feature mapping is used within the PECANN-CAPU framework.}
    \label{fig:1d_psn_fourier_sigma_l2}
\end{figure}

Figure~\ref{fig:1d_psn_fourier_sigma_l2} summarizes the mean and standard deviation of the relative $l^2$ error ($\mathcal{E}_r$) over 10 trials for varying Gaussian deviations $\sigma$ and two batch sizes (128 and 512). Best-performing trial results are also shown. The mean $\mathcal{E}_r$ remains stable below $10^{-2}$ for $\sigma$ values between 0.25 and 1.
As $\sigma$ increases beyond 1, the mean $\mathcal{E}_r$ rises, indicating reduced prediction accuracy.
In contrast, larger batch sizes generally yield lower errors, particularly for smaller $\sigma$.
Notably, $\sigma = 1$ provides the best overall performance and consistency; increasing the batch size to 512 further reduces the mean error to the order of $10^{-5}$.
These findings indicate that a single Fourier feature mapping with $\sigma = 1$ is sufficient to capture both low- and high-frequency components in multi-scale solutions when using PECANN-CAPU algorithm. This contrasts with prior findings in \citet{wang_eigenvector_2021}, which suggests the need for multiple Fourier features with tuning different $\sigma$ values to separately address different frequency components.

\begin{figure}[!h]
\centering
    \subfloat[]{\includegraphics[width=0.48\textwidth]{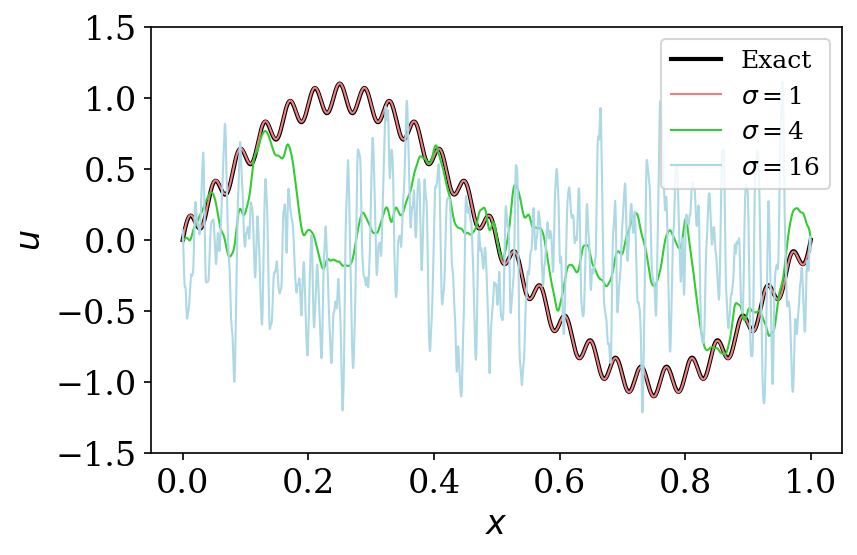}}\quad 
    \subfloat[]{\includegraphics[width=0.48\textwidth]{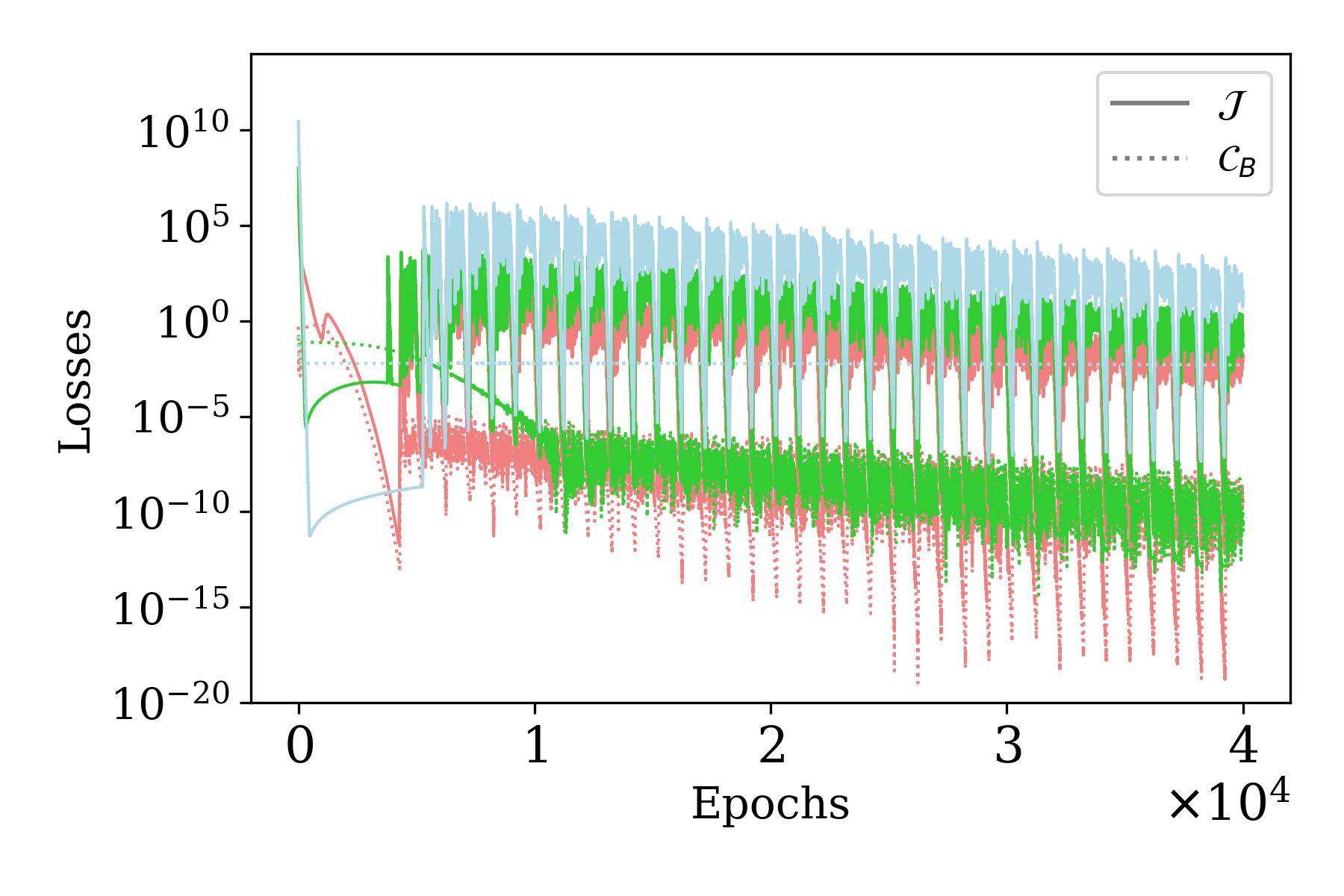}}
    \caption{Poisson’s equation ($b = 50$ in Eq. \ref{eq:1d_psn_sol}): (a) Prediction from the best-performing trials for $\sigma = 1$, 4, and 16 with a batch size of 128; (b) corresponding evolution of the objective (solid lines) and boundary constraint (dotted lines). A single Fourier feature mapping is used within the PECANN-CAPU framework.}
    \label{fig:1d_psn_fourier_evol_loss}
\end{figure}

An important observation is that the performance of the CAPU algorithm is sensitive to the choice of Gaussian deviation $\sigma$, with larger values leading to significant degradation in accuracy. To investigate this further, we examine $\sigma = 1$, $4$, and $16$, and present the best-performing predictions along with the evolution of loss terms in Figure~\ref{fig:1d_psn_fourier_evol_loss}. As shown in panel (a), only $\sigma = 1$ produces smooth and accurate predictions, while larger values—especially $\sigma = 16$—lead to discontinuities and wedge-like artifacts.
Panel (b) shows that the objective (averaged residual loss) initially drops sharply in all cases and oscillates around $10^{-2}$—a lower level than in standard networks trained with CAPU—indicating that Fourier features enhance expressiveness. However, this increased capacity also raises the risk of overfitting, particularly for large $\sigma$, as reflected in poor predictive performance. This also explains the improved accuracy at batch size 512, where more residual points help counteract overfitting.
Notably, for $\sigma = 16$, the boundary constraint $\mathcal{C}_B$ fails to decrease, despite continued growth of the Lagrange multiplier $\lambda_B$ under CAPU. While increasing the penalty scaling factor $\eta_B$ might accelerate $\lambda_B$ and better enforce the constraint, we recommend simply using the standard Gaussian deviation $\sigma = 1$ when applying CAPU with a single Fourier feature mapping, as it yields accurate, consistent predictions and faster convergence.

\begin{table}[h!]
    \centering
    \footnotesize
    \begin{tabular}{c|r|r}
        \hline
        \multirow{2}{*}{$b$} &  \multicolumn{2}{c}{Relative $l^2$} \\ \cline{2-3}
         & mean $\pm$ std & best \\
         \hline
            50 & $3.29\pm 0.88\times10^{-5}$ & $2.02\times10^{-5}$  \\ 
            64 & $5.85\pm 3.23\times10^{-5}$ & $2.86\times10^{-5}$ \\ 
            \boldsymbol{$128$} & \boldsymbol{$4.36\pm 3.14\times10^{-3}$} & \boldsymbol{$6.50\times10^{-4}$} \\ 
    \end{tabular}
    \caption{Poisson's equation: relative $l^2$ errors for different wavenumbers $b$ of the exact solution (Eq. \ref{eq:1d_psn_sol}) using a single Fourier feature mapping with standard normal distribution $\sigma = 1$. Results are obtained using the CAPU algorithm with the Adam optimizer over $4\times10^4$ epochs and a batch size of 512.}

    \label{tab:1d_psn_l2_fourier}
\end{table}


To examine the limitations of the CAPU algorithm with a single Fourier feature mapping ($\sigma = 1$), we increase the wavenumber of the exact solution up to $b = 128$. To mitigate overfitting, 512 residual points are randomly sampled at each epoch. Table~\ref{tab:1d_psn_l2_fourier} reports the relative $l^2$ error statistics. Although performance slightly declines with increasing $b$, the method still achieves accurate and consistent results even at $b = 128$, demonstrating its effectiveness on highly multiscale problems.


The 1D Poisson's equation with a multi-scale solution example show that, given a sufficiently large number of collocation points, standard MLPs are capable of learning the solutions despite the claims made in \citet{wang_kolmogorovarnold-informed_2025}. Moreover, when the number of collocation points is reduced, the incorporation of a single Fourier feature mapping proves sufficient to achieve excellent accuracy. Overall, the performance of the PECANN-CAPU framework substantially improves with a $\sigma = 1$ Fourier feature mapping, enabling accurate resolution of both low- and high-frequency components in multiscale problems when sufficient residual points are used. 

Our empirical observations underscore the critical importance of the underlying constrained optimization formulation and its associated algorithms in physics-informed learning, particularly for accurately capturing multi-scale and highly oscillatory solutions. We note that our conclusion does not diminish the relevance of advances in neural network architectures; rather, it highlights the complementary importance of optimization strategies in achieving robust and efficient learning.

\subsection{Forward problem: 2D Helmholtz equation with high-frequency and multi-scale solutions}\label{sec:2d_helm}
The solution of the Helmholtz equation, across varying levels of complexity, has been extensively utilized within the physics-informed machine learning community as a benchmark to evaluate the effectiveness of different methodologies, including APINNs \cite{hu_augmented_2023}, FBPINNs \cite{DOLEAN2024}, and PIKANs \cite{SHUKLA2024117290}. 

The 2D Helmholtz equation in the domain $\Omega = \{(x, y) \mid -1 \leq x, y \leq 1\}$ is defined as:
\begin{equation}
    \begin{aligned}
        \nabla^2 u + k^2 u & = s, && \text{in} \quad \Omega, \\
                         u & = g, && \text{on} \quad \partial \Omega,
         \label{eq:helmholtz_eqn}
    \end{aligned}
\end{equation}
where $k$ is the Helmholtz wavenumber.

Here, we will consider two categories of problems. The first adopts the same setup used to assess cPIKANs \cite{SHUKLA2024117290}. This problem is first employed to better understand the sensitivity of the predicted solution to the hyperparameters of the CAPU algorithm (see~\ref{sec:appendixb}), and to highlight that algorithmic innovation plays a role as crucial as architectural design. The second problem is much more challenging than the first one. It involves a localized Gaussian-like source term with a highly oscillatory solution. This case is used to further assess the PECANN-CAPU approach augmented with a single Fourier feature mapping. 

The exact solution to this first problem with $k=1$ is given by:
\begin{equation}
    u(x,y) = \sin(a_1 \pi x)\sin(a_2 \pi y), \quad \forall (x,y) \in \Omega.
    \label{eq:helmholtz_sol}
\end{equation}
Given the wavenumbers $(a_1, a_2)$, the corresponding source terms $s(x,y)$ and $g(x,y)$ are computed by substitution into Eq.~\eqref{eq:helmholtz_eqn}. Here, $a_1 = a_2 = 6$ is set to produce a high-frequency solution.

A MLP with $H=6$ hidden layers with $W=128$ neurons per layer is employed.
The model is trained for 500{,}000 epochs using the Adam optimizer on a $51 \times 51$ uniform collocation grid, and evaluated on a finer $201 \times 201$ mesh. Given Adam's first-order nature, a decaying learning rate is essential for convergence over long durations, and we adopt the same \textit{ReduceLROnPlateau} scheduler as in the previous subsection. The CAPU algorithm is applied with a penalty scaling factor of $\eta = 1$, following the insights gained from the sensitivity study presented in~\ref{sec:appendixb}.

\begin{table}[]
    \centering
    \footnotesize
    \begin{tabular}{l|c|c|c|c|r|r}
        \hline
        \multirow{2}{*}{Method} & \multirow{2}{*}{N. Params} & \multirow{2}{*}{$H$} & \multirow{2}{*}{$W$} & \multirow{2}{*}{Optim.} & \multicolumn{2}{c}{Relative $l^2$} \\ \cline{6-7}
            &   &   &   &   & mean $\pm$ std & best \\ \hline
        PINN \cite{SHUKLA2024117290} & 82304 & 6 & 128 & \multirow{5}{*}{Adam} & - & $4.30\times10^{-2}$ \\ 
        cPIKAN+RBA \cite{SHUKLA2024117290} & 20960 & 5 & 32 &  & - & $3.81\times10^{-3}$ \\ 
        \cline{1-4} \cline{6-7}
        PECANN-MPU & \multirow{3}{*}{82304} & \multirow{3}{*}{6} & \multirow{3}{*}{128} &  & $12.53 \pm 7.93\times10^{-1}$ & $1.20\times10^{-1}$  \\ 
        PECANN-CPU &   &   &   &  & $1.30 \pm 2.54 \times10^{-1}$ & \boldsymbol{$3.36\times10^{-3}$} \\ 
        PECANN-CAPU &   &   &   &  & \boldsymbol{$6.69 \pm 3.38\times10^{-3}$} & $3.68\times10^{-3}$  \\
        \hline
        PECANN-CAPU & 3441 & 3 & 40 & L-BFGS & \textcolor{blue}{$1.33 \pm 0.653\times10^{-3}$}  & \textcolor{blue}{$5.61\times10^{-4}$} \\ 
    \end{tabular}
    \caption{Helmholtz equation (Eq. \ref{eq:helmholtz_sol} with $a_1 = a_2 = 6$): Relative $l^2$ comparison between different methods and the PECANN-CAPU method. All methods use the Adam optimizer for $5\times10^5$ epochs, except the last case that adopt L-BFGS for $5\times10^4$ epochs.}
    \label{tab:helm_pikan_case_c_l2}
\end{table}

Table~\ref{tab:helm_pikan_case_c_l2} presents a statistical comparison of relative $l^2$ norms, $\mathcal{E}_r$, across different methods. Under the same optimizer, both CPU and CAPU achieve the best-trial performance, with $\mathcal{E}_r \approx 3\times10^{-3}$, comparable to cPIKAN with RBA. However, CAPU consistently attains high accuracy, exhibiting a mean error approximately two orders of magnitude lower and the smallest standard deviation among the PECANN framework.
An additional configuration further enhances accuracy: when trained with the L-BFGS optimizer, even a shallow network comprising three hidden layers with 40 neurons each attains both accuracy and robustness, achieving a mean $\mathcal{E}_r$ of $1.33\times10^{-3}$ and a minimum of $5.61\times10^{-4}$.

\begin{figure}[!h]
\centering
    \subfloat[]{\includegraphics[scale=0.45]{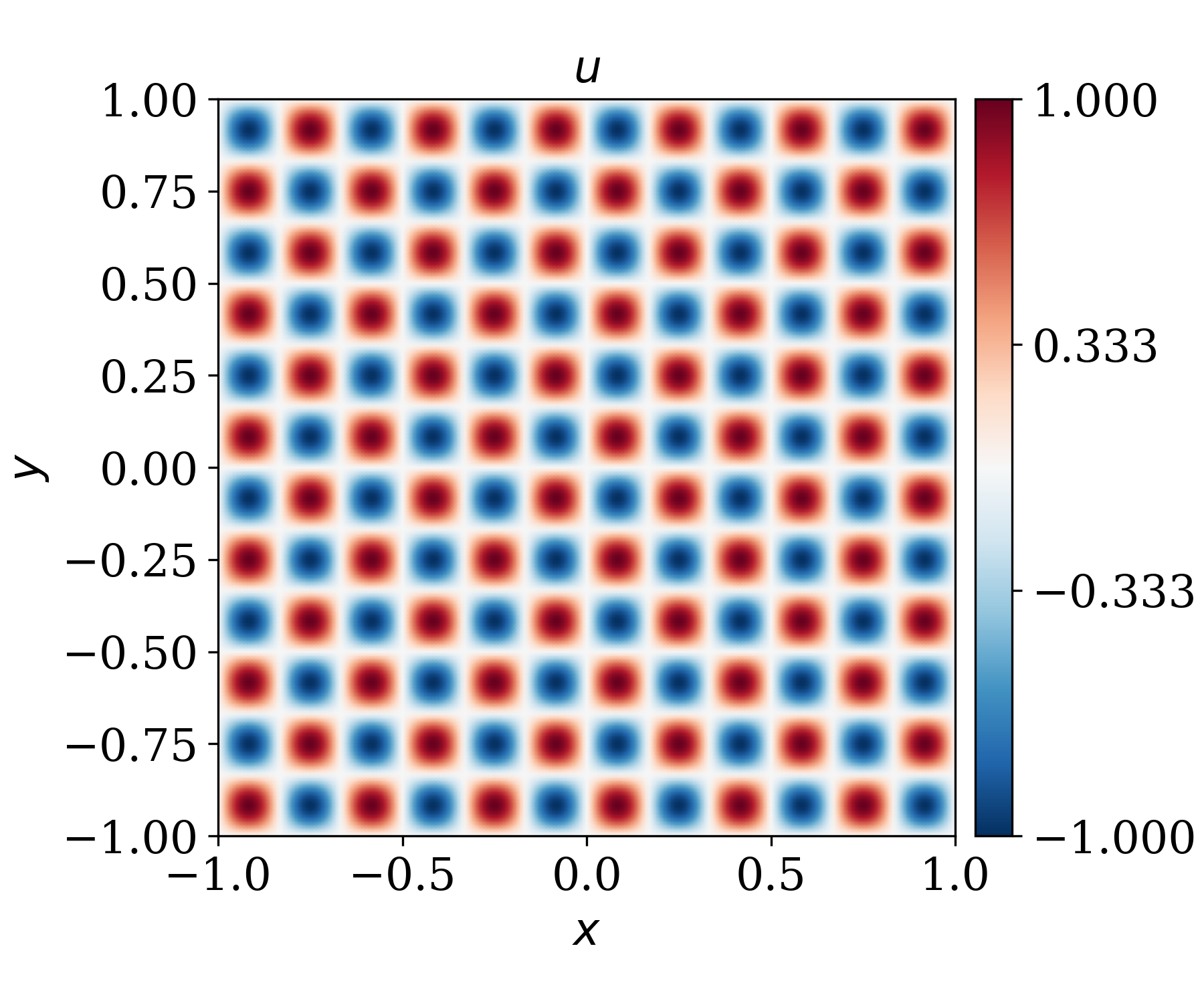}}\quad
    \subfloat[]{\includegraphics[scale=0.45]{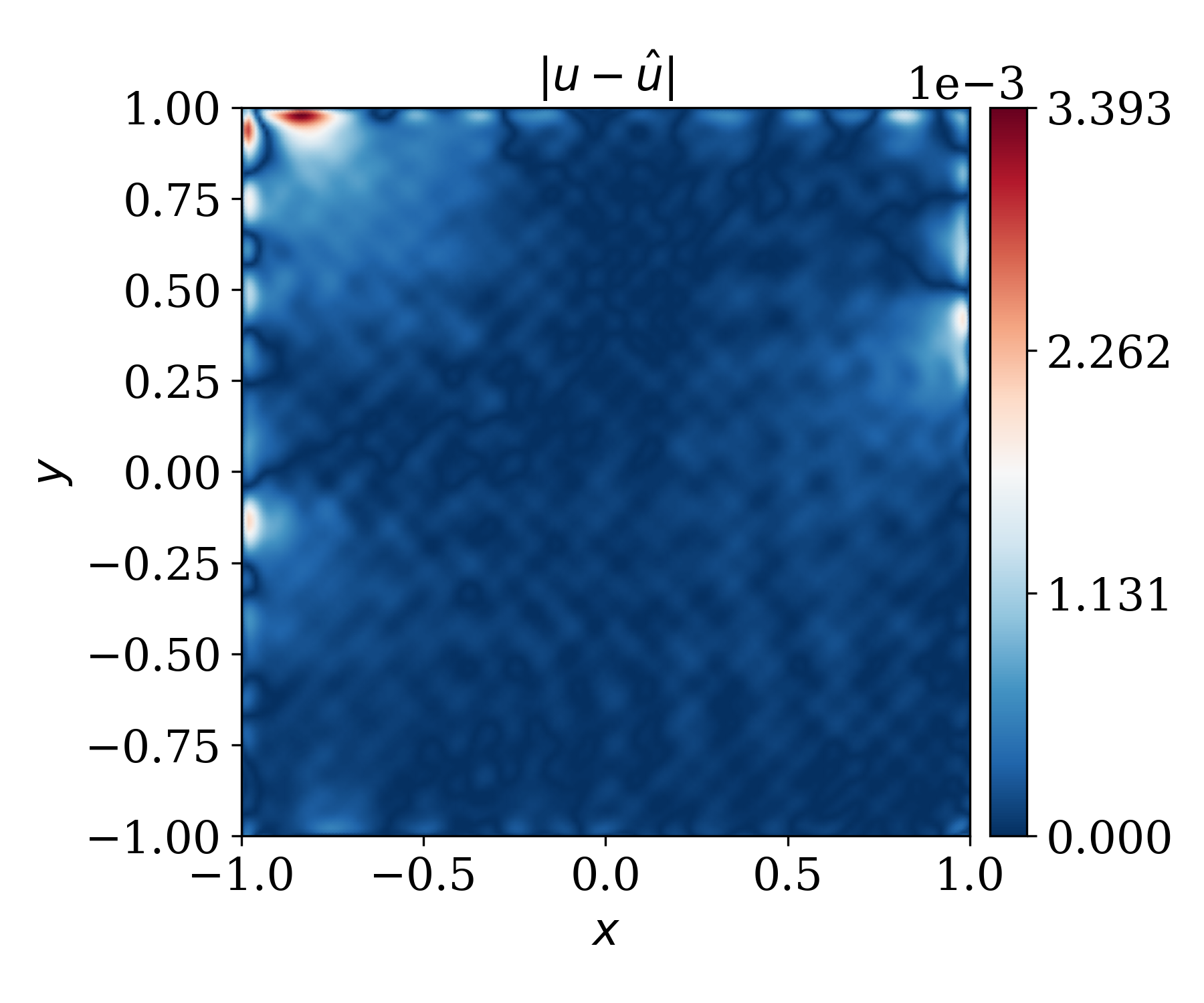}}
    \caption{Helmholtz equation (Eq. \ref{eq:helmholtz_sol} with $a_1 = a_2 = 6$): (a) PECANN-CAPU best trial prediction, (b) the corresponding absolute point-wise error.}
    \label{fig:helm_pikan_case_c_contour}
\end{figure}

Figure~\ref{fig:helm_pikan_case_c_contour}(a) displays PECANN predictions obtained using the CAPU algorithm. As seen in panel (b), we observe that absolute error level is generally very low, around $10^{-3}$.




\begin{figure}[!h]
\centering
    \subfloat[]{\includegraphics[scale=0.55]{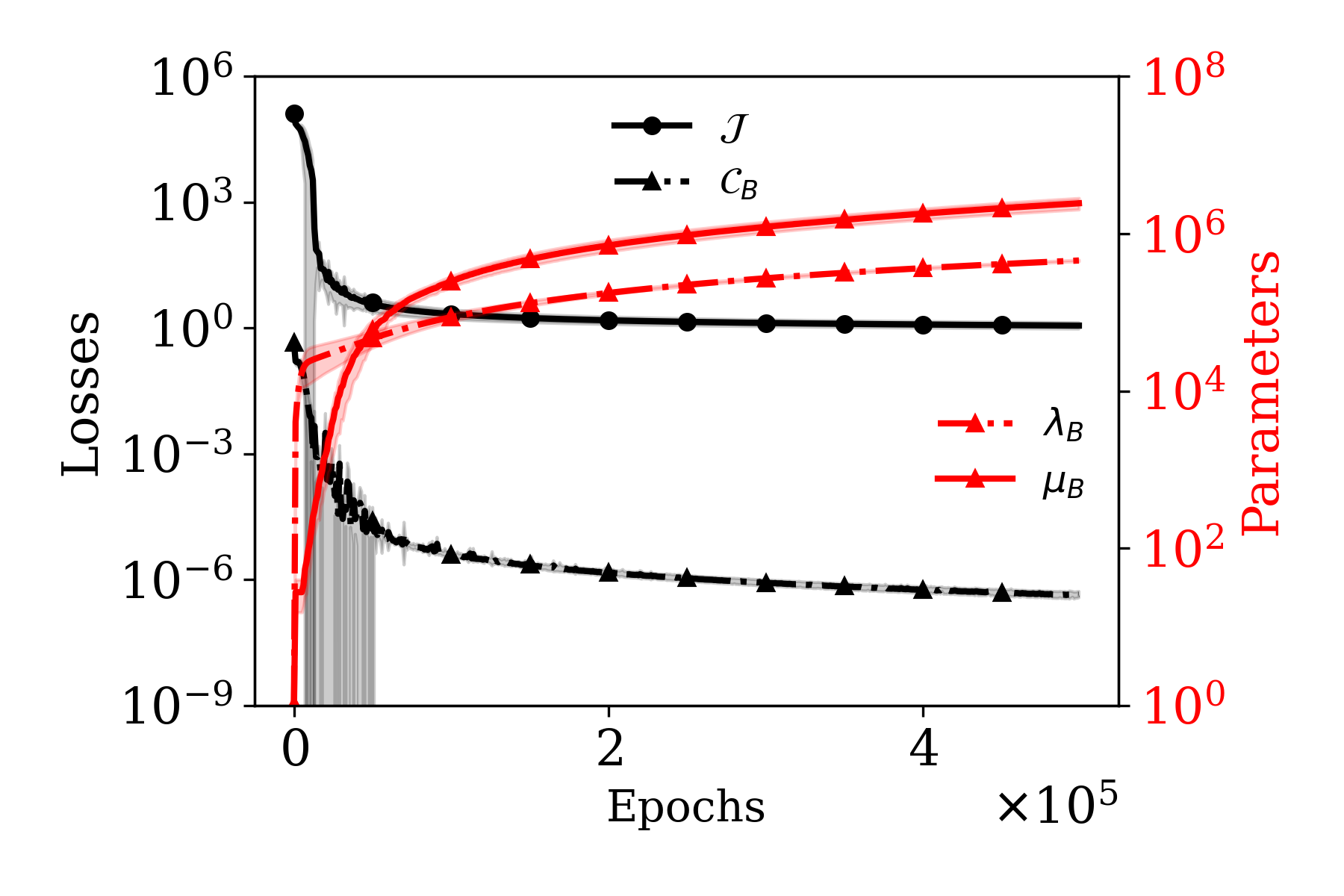}}\quad
    \subfloat[]{\includegraphics[scale=0.55]{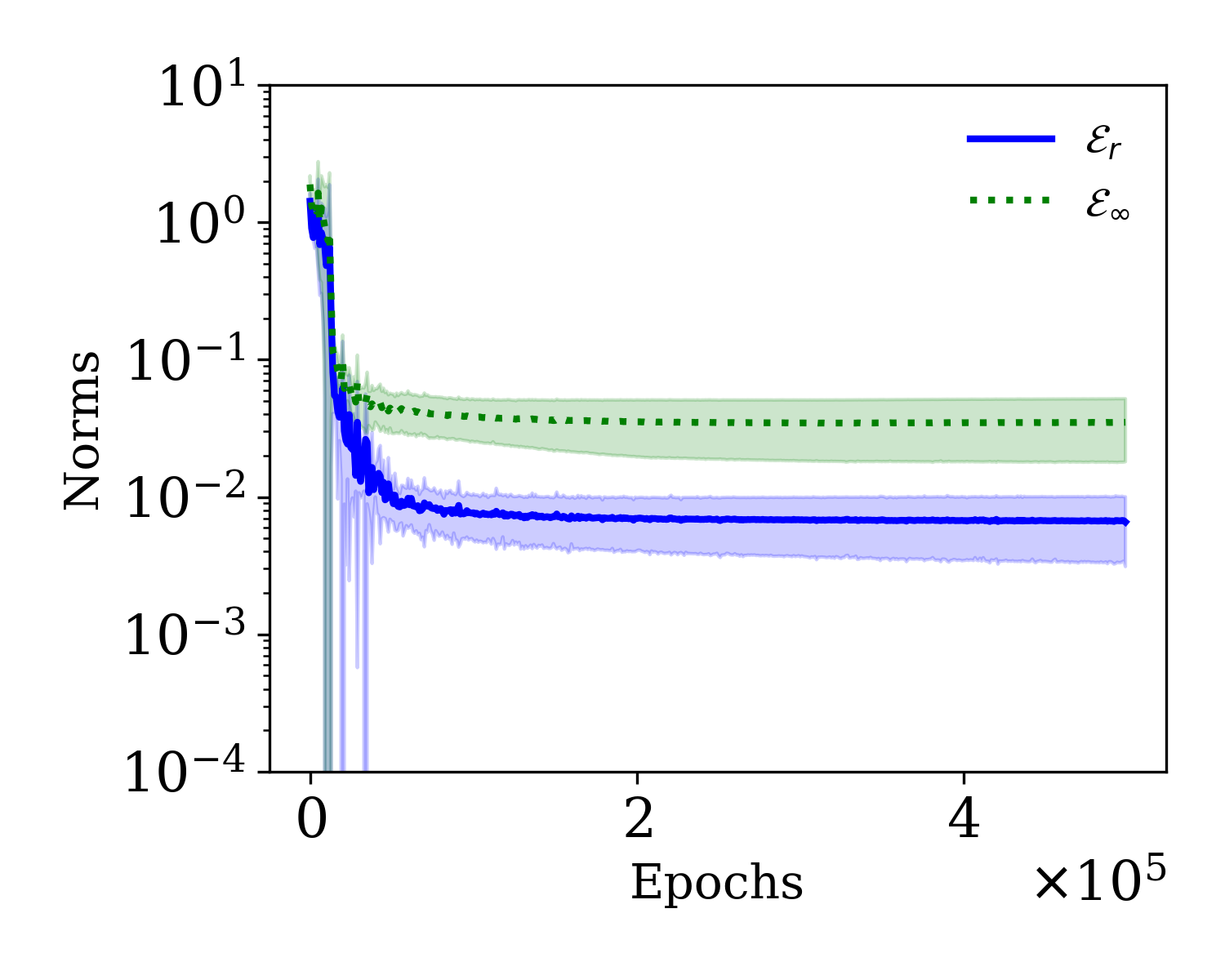}}
    \caption{Helmholtz equation (Eq. \ref{eq:helmholtz_sol} with $a_1 = a_2 = 6$): Evolution of mean and standard deviation of (a) the objective and constraint losses, (b) $l^\infty$ and relative $l^2$ norms using CAPU.}
    \label{fig:helm_pikan_case_c_evolution}
\end{figure}

Figure~\ref{fig:helm_pikan_case_c_evolution} shows the evolution of all the loss terms and evaluation metrics for the CAPU algorithm with Adam. In panel (a), a smooth and gradual increase of the penalty parameter $\mu_B$ is observed, interacting with the progressive decrease of the boundary constraint $\mathcal{C}_B$, as governed by the update rule in Eq.~\ref{eq:capu_mu}. This interaction rapidly amplifies the Lagrange multiplier, leading to $\lambda_B > \mu_B$ during the early training stage; subsequently, the growth rate slows, and the constraint stabilizes. In panel (b), the network converges after approximately $10^5$ epochs, with the relative $l^2$ norm stabilizing below $10^{-2}$. 
Note that the shallow network trained with L-BFGS converges also fast-by $10^4$ epochs-with a lower mean error near $10^{-3}$. 

\subsubsection{A much more challenging Helmholtz equation problem}\label{sec:helmholtz_challenging}
In their FBPINN work, \citet{DOLEAN2024} considered the 2D Helmholtz equation~\eqref{eq:helmholtz_eqn} on the domain 
\(\Omega = \{(x, y) \mid 0 \leq x, y \leq 1\}\), with zero Dirichlet boundary conditions \(g = 0\) and a localized Gaussian-like source term:
\begin{equation}
    s(\mathbf{x}) = \frac{1}{2\pi \sigma^2} 
    \exp\left(-\frac{\|\mathbf{x} - 0.5\|_2^2}{2\sigma^2}\right),
    \label{eq:helm_multi_source}
\end{equation}
The wavenumber and the width of the Gaussian source term are defined as:
\begin{equation}
    k = \frac{2^L\pi}{1.6}, \quad \sigma = \frac{0.8}{2^L},
    \label{eq:helm_multi_para}
\end{equation}
The solution complexity can be increased systematically by increasing the parameter \(L\).

This specific problem represents a more realistic application of the Helmholtz equation and has no analytical solution. Therefore, we generate reference data using a second-order central difference scheme on a $361 \times 361$ uniform grid, using the same finite difference code adopted in \cite{DOLEAN2024}. 
It is worth noting that for $L \ge 6$, the finite-difference solution shows inconsistencies, even after increasing the mesh resolution. The four different methods deployed in \citet{DOLEAN2024} failed to produce a common solution. Therefore, we restrict our analysis to cases with $L \le 5$.
We used the less complex $L = 4$ case to better understand the improvement of our CAPU algorithm relative to the original RMSProp scheme. Those results are presented in \ref{sec:appendixa}.

Setting $L = 5$ results in a highly concentrated source term and a large Helmholtz wavenumber, thereby significantly increasing the complexity of the problem. To enhance the approximation capability of the PECANN framework with the CAPU algorithm, we augment the network architecture by incorporating a low-frequency Fourier feature mapping, sampled from a standard Gaussian distribution ($\sigma = 1$). The training data consist of $160^2$ randomly sampled residual points and 160 boundary points per edge.

Table~\ref{tab:helm_multi_l5_norms} presents a comparison of the relative $l^2$ errors between PECANN predictions using standard MLPs and those using Fourier feature mappings, with comparable parameter counts. Additional comparisons investigate various Fourier-based architectures by varying the number of hidden layers ($H$) and the number of neurons per layer ($W$).

\begin{table}[h!]
    \centering
    \footnotesize
    \begin{tabular}{l|c|c|c|r|r}
        \hline
        \multirow{2}{*}{Architecture} & \multirow{2}{*}{N. Params} & \multirow{2}{*}{$H$} & \multirow{2}{*}{$W$} & \multicolumn{2}{c}{Relative $l^2$} \\ \cline{5-6}
            &   &   &   & mean $\pm$ std & best \\ 
        \hline
        MLP & 921 & 3 & 20 & $9.20 \pm 0.24\times10^{-1}$ & $8.83\times10^{-1}$ \\
        Fourier ($\sigma=1$) & 861 & 3 & 20 & $7.42 \pm 0.91\times10^{-1}$ & $6.14\times10^{-1}$ \\ 
        \hline
        MLP & 3441 & 3 & 40 & $8.61 \pm 0.18\times10^{-1}$ & $8.43\times10^{-1}$ \\  
        Fourier ($\sigma=1$) & 3381 & 9 & 20 & $1.78 \pm 0.43\times10^{-1}$ & $1.33\times10^{-1}$ \\
        Fourier ($\sigma=1$) & 3321 & 3 & 40 & $1.45 \pm 0.38\times10^{-1}$ & $9.64\times10^{-2}$ \\ 
        \hline
        \multirow{3}{*}{Fourier ($\sigma=1$)} & 7161 & 18 & 20 & $5.25 \pm 2.52\times10^{-1}$ & $2.19\times10^{-1}$ \\
            & 7381 & 3 & 60 & $8.89 \pm 1.16\times10^{-2}$ & \boldsymbol{$6.73\times10^{-2}$} \\
            & 13041 & 3 & 80 & $9.22 \pm 0.54\times10^{-2}$ & $8.28\times10^{-2}$ \\
    \end{tabular}
    \caption{2D Helmholtz equation with a localized source term (Eq. \ref{eq:helm_multi_source} with $L=5$): Relative $l^2$ norms of PECANN predictions with standard MLPs and networks with a $\sigma=1$ Fourier feature mapping, using the CAPU algorithm. Models are trained with the L-BFGS optimizer for $8\times10^4$ epochs and the CAPU algorithm.}
    \label{tab:helm_multi_l5_norms}
\end{table}

High mean $\mathcal{E}_r$ values on the order of $10^{-1}$ indicate that a width of $W = 20$ is insufficient for solving the $L = 5$ case, regardless of using MLP or Fourier-feature architectures. Doubling the width to $W = 40$ markedly improves both accuracy and consistency when Fourier features are included, with the best-performing trial achieving $\mathcal{E}_r = 9.642 \times 10^{-2}$—outperforming the corresponding MLP by an order of magnitude less error. Among models with similar parameter counts, shallower, wider Fourier networks (e.g., $H = 3$, $W = 40$) outperform deeper ones (e.g., $H = 9$, $W = 20$), with the gap widening at matched sizes around 7200 parameters. Accuracy continues to improve with model size in wide configurations, converging at $H = 3$, $W = 60$ with a minimum $\mathcal{E}_r = 6.726 \times 10^{-2}$. This degradation in deeper networks may stem from vanishing gradients and the limitations of the $\tanh$ actuation function in representing periodic nonlinearities \cite{sitzmann2020periodic}, which motivates further investigation into the role of Fourier feature mapping at the input layer.

\begin{figure}[!h]
\centering
\includegraphics[width=0.6\textwidth]{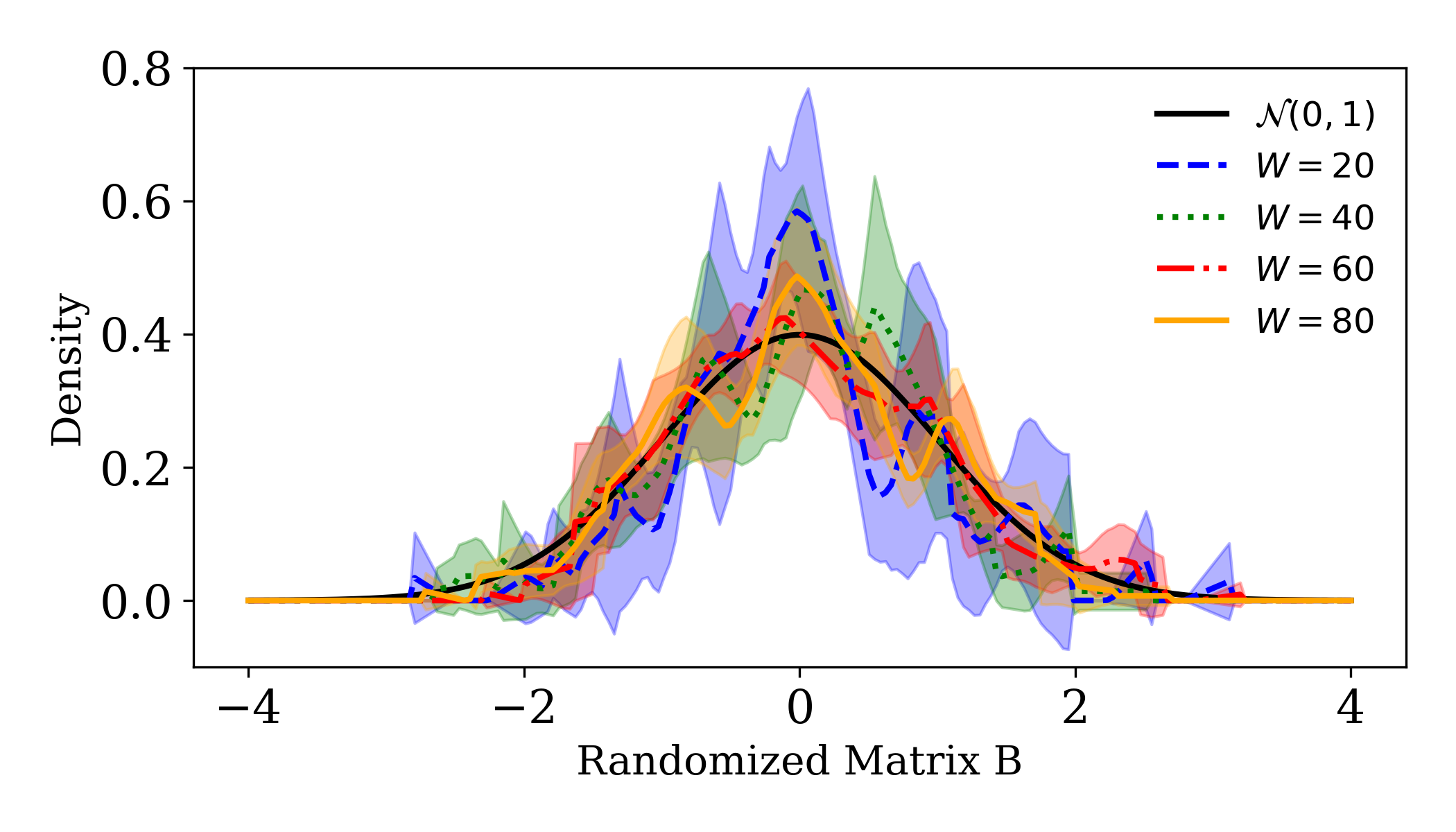}
\caption{2D Helmholtz equation with a localized source term (Eq. \ref{eq:helm_multi_source} with $L=5$): Mean and deviation bands of probability density distributions with elements in the random matrix $\mathbf{B}$ over five trials for Fourier networks with varying neuron counts, compared to the standard normal distribution.}
\label{fig:helm_multi_l5_B_distri}
\end{figure}

To better understand the impact of Fourier feature mapping under similar parameter counts, we analyze the probability density distributions of elements in the randomized matrix $\mathbf{B}$. Figure~\ref{fig:helm_multi_l5_B_distri} shows the mean and standard deviation bands across five trials for networks of varying width, compared to the standard normal distribution $\mathcal{N}(0,1)$. 
For $W = 20$, the distributions deviate significantly: although the mean remains symmetric, the standard deviation is clearly below 1, with wide variation. Increasing to $W = 40$ reduces this mismatch, consistent with earlier results—models with $W = 20$ consistently underperform, while doubling the width improves accuracy with some variability. As $W$ increases to 60 and 80, the distributions increasingly align with $\mathcal{N}(0,1)$, yielding further but saturating gains. This plateau in accuracy, as shown in Table~\ref{tab:helm_multi_l5_norms}, suggests diminishing returns from further distributional alignment. 
Given the lack of an exact analytical solution, we do not seek higher accuracy relative to the reference, which may itself be affected by mesh resolution or discretization scheme.

\begin{figure}[!h]
\centering
    \subfloat[]{\includegraphics[width=0.48\textwidth]{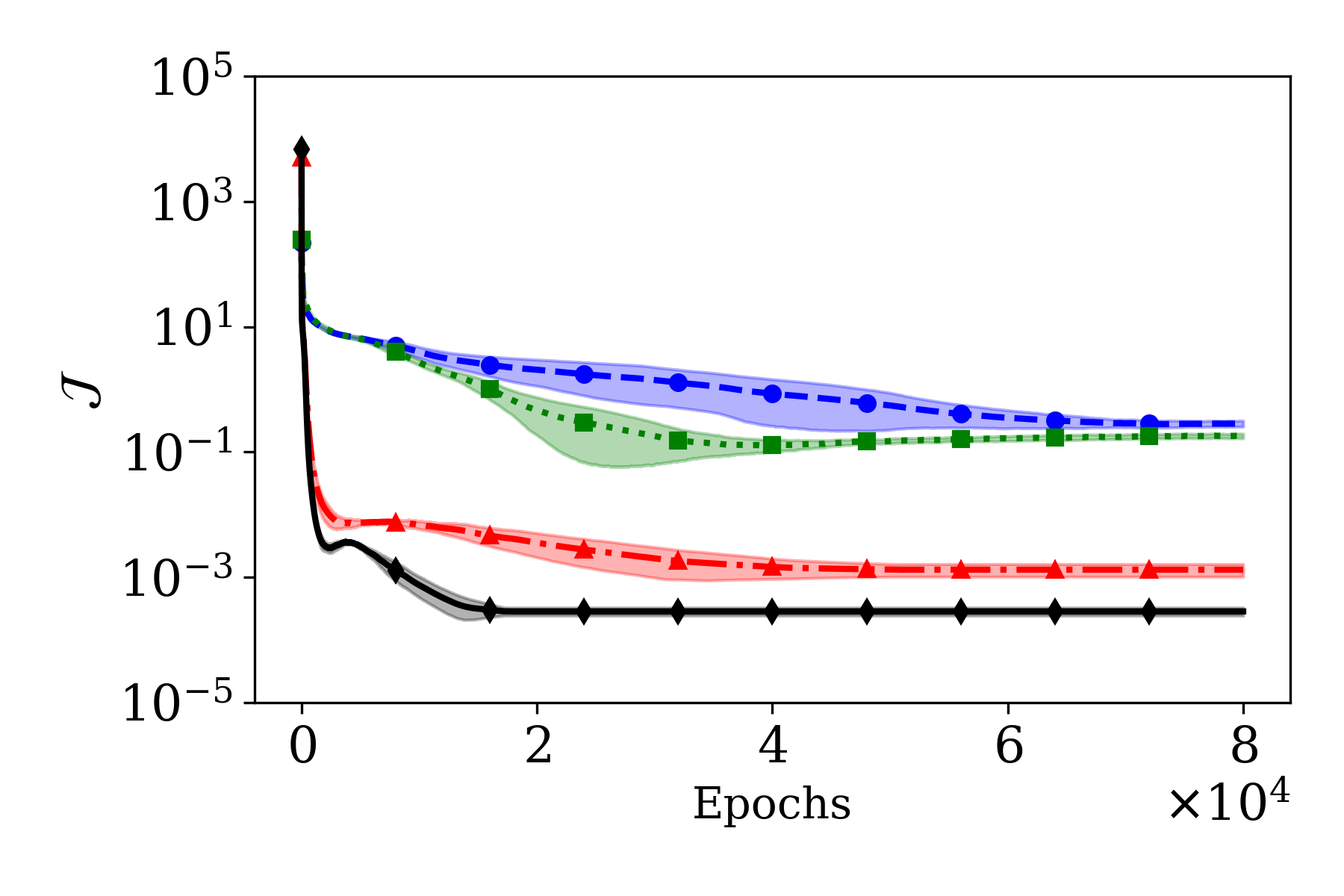}}\quad
    \subfloat[]{\includegraphics[width=0.48\textwidth]{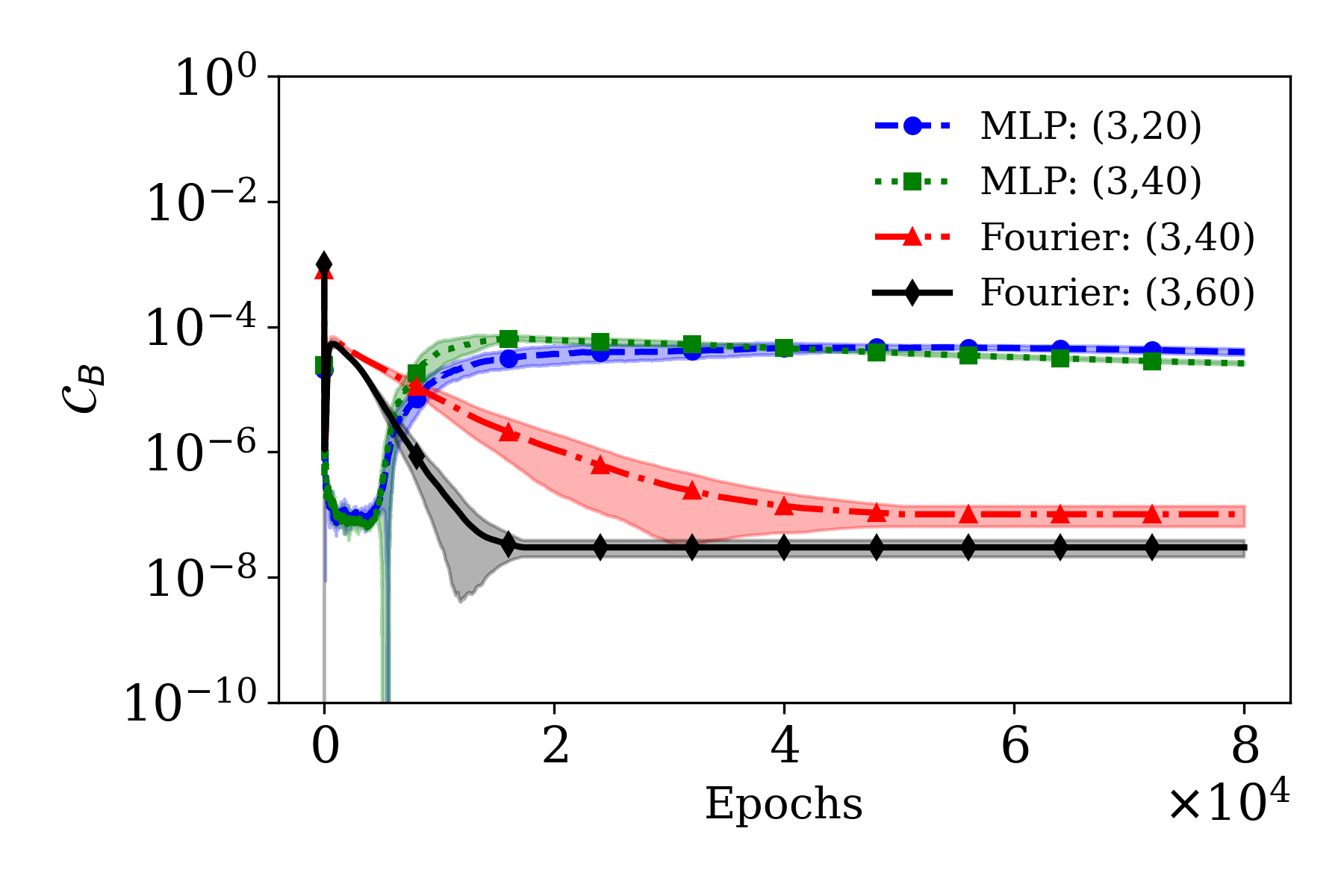}}\quad
    \caption{2D Helmholtz equation with a localized source term (Eq. \ref{eq:helm_multi_source} with $L=5$): Evolution of the mean and standard deviation of the objective (a) and boundary constraint (b) for different MLP and Fourier-featured network architectures.}
    \label{fig:helm_multi_l5_evol}
\end{figure}

Figure~\ref{fig:helm_multi_l5_evol} compares MLP and Fourier-featured networks by tracking the evolution of the objective $\mathcal{J}$ and boundary constraint $\mathcal{C}_B$. In panel (a), MLPs with $W = 20$ and $W = 40$ converge to $\mathcal{J} \approx 10^0$, while in panel (b), their $\mathcal{C}_B$ values rise and plateau just below $10^{-4}$, indicating boundary violations and waveform leakage beyond the domain for $L = 5$. This reflects the MLPs’ limited capacity at higher Helmholtz wavenumber, in contrast to their adequate performance at $L = 4$ (\ref{sec:appendixa}). While increasing the penalty scaling factor (e.g., to 10) may enhance boundary enforcement, the slow decay of $\mathcal{J}$ highlights the inherent challenge of this high-frequency case.

With a low-frequency Fourier mapping ($\sigma = 1$) at the input, the $W = 40$ network shows a rapid initial drop in $\mathcal{J}$—well below that of the corresponding MLP—followed by a slower decline around epoch $10^4$ in Fig.~\ref{fig:helm_multi_l5_evol}(a). This slowdown corresponds to a gradual reduction in the boundary constraint $\mathcal{C}_B$, as shown in Fig.~\ref{fig:helm_multi_l5_evol}(b), where CAPU steadily enforces the boundary condition. Both losses converge around epoch $3 \times 10^4$, with $\mathcal{J} \approx 10^{-3}$ and $\mathcal{C}_B$ below $10^{-8}$. 
Increasing the width to $W = 60$ further improves performance: $\mathcal{J}$ rapidly drops below $10^{-2}$ and briefly plateaus, while CAPU effectively suppresses boundary violations, leading both terms to converge before epoch $2 \times 10^4$.

\begin{figure}[!h]
\centering
    \subfloat[]{\includegraphics[width=0.3\textwidth]{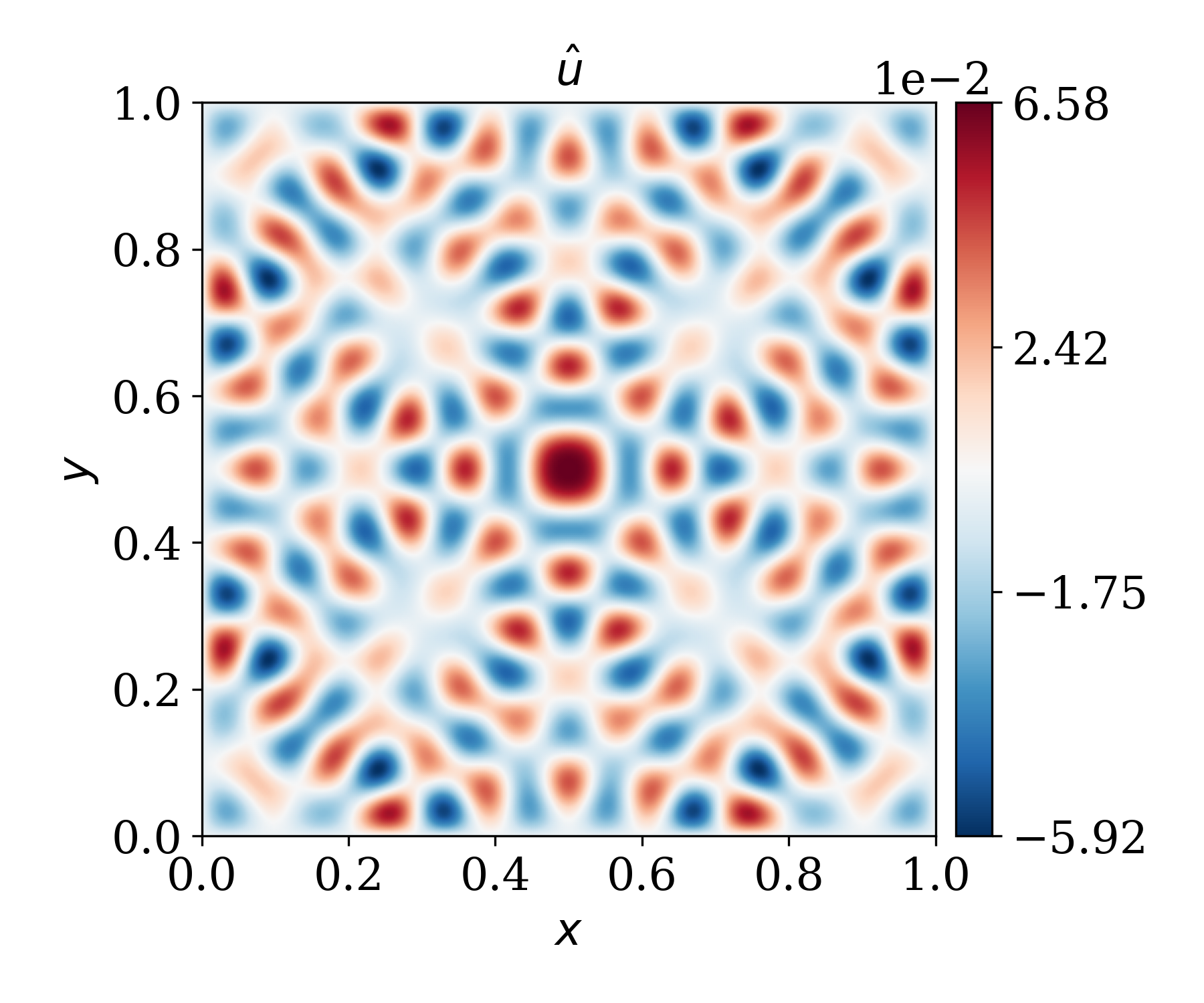}}\quad
    \subfloat[]{\includegraphics[width=0.3\textwidth]{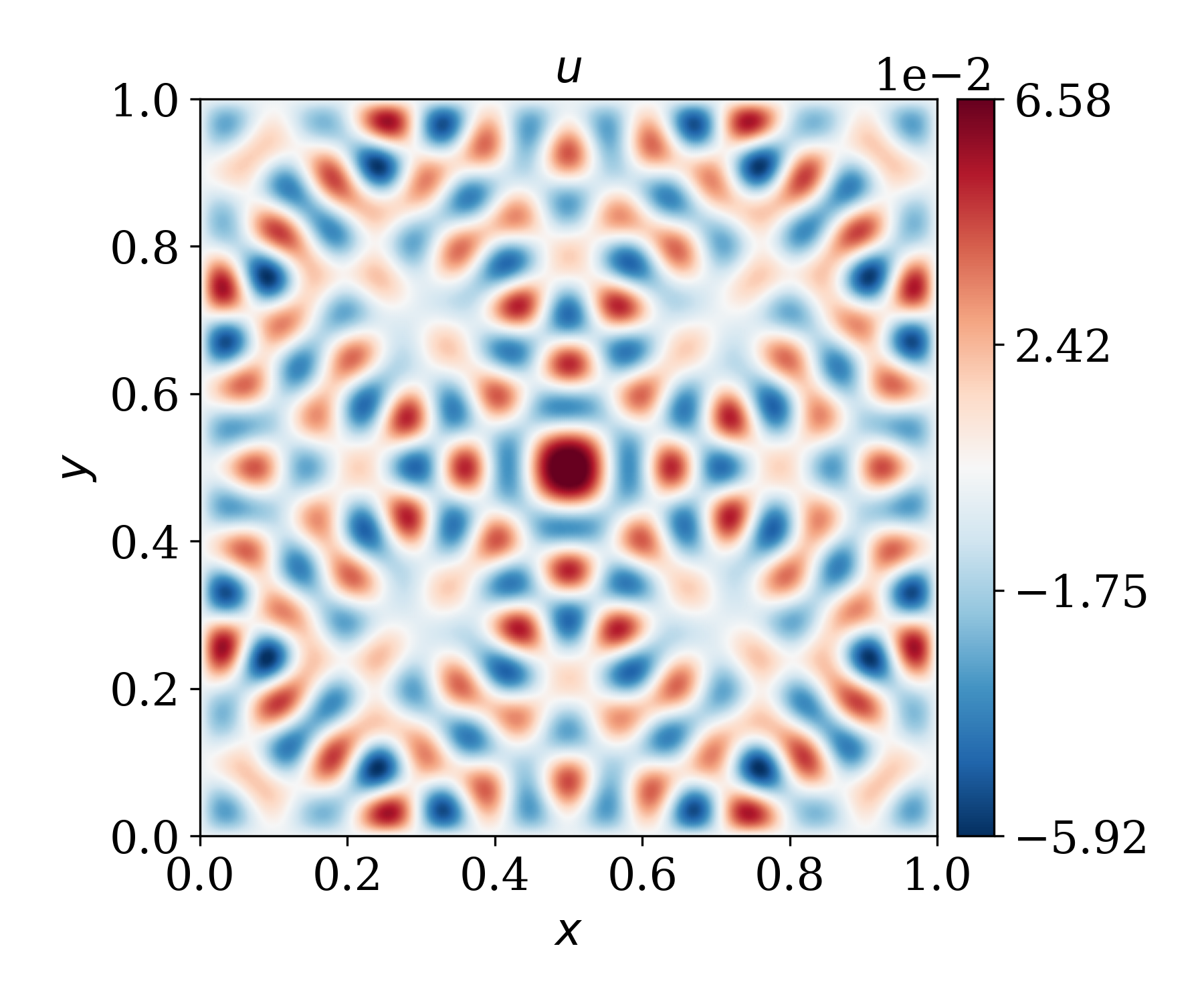}}\quad
    \subfloat[]{\includegraphics[width=0.3\textwidth]{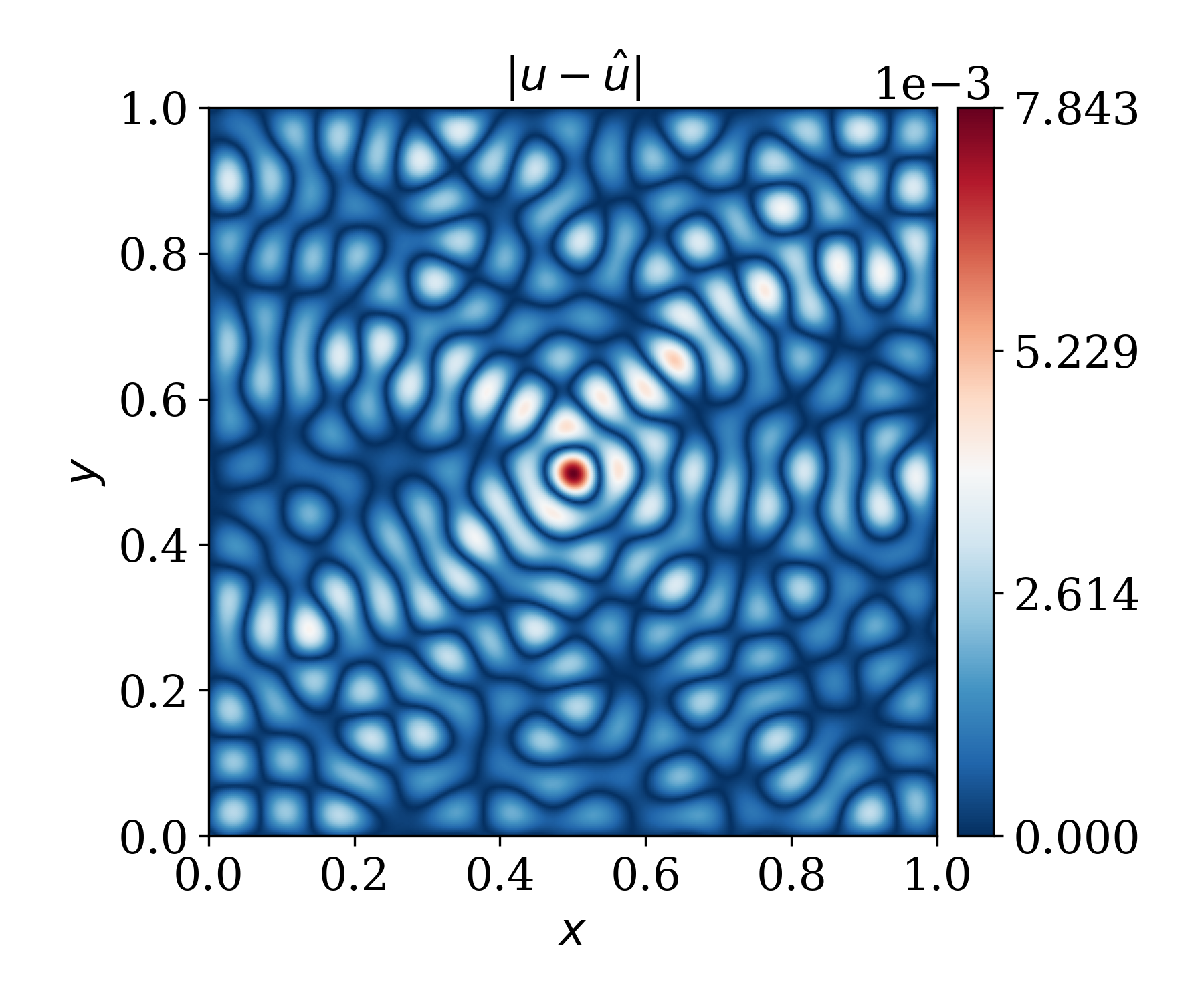}}
    \caption{2D Helmholtz equation with a localized source term (Eq. \ref{eq:helm_multi_source} with $L=5$): (a) reference solution, (b) prediction of the best-performing trial using the CAPU algorithm with Fourier-featured network of 3 hidden layers and 60 neurons per layer, (c) the corresponding absolute point-wise error.}
    \label{fig:helm_multi_l5_contour}
\end{figure}

Figure~\ref{fig:helm_multi_l5_contour} illustrates the reference solution of $L=5$, and the prediction from the lowest-$\mathcal{E}_r$ trial, as well as the absolute point-wise error. Overall, the agreement between the exact and numerical solution is very good for this challenging case. The primary absolute error is concentrated at the center of the square domain, with a maximum value of $7.843 \times 10^{-3}$.

\subsection{Forward problem: reversible advection of a passive scalar by a single vortex}\label{sec:reversible_adv}
Baseline PINN models struggle with long-time dynamics when trained over the entire spatio-temporal domain. \citet{Wang2024causality} showed that this failure happens because baseline PINN models exhibit an implicit bias toward minimizing PDE residuals at later times before fitting initial conditions. A practical alternative to address this issue is to use discrete time models with time-marching schemes such as the Runge-Kutta methods \cite{raissi2019pinn}. \citet{Wang2024causality} proposed a causal training algorithm that restores physical causality in PINNs by adaptively re-weighting the PDE residual loss toward initial times during training of continuous time models. 

In contrast to previously discussed approaches, we adopt a time-windowing strategy that is enabled by the constrained optimization formulation employed in the PECANN framework. In our approach the final solution of a time window is enforced as an initial condition constraint for the next time window in a principled fashion. 

To evaluate this time-windowing strategy, we consider a challenging benchmark problem: reversible scalar advection by a single vortex flow. Originally introduced by \citet{leveque1996} and \citet{rider1998singlevortex}, this benchmark is well-established for assessing  various interface tracking and front capturing techniques used in two-phase flow simulations, including the volume-of-fluid (VOF) method \citep{rider1998singlevortex, harvie2000singlevortex}, the level-set method \citep{ENRIGHT2002, HIEBER2005}, and the moment-of-fluid (MOF) method \citep{AHN2009}.

In this benchmark, a circular dye distribution is advected and then the flow is reversed to observe to what accuracy level the original circular shape is recovered. This process serves as a stringent test of a numerical method’s conservation and diffusion properties. For PECANN, successful recovery also highlights its capability for accurate long-time PDE integration.

The test problem is governed by the advection of a passive scalar $\phi$ by a predefined time-dependent flow field without any physical diffusion in the scalar transport equation
\begin{equation}
    \frac{\partial \phi}{\partial t} + \mathbf{u}\cdot \nabla \phi =0,
\end{equation}
where $\phi \in [0,1]$ is the volume fraction function for a scalar field, which is advected by the velocity field $\mathbf{u}$. The initial condition is defined by setting \(\phi = 1\) within a circle of radius 0.15 centered at \((0.5, 0.75)\) inside a unit square domain, and $\phi=0$ elsewhere. This results in a scalar field with a discontinuous interface. For the domain boundaries, $\phi$ is always set to zero. 

A time-dependent, divergence-free velocity field is defined by the following stream function:
\begin{equation}
    \Psi(x, y, t) = \frac{1}{\pi} \sin^2(\pi x) \sin^2(\pi y) \cos\left(\frac{\pi t}{T}\right), \quad x, y \in [0,1],
\end{equation}
where \(T\) is the specified period, chosen as $8.0$ for this test. 

To predict the long-time evolution of this scalar advection problem, we propose a non-overlapping time-marching strategy. Unlike conventional finite-difference-based time-marching schemes, our approach fundamentally differs in its formulation and implementation. Specifically, the entire temporal domain is partitioned into smaller, non-overlapping subdomains. The predicted solution within each temporal-spatial subdomain is subsequently employed as an initial condition constraint for the succeeding subdomain. This sequential conditioning is facilitated by the underlying constrained optimization framework of PECANN, which enables principled enforcement of temporal-spatial consistency within each subdomain.

In the present problem, each time window spans \(\Delta t = 0.2\), dividing the total simulation period into 40 uniform intervals. A separate neural network models each subdomain, taking $(x, y, t)$ as input and producing a scalar output $\phi$. The network architecture consists of 6 hidden layers ($H=6$), each with 40 neurons. Since the passive scalar $\phi$ is bounded between 0 and 1 as prior knowledge, a sigmoid activation function is employed in the output layer.
Accordingly, the output in the $i$th subdomain is given by
\begin{equation}
\phi_i = f_{\theta^{i}}(x, y, t) = \mathrm{sigmoid}(\mathbf{W}_{H+1} \cdot \mathbf{H}_H + \mathbf{b}_{H+1}), \quad \text{for } t \in [(i - 1)\Delta t, i\Delta t],
\end{equation}
where $\theta^{i}$ denotes the trainable parameters of the $i$th model.
The time-marching update is performed via:
\begin{equation}
\phi(x, y, {(i - 1)\Delta t}; \theta^{i}) \leftarrow \phi(x, y, (i - 1)\Delta t; \theta^{i-1}).
\end{equation}
Each subdomain is sampled using two levels of uniform meshes: a coarse mesh \(129 \times 129 \times 11\) and a fine mesh \(257 \times 257 \times 11\). The first two dimensions correspond to equally spaced points in the spatial domain, aligned with the setup in \citet{xie_consistent_2020}, while the third dimension denotes the number of points along the temporal axis spanning the time window $\Delta t =0.02$. The sampled points include interior residual points as well as boundary and initial condition points.
Training begins with the Adam optimizer for 3000 epochs, followed by L-BFGS for 2000 epochs. The CAPU algorithm is applied with a penalty scaling factor $\eta = 0.01$ for both initial and boundary constraints.
\begin{figure}[!h]
\centering
    \includegraphics[width=1\textwidth]{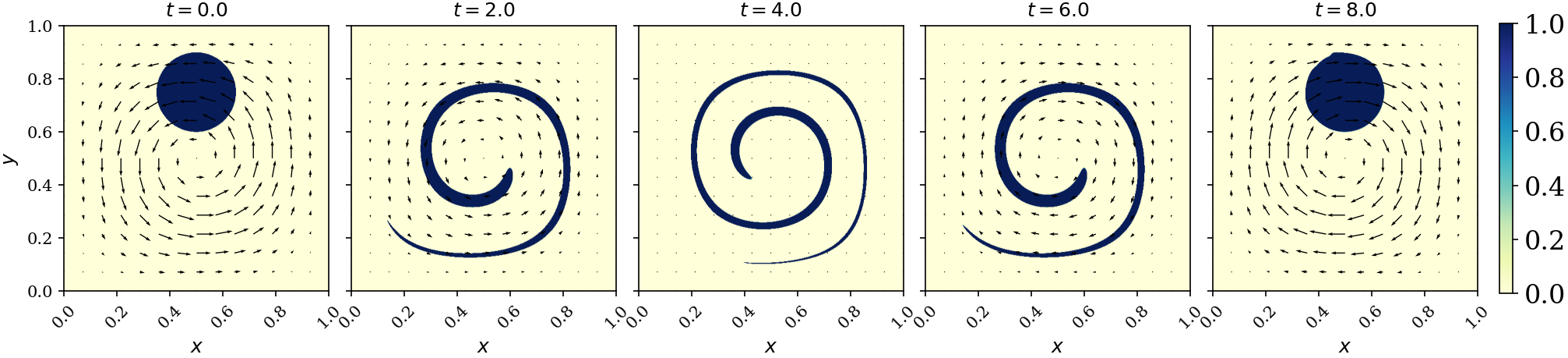}
    \caption{Reversible transport of a passive scalar field by a vortex. Scalar field at $t = 0$, $2$, $4$, $6$, and $8$ on the fine mesh using the CAPU algorithm.}
    \label{fig:vortex_flow_t8_pred_evol}
\end{figure}

Figure~\ref{fig:vortex_flow_t8_pred_evol} displays the time evolution of predicted scalar on the fine mesh at five snapshots ($t = 0, 2, 4, 6, 8$). Initially, a circular passive scalar is accurately captured at $t=0$. As the vortex flow progresses, the scalar is stretched into a spiral structure with a thin, elongated tail, reaching maximum deformation at $t=4$. This process introduces strong interface distortion and broadens the discontinuity, creating a highly challenging scenario for interface tracking methods to capture. In the second half of the simulation, the vortex rotation reverses direction to recover the original shape. At $t=6$, the interface resembles that at $t=2$, indicating a near-symmetric reversal of deformation. By the end of the period ($t=8$), the original circular shape is nearly restored, demonstrating superior conservation and minimal numerical smearing in the method.

\begin{figure}[!h]
\centering
    \subfloat[]{\includegraphics[width=0.48\textwidth]{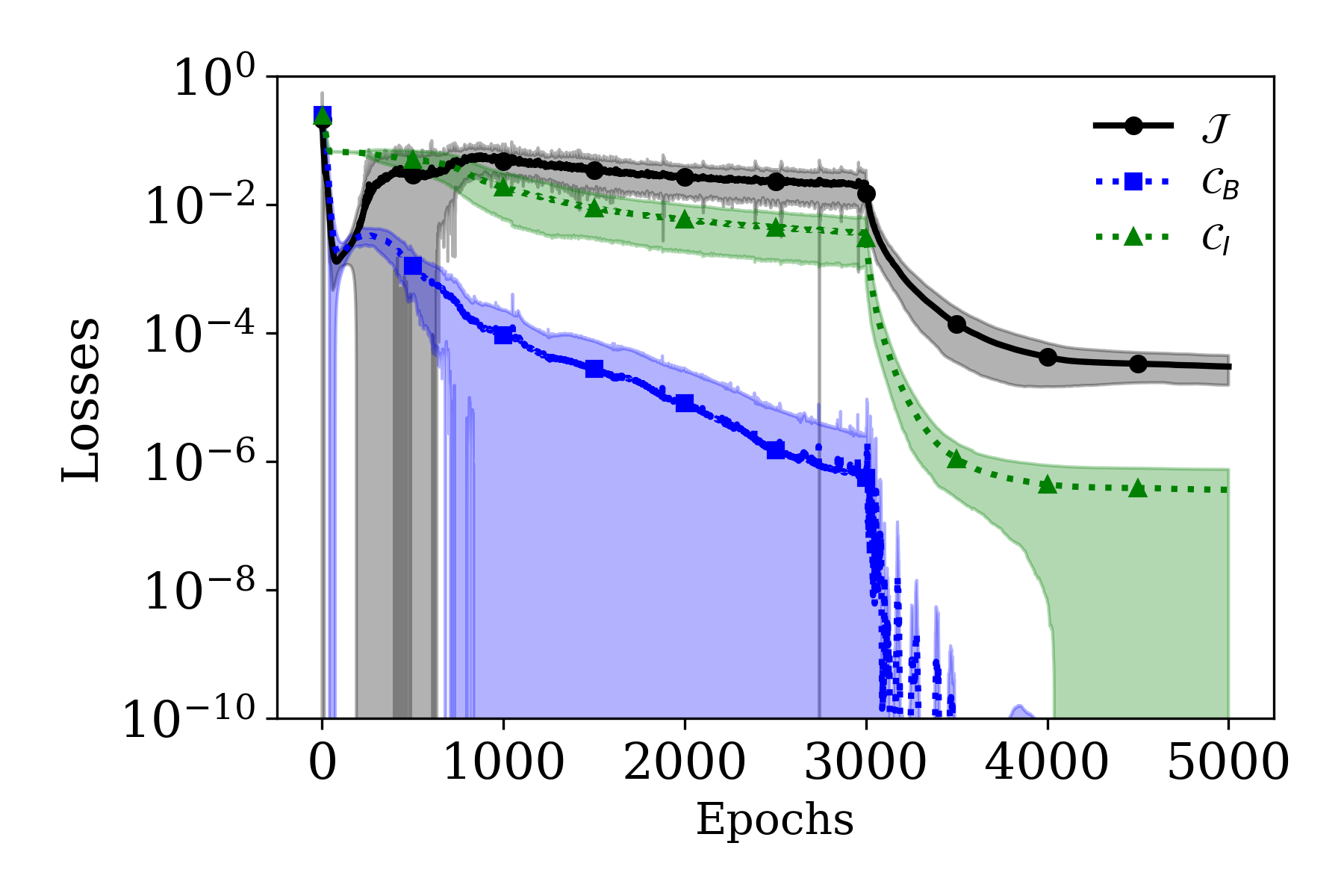}} \quad
    \subfloat[]{\includegraphics[width=0.48\textwidth]{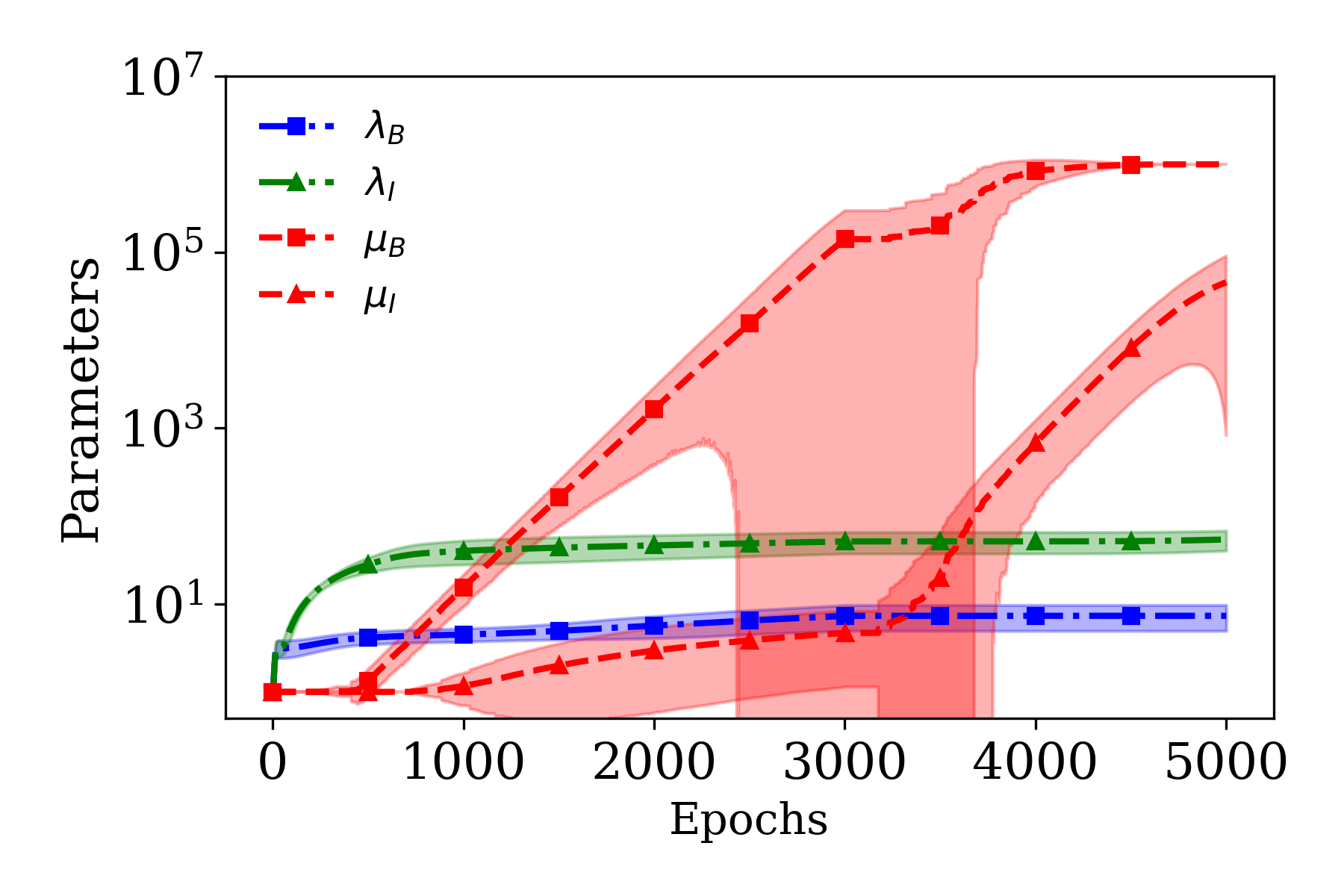}} \quad
    \caption{Reversible transport of a passive scalar field by a vortex: Evolution of the mean and standard deviation during training across all 40 subdomain models on the fine mesh using the CAPU algorithm. (a) Loss terms (objective, boundary, and initial constraints); (b) Corresponding Lagrange multipliers and penalty parameters.}
    \label{fig:vortex_flow_t8_loss_para_evol}
\end{figure}

Figure~\ref{fig:vortex_flow_t8_loss_para_evol} shows the evolution of loss terms and optimization parameters across the 40 subdomain models trained with the CAPU algorithm. In panel (a), switching from Adam to L-BFGS yields significant convergence improvements, with the objective $\mathcal{J}$ dropping by nearly three orders of magnitude and stabilizing below $10^{-4}$. The initial constraint $\mathcal{C}_I$ consistently decreases to around $10^{-6}$, while the boundary constraint $\mathcal{C}_B$ falls sharply below $10^{-10}$—demonstrating CAPU’s effectiveness in enforcing constraints within each time window. This robust performance ensures more reliable predictions for subsequent windows and contributes to the method’s superior accuracy.

In Fig.~\ref{fig:vortex_flow_t8_loss_para_evol}(b), the boundary penalty parameter $\mu_B$ rapidly approaches its upper bound of $10^6$, triggered by the sharp decline in $\mathcal{C}_B$. This limit is implicitly imposed by the adaptive update rule in Eq.~\eqref{eq:capu_mu}, given $\bm{\eta}=0.01$ and $\epsilon=10^{-16}$. In contrast, Lagrange multipliers stabilize early in training. 
\begin{figure}[!h]
\centering
    \subfloat[]{\includegraphics[width=0.35\textwidth]{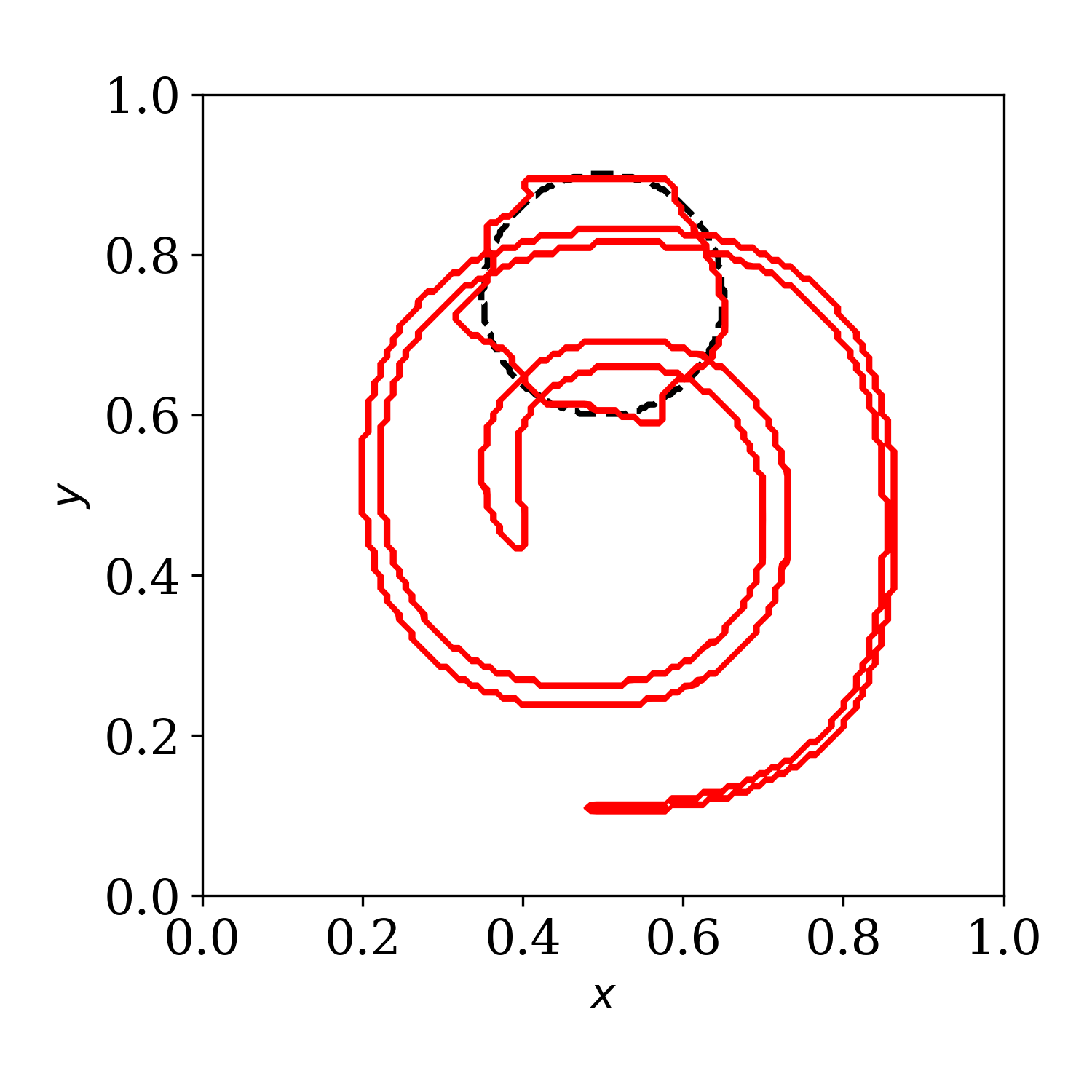}}\quad
    \subfloat[]{\includegraphics[width=0.35\textwidth]{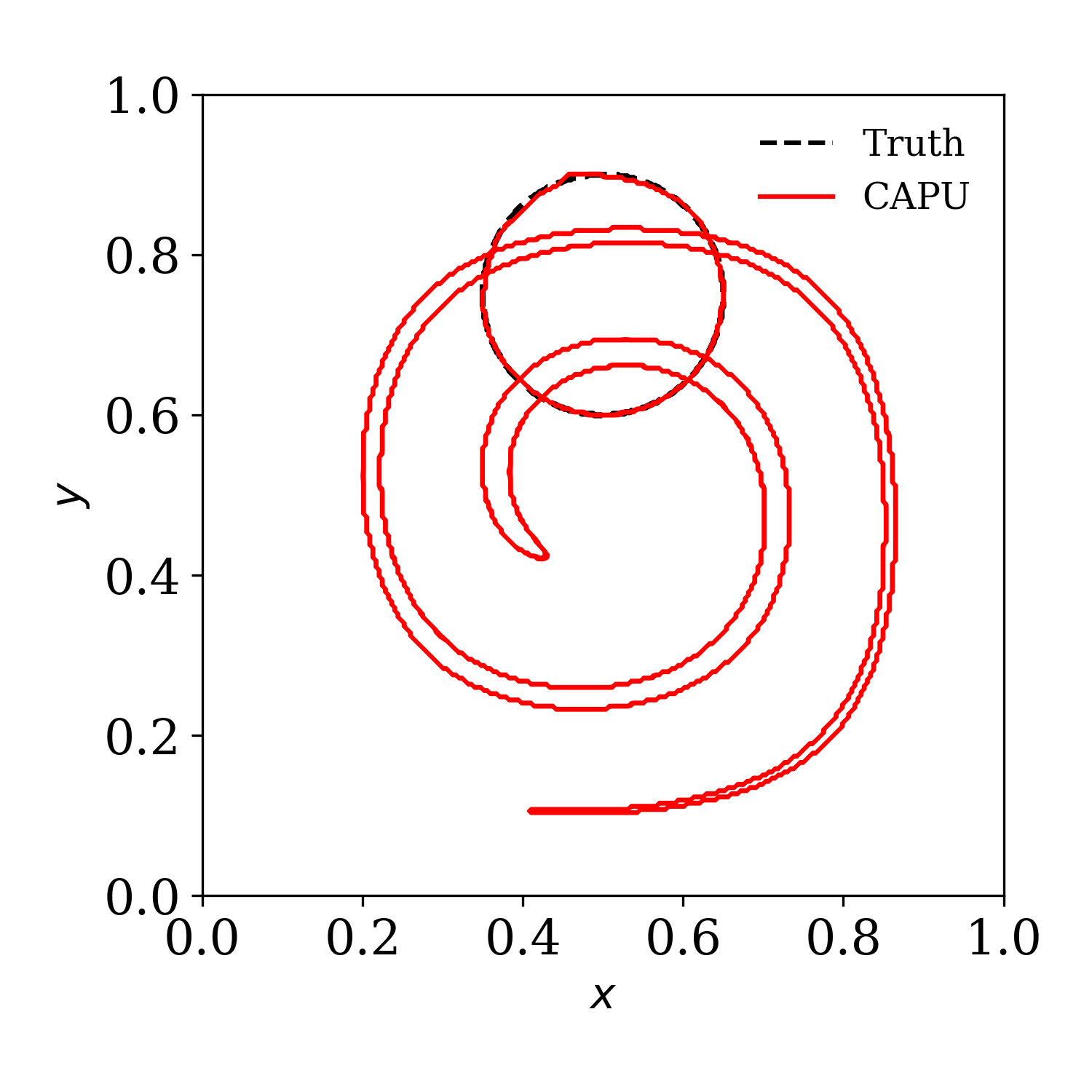}}
    \caption{Reversible transport of a passive scalar field by a vortex: Predicted interfaces at intermediate and final times, represented by the iso-contours $\phi = 0$ to 1 in increments of $0.2$, using the CAPU algorithm with (a) the coarse mesh and (b) the fine mesh.}
    \label{fig:vortex_flow_t8_mid_end_pred}
\end{figure}

By plotting scalar contour lines at $\phi = 0$ to $1$ in increments of 0.2, Figure~\ref{fig:vortex_flow_t8_mid_end_pred} presents the predicted interface locations for both coarse (panel a) and fine (panel b) meshes at midpoint ($t=4$) and at final time ($t=8$).
In panel (a), the CAPU algorithm consistently captures the interface over the entire simulation, even on the coarse mesh. At $t=4$, the prediction exhibits a tightly wound spiral, while at $t=8$, it evolves into an irregular pattern around the original circular region, indicating degradation of scalar information.

Although the predicted scalar field on the coarse mesh yields acceptable results, it exhibits noticeable imperfections in reconstructing the original circular geometry. In contrast, simulations performed on the fine mesh demonstrate significant improvements: the solution evolves into a smooth and coherent spiral structure at the midpoint and accurately recovers an almost perfectly circular profile by the final time.

\begin{table}[!h]
\centering
\footnotesize
\begin{tabular}{l|r}
\hline
Algorithm & Error \\ \hline
Rider and Kothe \citep{rider1998singlevortex}  & $1.44 \times 10^{-3}$ \\
EMFPA/Youngs \citep{lopez2004singlevortex}  & $2.13 \times 10^{-3}$  \\
Stream/Youngs \citep{harvie2000singlevortex}  & $2.16 \times 10^{-3}$  \\
CLSVOF \citep{singh2018singlevortex} & $1.93 \times 10^{-3}$  \\
THINC/QQ \citep{xie2017singlevortex} & $3.12 \times 10^{-3}$ \\
CBLSVOF \citep{xie_consistent_2020} & $2.73 \times 10^{-3}$ \\
Mean of various works from \citep{xie_consistent_2020} & $2.255  \pm 1.167 \times 10^{-3}$ \\
\textbf{Current Method} & \boldsymbol{$1.113 \pm 0.210 \times 10^{-3}$} \end{tabular}
\caption{Reversible transport of a passive scalar field by a vortex: numerical errors at $T=8$ with different numerical methods and our proposed method}
\label{tab:single_vortex_flow_errors}
\end{table}

Next, we compare the quality of our predictions for this challenging benchmark problems against published studies that implement a variety of interface tracking methods using finite difference/volume type of methods. \citet{xie_consistent_2020} compiled results from 17 different studies including structured and unstructured grids. The following metric was used to assess the accuracy of each method  
\begin{equation}
\text{Error} = \sum_{i=1}^{N_e} |\phi_{pi} - \phi_{ei}| \Omega_i,
\end{equation}
where \(\phi_{pi}\) and \(\phi_{ei}\) denote the predicted and exact values on grid cell $i$, and \(\Omega_i\) represents the area of that cell. Error values range from a worst case of $5.02\times10^{-3}$ to a best case of $5.09\times10^{-4}$ for the present problem.

Table~\ref{tab:single_vortex_flow_errors} presents a comparison of our method’s mean error estimate and its standard deviation--computed over five trials--at the final time ($t = 8.0$) against selected benchmark studies, including the average of the error from 17 different methods as compiled in \cite{xie_consistent_2020}. Our approach achieves a mean error of $1.113 \times 10^{-3}$, underscoring its competitive accuracy relative to established interface tracking methods.

\subsection{Inverse problem: estimation of space-wise heat source in transient heat conduction}\label{sec:heat_source_reconstruction}
In Section~\ref{subsec:inverse_formulation}, we extended the PECANN-CAPU framework with constraint aggregation to inverse PDE problems. Here, we apply this formulation to jointly learn the temperature field and a spatially dependent heat source from noisy temperature measurements. Specifically, we consider the inverse identification of a space-wise heat source term problem from \citet{hasanov2012identification}, in which the temporal component of the source is known and aim to infer its spatial part. 
\begin{figure}[!h]
    \centering
    \includegraphics[width=0.8\linewidth]{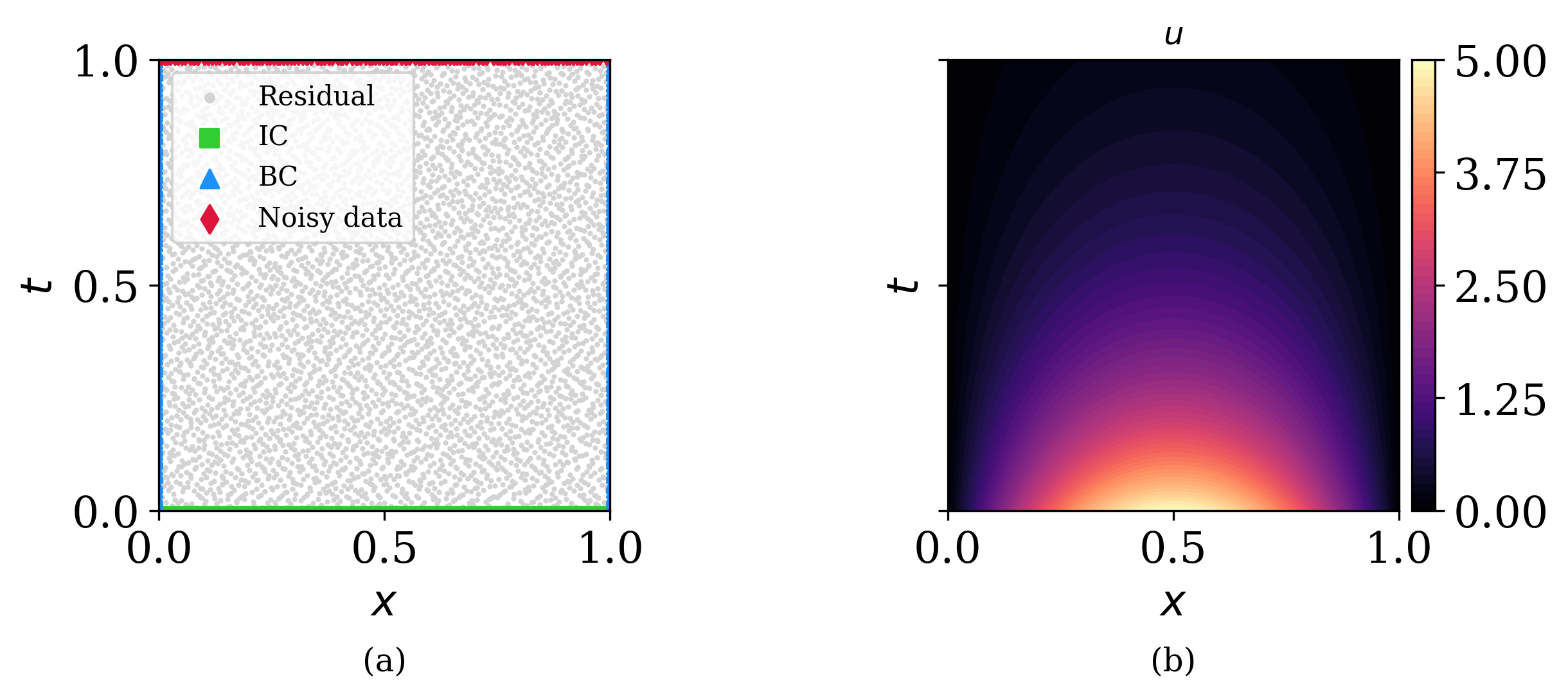}
    \caption{Inverse problem: (a) distribution of residual, boundary, and initial points along with noisy data from the final-time measurement; (b) exact, time-dependent temperature field, $u(x, t)$.}
    \label{fig:inv_ispf_sampling_solution}
\end{figure}

The governing equation \eqref{eq:heat_pde} defined in the domain $\mathcal{U} = \{(x,t)| 0\leq x,t\leq 1\}$ is adopted with the following separable source term,
\begin{equation}
    s(x,t) = F(x)H(t),
\end{equation}
where $H(t) = 5\exp(-3t)$ is prescribed and the thermal conductivity is given by $\kappa (x) = 4 + x^2$. This type of inverse problem is designated as ISPF in \cite{hasanov2012identification}. In addition to known boundary and initial conditions, we incorporate final-time temperature measurements corrupted by Gaussian noise:
\begin{equation}
    u_T(x) = (1 + n)\, u(x, T_f), \quad n \sim \mathcal{N}(0, 1).
\end{equation}
The exact solution and the corresponding source are as follows:
\begin{equation}
    \begin{aligned}
        u(x,t) & = 5\exp(-3t)\sin(\pi x), \\
        F(x) & = (-3 + \pi^2(4 + x^2))\sin(\pi x) - 2\pi x.
    \end{aligned}
\end{equation}
Figure~\ref{fig:inv_ispf_sampling_solution}(a) indicates the problem setup via the sampling configuration, which includes the known boundary and initial points, noisy measurements, and residual points representing the PDE.
The corresponding exact, time-dependent temperature field is shown in Fig.~\ref{fig:inv_ispf_sampling_solution}(b).

We cast this inverse problem as a PDE-constrained optimization problem using a neural network with inputs $(x, t)$, one hidden layer of 40 units, and outputs $(u, F)$. To ensure that $F$ depends only on $x$, we evaluate $F$ with a constant temporal input, effectively removing any influence from $t$. This is implemented as:
\begin{equation}
\begin{aligned}
    u(x,t),\, \_ & = f_{\theta}(x, t), \\
    \_\, ,\, F(x) & = f_{\theta}(x, t_{\mathrm{const}}),
\end{aligned}
\end{equation}
where $\_$ denotes a placeholder output that is computed but discarded, $t_{\mathrm{const}}$ is a fixed value as 0.
Thus, the network performs two forward passes: first, using the true $(x, t)$ to obtain $u$; second, using $(x, t_{\mathrm{const}})$ to obtain $F$. 
The loss formulation follows Eq.~\eqref{eq:inv_formulation}, where the averaging measurement loss is treated as the objective $\mathcal{J}$, and the PDE residual $\mathcal{C}_F$, boundary $\mathcal{C}_B$, and initial condition $\mathcal{C}_I$ losses are enforced as constraints. 
Training samples include $10^4$ collocation points for the PDE residual, 128 points for the initial condition, each boundary, and the final-time measurement. The model is trained for $10^4$ epochs using the L-BFGS optimizer, with the CAPU algorithm applied. Penalty scaling factors are set to $\eta_F = 0.01$ for $\mathcal{C}_F$ and 1 for the remaining constraints, as detailed in subsection~\ref{subsec:inverse_formulation}.

\begin{figure}[!h]
\centering
    \subfloat[]{\includegraphics[width=0.32\textwidth]{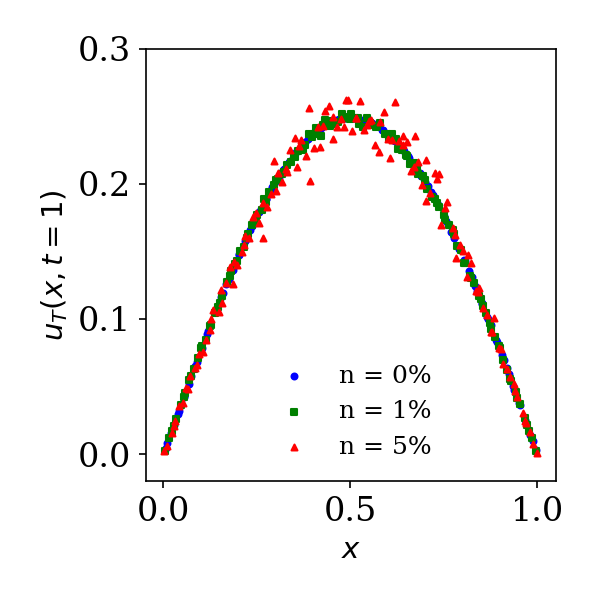}}\quad
    \subfloat[]{\includegraphics[width=0.32\textwidth]{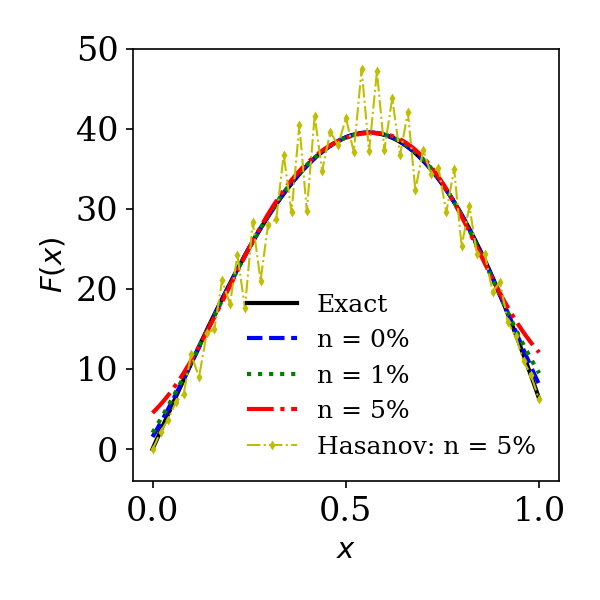}}\quad
    \subfloat[]{\includegraphics[width=0.32\textwidth]{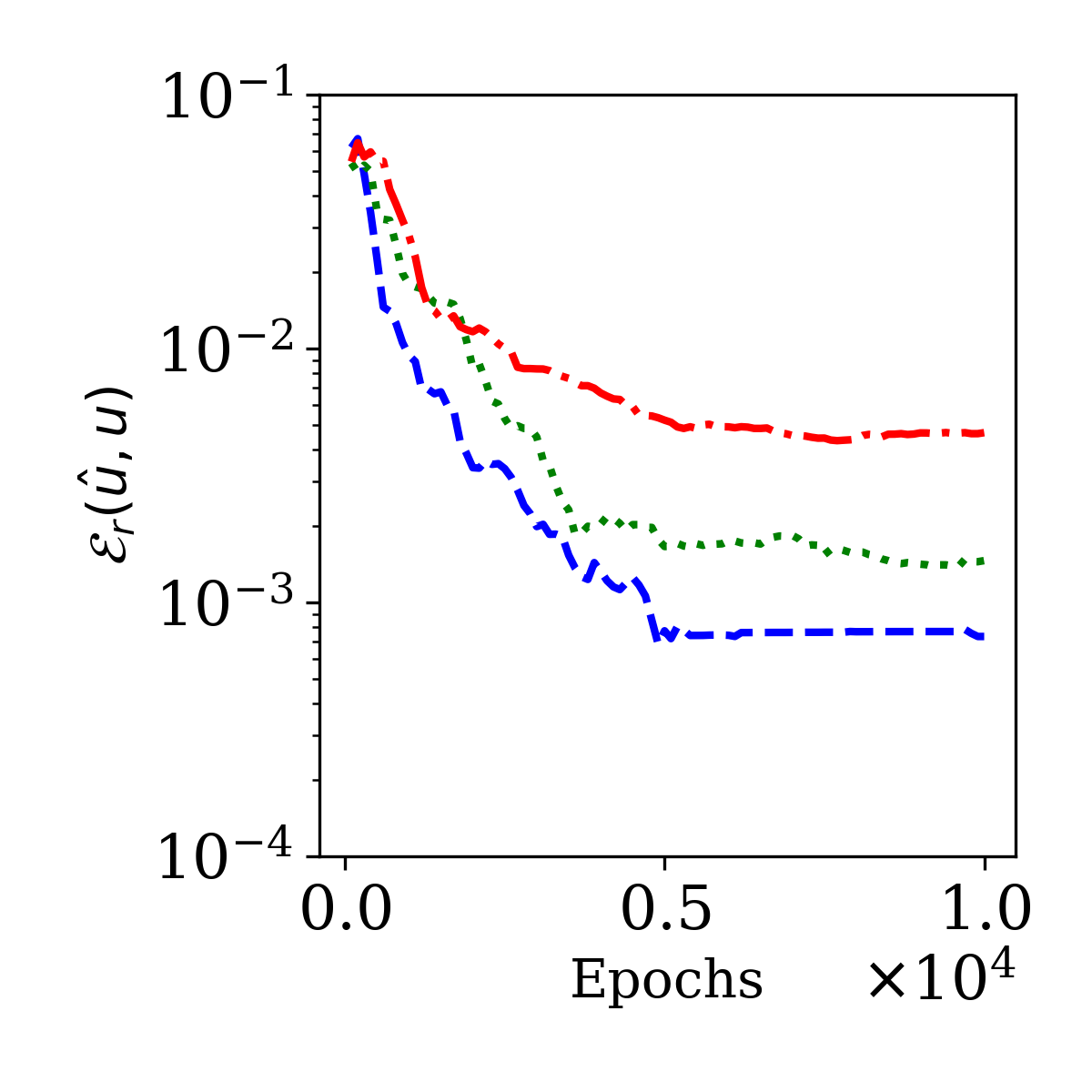}}
\caption{Inverse problem: (a) temperature measurements at the final time with different noise levels, (b) predicted spatial source $F(x)$ compared with the exact solution and numerical results from \citet{hasanov2012identification}, and (c) evolution of the relative $l^2$ error in the predicted temperature over training.}
\label{fig:inv_ispf_pred}
\end{figure}

Figure~\ref{fig:inv_ispf_pred} illustrates the effect of final-time measurements $u_T(x)$ at three noise levels ($n = 0$, $1\%$, $5\%$) on the predicted spatial source $F(x)$, compared against the exact solution and numerical results from \citet{hasanov2012identification}, along with the evolution of the relative $l^2$ error $\mathcal{E}_r(\hat{u}, u)$ during training. As noise increases, the measurements in panel (a) scatter around the noiseless data. In panel (b), the network-based prediction of $F(x)$ remains smooth and accurate in the central region but shows increasing deviation near the boundaries. In contrast, the numerical result of \citet{hasanov2012identification} exhibits noticeable oscillations near the center, despite accurate boundary predictions. Panel (c) illustrates that the relative error, $\mathcal{E}_r(\hat{u}, u)$, consistently falls below $10^{-2}$ across all noise levels, albeit with a gradual degradation in performance as the noise increases. This example highlights a key advantage of the PECANN-CAPU framework over traditional approaches to inverse problems--namely, its robustness in producing smooth and accurate predictions even under high observational noise.

\section{Conclusion}\label{sec:Conclusion}

The PECANN framework \cite{PECANN_2022} relies on constrained optimization via the augmented Lagrangian method (ALM) \cite{hestenes1969multiplier, powell1969method} to enforce boundary and initial conditions on PDE solutions as equality constraints. We showed that conventional ALM with a single penalty parameter update strategy often fail for complex problems with multiple, diverse constraints. We addressed this limitation by introducing the ``conditionally adaptive penalty update (CAPU)'' strategy, which assigns unique penalty parameters to each constraint and updates them based on the magnitude of constraint violations. Theoretical analysis and empirical examples confirm CAPU's ability to ensure boundedness of penalty parameters and convergence of Lagrange multipliers.

To further expand the PECANN–CAPU framework's capacity for complex PDE problems in forward and inverse modeling scenarios, we incorporated constraint aggregation---a common technique in design and topology optimization---into its formulation. The original PECANN approach enforced constraints point-wise at every collocation point, a method that becomes computationally cumbersome and degrades performance as problem complexity increases, especially when using the proposed CAPU algorithm. Our experiments revealed that the Lagrange multipliers associated with a given constraint type follow a well-defined distribution and mean. Motivated by this observation, we aggregated constraints using a mean squared residual metric, which drastically reduced the number of Lagrange multipliers and penalty parameters from thousands to just a few. Equally important, we demonstrated through numerical examples and theoretical analysis that constraint aggregation enhances accuracy by eliminating the additional variance term in the objective function that arises from the point-wise enforcement of a large number of constraints.

Our revised formulation significantly improves the expressive power of the original PECANN framework. Standard MLPs often struggle to represent highly oscillatory PDE solutions with multi-scale features. Through case studies involving 1D Poisson's and 2D Helmholtz equations of varying complexity, we demonstrated that PECANN–CAPU with constraint aggregation substantially enhances MLP capacity. Our approach achieves accuracy on par with physics-informed methods based on Kolmogorov–Arnold networks. An interesting implication of this demonstration is that pure regression is insufficient to evaluate network architecture in physics informed learning. Recognizing the inherent limitations of MLPs, we further enhanced expressive power by incorporating a single Fourier feature mapping with $\sigma=1$. Another key strength of PECANN–CAPU lies in training efficiency. In our experiments, shallow and compact networks trained with L-BFGS consistently outperformed deeper networks trained with Adam.

To assess the PECANN–CAPU framework's capability in predicting long-time PDE evolution, we applied it to the challenging benchmark of reversible transport of a passive scalar by a single vortex, a problem often used to evaluate interface-tracking methods in two-phase flow simulations. To enable this long-time integration, we introduced a time-windowing strategy within the constrained optimization formulation. In this approach, the terminal state of each time window is enforced as an initial constraint for the subsequent one. This strategy is uniquely enabled by the CAPU design, which inherently supports multiple independent constraints, each evolving with its own adaptively updated penalty parameter. Our results show that PECANN–CAPU achieves accuracy comparable to advanced schemes developed specifically for two-phase flows.


Overall, the success of PECANN–CAPU with constraint aggregation across a diverse set of PDE problems—in both forward and inverse scenarios—demonstrates that the introduced enhancements substantially expand the scope of the original PECANN formulation \cite{PECANN_2022} for tackling challenging problems in computational science. More broadly, the augmented Lagrangian method, when integrated with CAPU and constraint aggregation, provides a principled and versatile mechanism for handling multiple, heterogeneous constraints, highlighting its promise beyond training physics-informed neural networks. 

\section*{CRediT authorship contribution statement}
\textbf{Qifeng Hu:} Conceptualization (Supporting), Data curation, Formal analysis (Lead), Investigation (Lead), Methodology (Equal), Software (Lead), Validation, Visualization, Writing - original draft (Lead), Writing – review \& editing (Equal) 
\textbf{Shamsulhaq Basir:} Conceptualization (Supporting), Investigation (Supporting), Methodology (Equal), Software - original version, Writing – original draft (Supporting), Writing – review \& editing (Supporting)
\textbf{Inanc Senocak:} Conceptualization (Lead), Formal analysis (Supporting), Funding acquisition, Investigation (Supporting), Methodology (Equal), Project administration, Resources, Supervision, Writing – original draft (Supporting), Writing – review \& editing (Equal).

\section*{Data availability}
All the codes used to produce the results in this work will be publicly available at \url{https://github.com/HiPerSimLab/PECANN/CAPU} upon acceptance of the manuscript.

\section*{Declaration of computing interests}
The authors declare that they have no known competing financial interests or personal relationships that could have appeared to influence the work reported in this paper.

\section*{Declaration of generative AI and AI-assisted technologies in the writing }
During the preparation of this work the author(s) used Microsoft Copilot and ChatGPT in order to assist with improving the clarity and quality of the English language. After using this tool/service, the author(s) reviewed and edited the content as needed and take(s) full responsibility for the content of the publication.

\section*{Acknowledgments}
This material is based upon work supported by the National Science Foundation under Grant No. 1953204 and in part by the University of Pittsburgh Center for Research Computing through the resources provided.


\appendix

\section{Evaluation of PECANN-CAPU and Direct PECANN-RMSProp}\label{sec:appendixa}
This appendix provides a comparison between the adaptive strategy from RMSProp and our proposed CAPU to better comprehend the merits of CAPU. Both strategies adopt the PECANN formulation with constraint aggregation.

We consider the 2D Helmholtz problem, Eq.~\eqref{eq:helmholtz_eqn}, defined by the source terms in Eqs.~\eqref{eq:helm_multi_source} and~\eqref{eq:helm_multi_para} with $L = 4$. Note that the more complex $L=5$ case is studied in Section \ref{sec:helmholtz_challenging}.
A compact MLP with three hidden layers of 20 neurons each is trained for 80{,}000 epochs using the L-BFGS optimizer. The training data consist of 6400 randomly sampled residual points and 80 boundary points per edge.

\begin{figure}[!h]
\centering
    \subfloat[]{\includegraphics[width=0.48\textwidth]{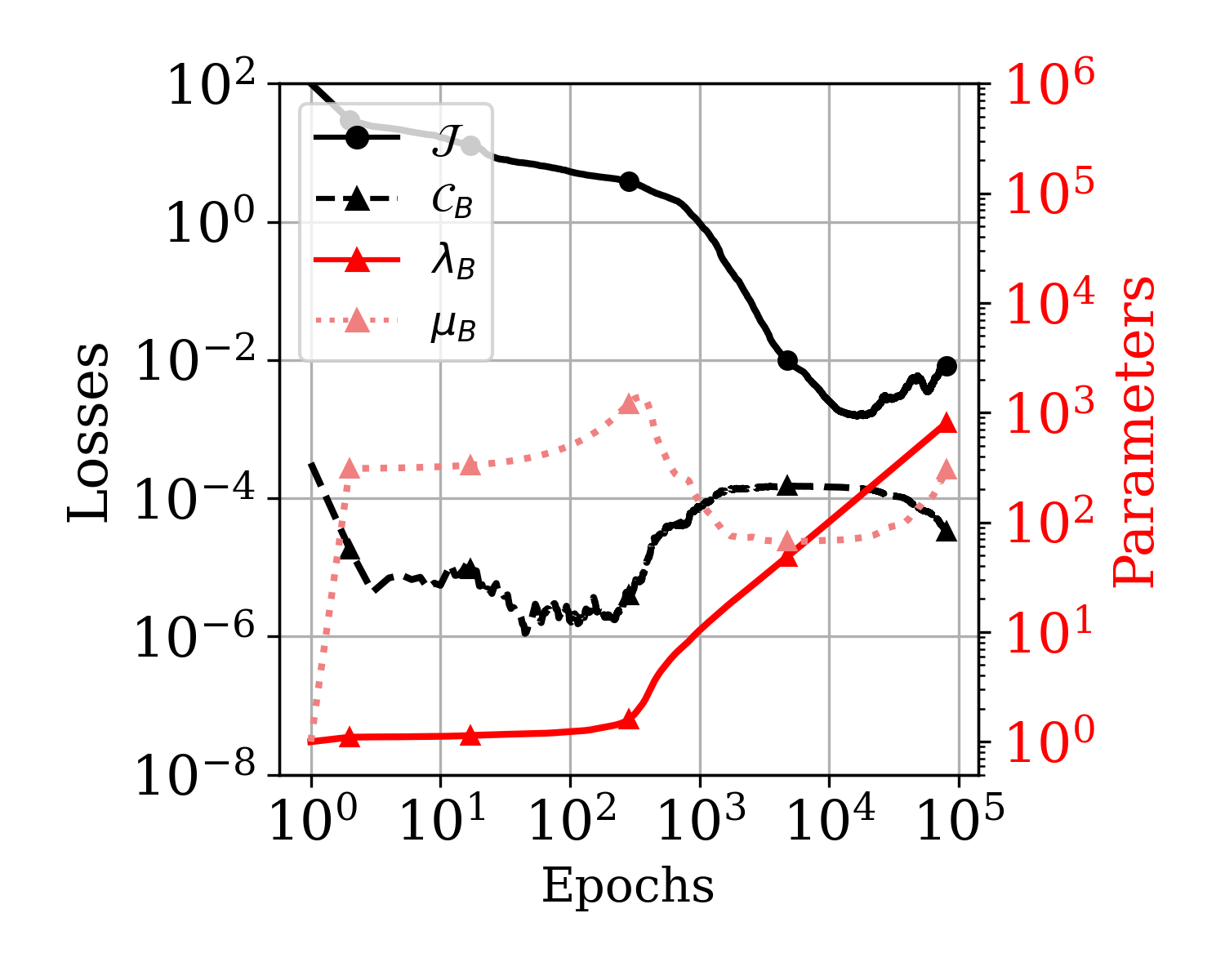}}\quad
    \subfloat[]{\includegraphics[width=0.48\textwidth]{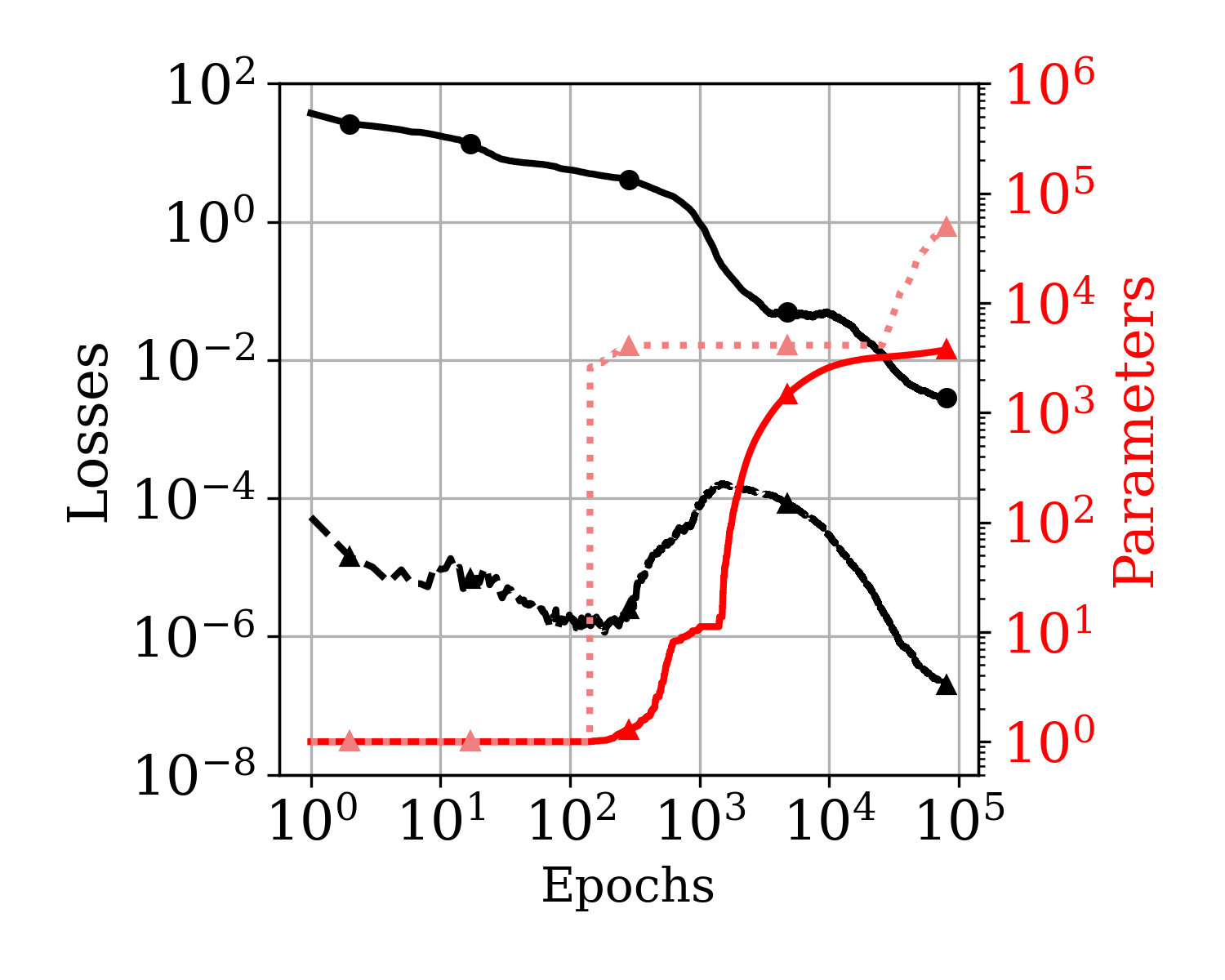}}\quad
    \
    \subfloat[]{\includegraphics[width=0.45\textwidth]{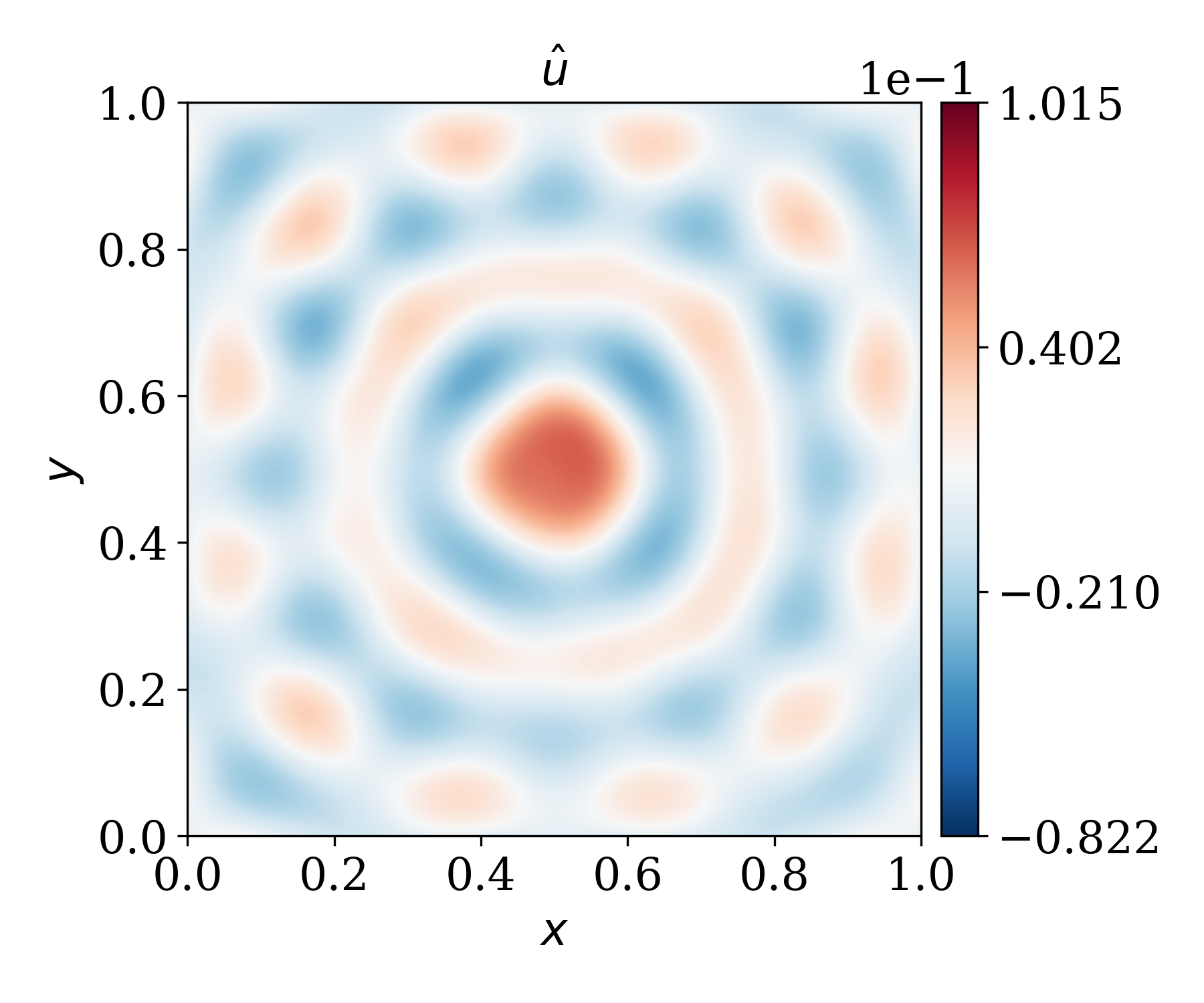}}\quad
    \subfloat[]{\includegraphics[width=0.45\textwidth]{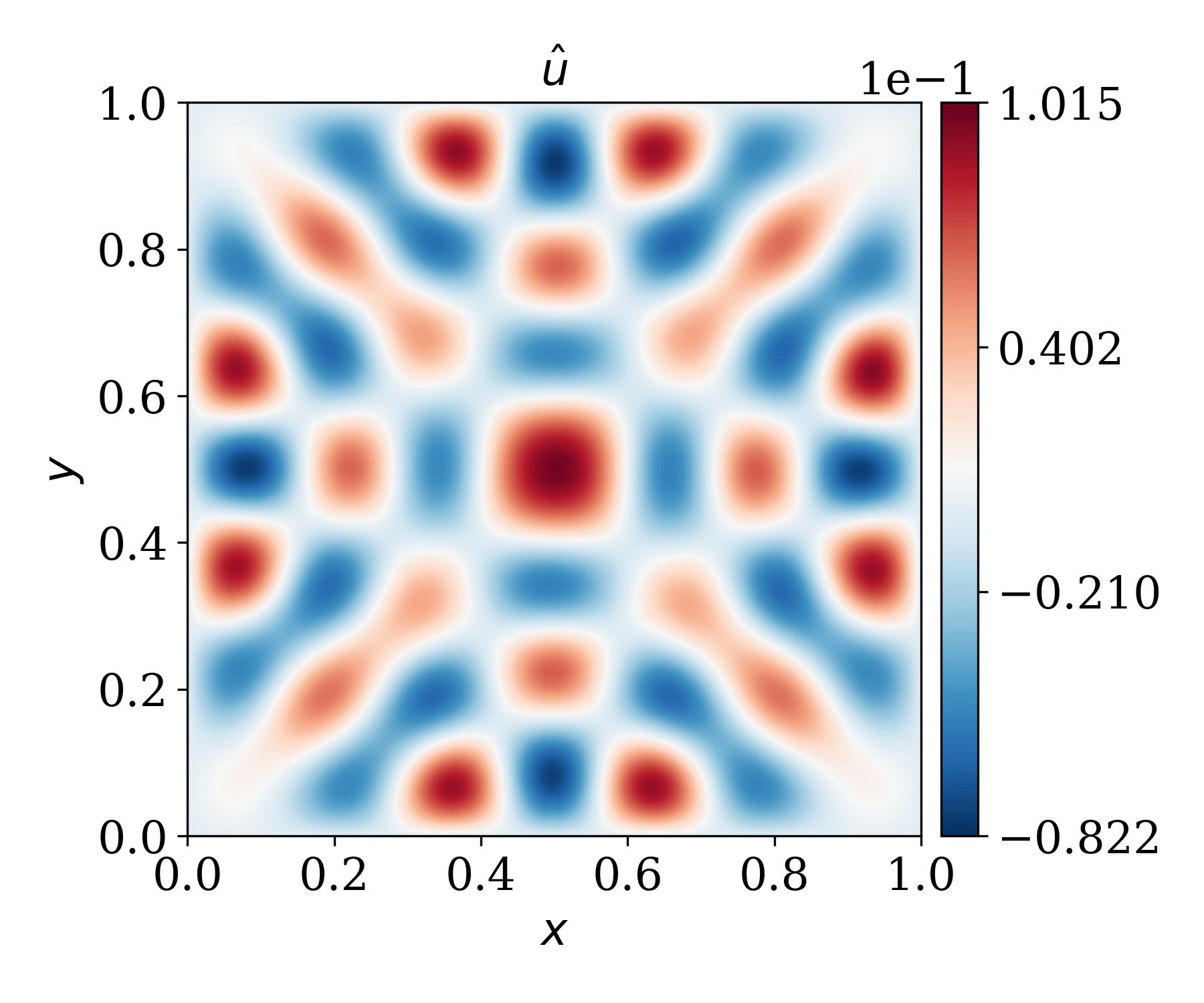}}\quad
    \caption{2D Helmholtz equation with a localized source term (Eq.~\ref{eq:helm_multi_source}) with $L = 4$: evolution of the objective, boundary constraint, Lagrange multiplier, and penalty parameter using (a) the PECANN-RMSProp algorithm, (b) the proposed PECANN-CAPU algorithm. Panels (c–d) show the corresponding predicted fields of these algorithms at their final epochs.}
    \label{fig:helm_multi_l4_pred_evol}
\end{figure}

Figure~\ref{fig:helm_multi_l4_pred_evol} show the evolution of the objective $\mathcal{J}$, boundary constraint $\mathcal{C}_B$, penalty parameter $\mu_B$, and Lagrange multiplier $\lambda_B$ over the epochs, for PECANN-RMSProp and PECANN-CAPU, respectively, with the corresponding final predictions.

During optimization, as shown in Figs.~\ref{fig:helm_multi_l4_pred_evol}(a–b), the constraint violation $\mathcal{C}_B$ starts to increase after approximately $10^2$ epochs, as the objective $\mathcal{J}$ continues to decrease, indicating that the training is progressively driven towards an infeasible region.
In this scenario, RMSProp in Fig.~\ref{fig:helm_multi_l4_pred_evol}(a) promotes a reduction of $\mu_B$ and produces an almost linear, weakly responsive update of $\lambda_B$, as discussed in Section~\ref{sec:proposed_capu}. 
Lagrange multipliers show no sign of convergence even after $10^5$ epochs. 
We also observe premature escalation of $\mu_B$ and unnecessary updates of $\lambda_B$ during the early training stage, when optimization automatically continues to improve feasibility.

In contrast, PECANN-CAPU in Fig.~\ref{fig:helm_multi_l4_pred_evol}(b) exhibits stronger enforcement response under larger constraint violations by maintaining $\mu_B$ above $10^3$, determined by the maximum operation in Eq.~\eqref{eq:capu_mu}. L
It is accompanied by a pronounced increase in $\lambda_B$ compared to other training stages. Consequently, $\mathcal{C}_B$ begins decreasing shortly after $10^3$ epochs, indicating that the optimization is driven back toward feasibility.
As $\mathcal{C}_B$ continues to diminish,Lagrange multiplier, $\lambda_B$, converges to $\approx 2{,}000$ after $\approx 5\times 10^4$ epochs.

The final predictions shown in Figs.~\ref{fig:helm_multi_l4_pred_evol}(c–d) further highlight the difference between feasible and infeasible solutions produced by the two algorithms.
The Gaussian-like source located at the geometric center of the domain induces outward wave propagation, leading to pronounced boundary violations around the $10^3$-epoch mark.
As illustrated in Fig.~\ref{fig:helm_multi_l4_pred_evol}(c), PECANN-RMSProp just begins to recover toward feasibility, suggesting that it would need order of magnitude more epochs than PECANN-CAPU to potentially achieve a feasible solution in panel (d).
This is evident from the evolution plots displayed on a $\log{10}$ scale along the x-axis.
Eventually, the CAPU algorithm achieves a relative $l^2$ norm of $\mathcal{E}_r = (8.906 \pm 5.189) \times 10^{-2}$, with the best-performing trial reaching $2.909 \times 10^{-2}$, confirming the efficacy of the CAPU strategy.

\section{Sensitivity to Hyperparameter Selection}\label{sec:appendixb}
We use the exact solution of the Helmholtz equation in Eq.~\eqref{eq:helmholtz_sol} as a reference case. To examine the sensitivity of PECANN-CAPU predictions to its hyperparameters, we adopt the configuration $a_1 = 1$ and $a_2 = 4$, following the setup in~\cite{SHUKLA2024117290}.

The sole external input to Algorithm~\ref{alg:capu_adaptive_training_algorithm}, the penalty scaling factor $\eta_B$ for BC, is varied from $0.01$ (as used in RMSProp) to $1$.
We also vary the internal hyperparameters: the smoothing coefficient $\zeta$ (from the default value of $0.99$ to $0.9$) and the conditional factor $\omega_t$ (from $0.999$ to $0.99$).
We adopt a shallow MLP with two hidden layers ($H = 2$) and 16 neurons per layer ($W = 16$) and train it using the L-BFGS optimizer with a limit of 1800 epochs.
The same number of collocation and evaluation points are used as described in Section~\ref{sec:2d_helm}.

Table~\ref{tab:helm_pikan_case_a_capu_l2_hyper} reports the relative $l^2$ norm of the mean error and its standard deviation computed over five trials for the PECANN-CAPU method.
The accuracy of PECANN-CAPU predictions shows a modest sensitivity to $\eta_B$, which controls the growth of the Lagrange multiplier. Setting $\eta_B = 1$ reduces both the mean and standard deviation of the relative error compared to $\eta_B = 0.01$ by approximately an order of magnitude. In contrast, variations in internal hyperparameters $\omega_t$ and $\zeta$ have only minor effects. 

While the best CAPU performance is achieved with $\eta_B = 1$ and a tuned value of $\zeta = 0.9$, we caution against excessive tuning of internal hyperparameters.
Reducing the smoothing coefficient to $\zeta = 0.9$ enhances responsiveness by assigning greater weight to recent changes in the squared averages. 
While this adjustment yields favorable results in the present simple case, where constraint violations decrease steadily during training, such aggressive responsiveness may neglect gradual trends, leading to undesirable behavior in more complex scenarios, such as the cases discussed in~\ref{sec:appendixa} and Section~\ref{sec:trans_rare}. Therefore, we recommend retaining the default values for the internal hyperparameters, and selecting the external input--the penalty scaling factors for BC and IC--as $1.0$ when using the L-BFGS optimizer, as analyzed in Section~\ref{sec:proposed_capu}.

\begin{table}[]
    \centering
    \footnotesize
    \begin{tabular}{l|c|c|c|c|r|r}
        \hline
        \multirow{2}{*}{Method} & \multirow{2}{*}{$\eta_B$} & \multirow{2}{*}{$\omega_t$} & \multirow{2}{*}{$\zeta$} & \multirow{2}{*}{Epochs} & \multicolumn{2}{c}{Relative $l^2$} \\ \cline{6-7}
            &   &   &   &  & mean $\pm$ std & best trial \\
        \hline
        \multirow{7}{*}{PECANN-CAPU}   & \multirow{3}{*}{0.01} & $d.v.$ & $d.v.$ & \multirow{6}{*}{1800} & $3.40 \pm 1.28\times10^{-2}$ & $1.74\times10^{-2}$ \\ \cline{3-4}
                                        &   & 0.99 & $d.v.$ &  & $1.75 \pm 0.63\times10^{-2}$ & $7.56\times10^{-3}$ \\  
                                        &  & $d.v.$ & 0.9 &   & $2.84\pm 0.90\times10^{-2}$ & $1.29\times10^{-2}$ \\  
                                        \cline{2-4} \cline{6-7}
                                        & \multirow{3}{*}{1}  & $d.v.$ & $d.v.$ &   & $8.20\pm 4.20\times10^{-3}$ & $4.22\times10^{-3}$ \\ \cline{3-4}
                                        &    & 0.99  & $d.v.$ &   & $8.60\pm 5.97\times10^{-3}$ & $4.18\times10^{-3}$ \\ 
                                        &    & $d.v.$  & 0.9 &   & $5.14\pm 2.88\times10^{-3}$ & $2.68\times10^{-3}$ \\ 
    \end{tabular}
    \caption{Helmholtz equation (Eq. \ref{eq:helmholtz_sol} with $a_1 = 1$ and $a_2 = 4$): Relative $l^2$ error comparison of PECANN-CAPU using one input and two internal hyperparameters, where the default values ($\omega_t = 0.999$, $\zeta = 0.99$).} are denoted with $d.v.$. The network has two hidden layers with 16 neurons per layer ($H=2, W=16)$),
    \label{tab:helm_pikan_case_a_capu_l2_hyper}
\end{table}

\begin{figure}[!h]
\centering
    \subfloat[]{\includegraphics[width=0.48\textwidth]{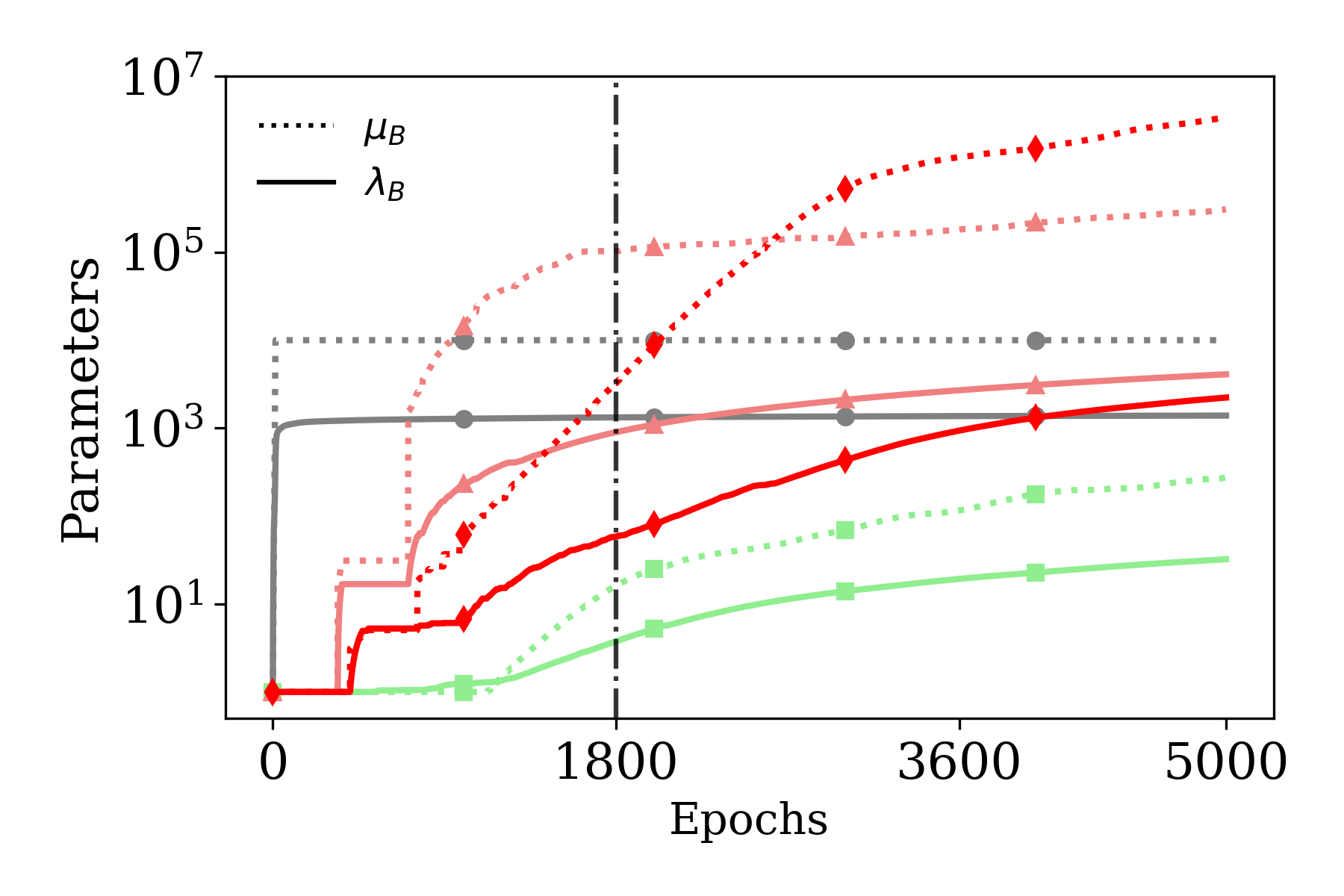}}\quad
    \subfloat[]{\includegraphics[width=0.48\textwidth]{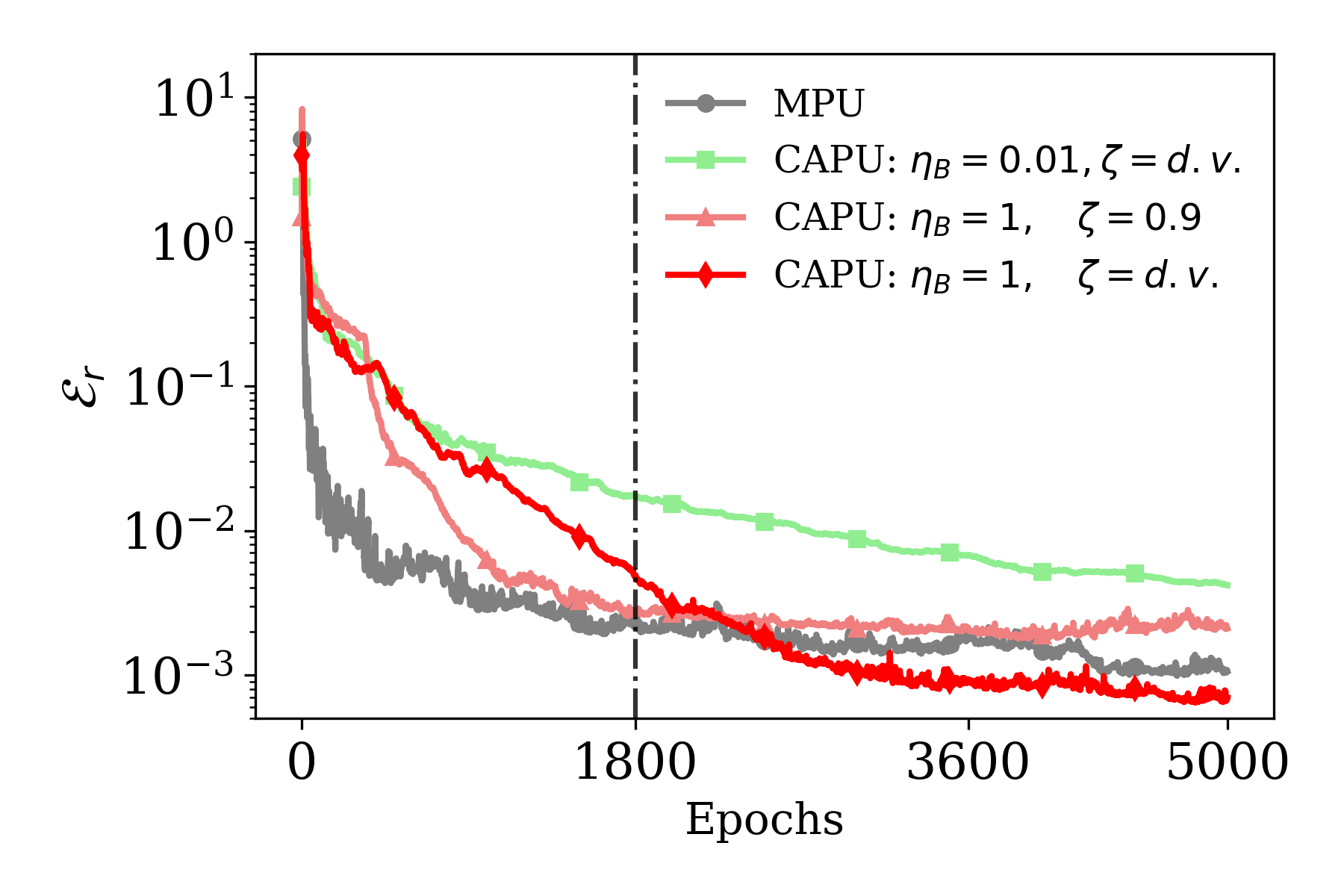}}
    \caption{Helmholtz equation (Eq. \ref{eq:helmholtz_sol} with $a_1 = 1$ and $a_2 = 4$): (a) Evolution of penalty parameters (dotted lines) for BC and the corresponding Lagrange multipliers (solid lines) of the best-performing trials using three CAPU configuration and MPU, extended to 5000 epochs. and (b) the corresponding evolution of the relative $l^2$ norm.}
    \label{fig:helm_pikan_case_a_evolution}
\end{figure}

To assess the influence of CAPU hyperparameters on convergence, three configurations, ($\eta_B = 0.01$, $\zeta=d.v.$), ($\eta_B = 1.0$, $\zeta=d.v.$), and ($\eta_B = 1.0$, $\zeta = 0.9$), are evaluated along with the MPU algorithm.
Figure~\ref{fig:helm_pikan_case_a_evolution} illustrates the evolution of the penalty parameter $\mu_B$, the corresponding Lagrange multiplier $\lambda_B$, and the relative $l^2$ error $\mathcal{E}_r$ for their best-performing trials.
As shown in Fig.~\ref{fig:helm_pikan_case_a_evolution}(a), all CAPU configurations exhibit a delayed increase in $\mu_B$ and $\lambda_B$ during the first 500 epochs compared to MPU, due to the convergence condition embedded in the CAPU algorithm.
Beyond this stage, configurations with larger penalty scaling factor $\eta_B$ begin to grow earlier, while the one with a smaller smoothing coefficient $\zeta$ rises more rapidly.
As discussed previously, this behavior associated with smaller $\zeta$ can only occur when the optimization continues toward the enhancement of feasibility.

Before the 1800th epoch, MPU shows a superior convergence performance in terms of $\mathcal{E}_r$ in Fig.~\ref{fig:helm_pikan_case_a_evolution}(b).
In fact, the ranking of configurations at this point closely follows the order of their Lagrange multipliers in Fig.~\ref{fig:helm_pikan_case_a_evolution}(a).
MPU appears to converge fastest early on because the constraint decreases to a negligible level, leading $\lambda_B$ to rapidly converge to about $10^3$ as $\mu_B$ reaches its upper bound at a fixed rate.
In contrast, CAPU resists such abrupt growth through its adaptive mechanism--beneficial for complex problems but limiting here--resulting in slower convergence.
With further training, CAPU with $\eta_B = 1$ gradually increases both $\mu_B$ and $\lambda_B$ until convergence near $\lambda_B \approx 10^3$, whereas the $\eta_B = 0.01$ configuration requires additional computation to reach convergence. Overall, CAPU achieves the lowest observed relative error.

\begin{table}[]
    \centering
    \footnotesize
    \begin{tabular}{l|c|c|c|c|c|r|r}
        \hline
        \multirow{2}{*}{Method} & \multirow{2}{*}{N. Params} & \multirow{2}{*}{$H$} & \multirow{2}{*}{$W$} & \multirow{2}{*}{Optim.} & \multirow{2}{*}{Epochs} & \multicolumn{2}{c}{Relative $l^2$} \\ \cline{7-8}
            &   &   &   &  &  & mean $\pm$ std & best trial\\ \hline
        baseline PINN \cite{SHUKLA2024117290}  & 337 & 2 & 16 & L-BFGS & 1800 & - & $1.03\times10^{-2}$ \\ 
        cPIKAN+RBA \cite{SHUKLA2024117290} & 350 & 2 & 8 & L-BFGS & 1800 &  - & $3.81\times10^{-3}$ \\  
        \multirow{2}{*}{SA-PINN \cite{mcclenny2020self}} & \multirow{2}{*}{7851} & \multirow{2}{*}{4} & \multirow{2}{*}{50} & \scriptsize{Adam \&} & \scriptsize{$10^4$ \&} & \multirow{2}{*}{$3.20 \pm 0.22 \times 10^{-3}$}  & \multirow{2}{*}{-} \\
          &   &   &   & \scriptsize{L-BFGS} & \scriptsize{$10^4$} &    &  \\
        \hline
        PECANN-MPU & \multirow{3}{*}{337} & \multirow{3}{*}{2} & \multirow{3}{*}{16} & \multirow{3}{*}{L-BFGS} & \multirow{3}{*}{5000} & $5.33 \pm 2.52 \times 10^{-3}$ & $1.06\times10^{-3}$  \\ 
        PECANN-CPU &   &   &   &  &  & $6.18 \pm 4.08 \times 10^{-3}$ & $4.32\times 10^{-3}$ \\ 
        PECANN-CAPU &   &   &   &  &  &  \boldsymbol{$4.88\pm 6.65\times10^{-3}$} & \boldsymbol{$6.96\times10^{-4}$} \\
        \hline
        baseline PINN & \multirow{2}{*}{337} & \multirow{2}{*}{2} & \multirow{2}{*}{16} & \multirow{2}{*}{L-BFGS} & \multirow{2}{*}{5000} & $5.64 \pm 5.47 \times 10^{-2}$ & $1.39 \times 10^{-2}$ \\
        PINN with $\lambda_B = 1000$ &  &  &  &  &  & \boldsymbol{$2.56 \pm 1.59 \times 10^{-3}$} & \boldsymbol{$5.87 \times 10^{-4}$}
    \end{tabular}
    \caption{Helmholtz equation (Eq. \ref{eq:helmholtz_sol} with $a_1 = 1$ and $a_2 = 4$): Relative $l^2$ comparison among several methods and the PECANN-CAPU algorithm. The SA-PINN \cite{mcclenny2020self} was trained with $10^5$ residual points and 400 boundary points. PINNs are re-evaluated over 5000 training epochs, both with and without the near-optimal Lagrange multiplier $\lambda_B = 1000$ retrieved from the PECANN CAPU run.}
    \label{tab:helm_pikan_case_a_l2_comp}
\end{table}

Table~\ref{tab:helm_pikan_case_a_l2_comp} presents a comparison of the three PECANN algorithms in this study with results reported in the literature \cite{mcclenny2020self, SHUKLA2024117290}, as well as our additional tests of PINNs with and without the near-optimal Lagrange multiplier $\lambda_B = 1000$ identified via the PECANN framework.
SA-PINN exhibits highly consistent performance in the cited work; however, it relies on a considerably larger network configuration ($H=4, W=50$) and requires substantially more training epochs and collocation points.
When using the PECANN framework, both the MPU and CPU algorithms show effectiveness with comparable accuracy, due to the relative simplicity of the PDE problem, 
The most effective PECANN-CAPU configuration ($\eta_B = 1$, default internal hyperparameters) was subjected to extended training up to 5000 epochs. The resulting predictions achieved an error level comparable to the mean error reported for SA-PINN~\cite{mcclenny2020self}. Across the five trials conducted, PECANN-CAPU yielded a minimum relative $l^2$ error of $6.96\times10^{-4}$.

In the re-evaluation, PINN achieves its best and most consistent performance when provided with a near-optimal Lagrange multiplier $\lambda_B = 1000$, retrieved from the PECANN-CAPU framework. Use of this optimal value yields a significant improvement over the baseline PINN, highlighting the advantage of a constrained-optimization perspective in determining the optimal weight.

While MPU exhibits superior computational efficiency in this specific, simple case, it fails to yield acceptable results for more complex problems. We assert that to develop a robust, generalizable method applicable across a range of challenges, it is pragmatic to accept a minor efficiency trade-off in simpler cases to guarantee superior performance in complex scenarios.
The computational overhead associated with training for 5000 epochs remains modest, as this is offset by the use of a compact network and a reduced number of collocation points relative to the cited works in Table~\ref{tab:helm_pikan_case_a_l2_comp}.


\section{Theoretical underpinnings of the PECANN-CAPU algorithm}\label{sec:appendixc}
This section begins with a derivation of the augmented Lagrangian method (ALM) \citep{hestenes1969multiplier, powell1969method}, providing a theoretical foundation for the proposed PECANN-CAPU algorithm. First, we examine the benefits of introducing quadratic penalty terms in classical Lagrange multipliers method. We then leverage this analysis to discuss the CAPU algorithm's resulting boundedness and convergence properties.

Starting from the method of Lagrange multipliers, the constrained optimization problem in Eq.~\eqref{eq:constrained_problem} can be expressed as an unconstrained saddle-point problem
\begin{align}
\min_{\theta} \max_{\bar{\bm{\lambda}}} ;
\mathcal{J}(\theta) + \bar{\bm{\lambda}}^{T} \bm{\mathcal{C}}(\theta),
\label{eq:unconstrained_lagrange_multiplier_method}
\end{align}
where $\bar{\bm{\lambda}}$ denotes the classical Lagrange multipliers, distinct from those utilized in the Augmented Lagrangian Method (ALM).

The maximization step in \eqref{eq:unconstrained_lagrange_multiplier_method} with respect to each $\bar{\lambda}_i \in \bar{\bm{\lambda}}$ can be highly non-smooth, since the multiplier becomes unbounded (i.e., approaches $\infty$) unless the constraint $\mathcal{C}_i = 0$ is exactly satisfied. 

To mitigate the issue, a smoothing mechanism is introduced by adding a proximal term that penalizes deviations of $\bar{\bm{\lambda}}$ from their previous estimates $\bm{\lambda}$.
This modification yields the following regularized formulation:
\begin{align}
\min_{\theta} \max_{\bar{\bm{\lambda}}} \mathcal{L}(\theta,\bar{\bm{\lambda}};\bm{\mu}) = \mathcal{J}(\theta) + \bar{\bm{\lambda}}^T \bm{\mathcal{C}}(\theta) - (\frac{1}{2 \bm{\mu}})^T \big[(\bar{\bm{\lambda}} - \bm{\lambda}) \odot (\bar{\bm{\lambda}} - \bm{\lambda})\big],
\label{eq:smoothed_unconstrained_lagrange_multiplier_method}
\end{align}
where $\bm{\mu}$ are positive penalty parameters, controlling the strength of the regularization.
Maximizing the regularized loss in Eq.~\eqref{eq:smoothed_unconstrained_lagrange_multiplier_method} with respect to $\bar{\bm{\lambda}}$ is straightforward.
By setting the gradient to zero, we obtain the first-order optimality condition:
\begin{align}
\bm{\mathcal{C}}(\theta) - \frac{1}{\bm{\mu}}(\bar{\bm{\lambda}} - \bm{\lambda}) = 0,
\label{eq:dual_optimal_cond}
\end{align}
which is equivalent to:
\begin{equation}
    \bar{\bm{\lambda}} = \bm{\lambda} + \bm{\mu} \odot \bm{\mathcal{C}}(\theta).
    \label{eq:dual_optimal_cond2}
\end{equation}
Substituting Eq.~\eqref{eq:dual_optimal_cond2} back into Eq.~\eqref{eq:smoothed_unconstrained_lagrange_multiplier_method} recovers the standard augmented Lagrangian form:
\begin{align}
    \min_{\theta} \max_{\bm{\lambda}} \mathcal{L}(\theta,\bm{\lambda};\mu) = \mathcal{J}(\theta) + \bm{\lambda}^T \bm{\mathcal{C}}(\theta) + \frac{1}{2} \bm{\mu}^T \big[\bm{\mathcal{C}}(\theta) \odot \bm{\mathcal{C}}(\theta)\big].
\end{align}
This confirms that the prior estimates $\bm{\lambda}$ act as the Lagrange multipliers in the ALM, while the smoothing parameters $\bm{\mu}$ govern the regularization strength of the quadratic penalty terms applied to the constraints.

Analyzing the first-order optimality condition given in Eq. \eqref{eq:dual_optimal_cond}, we observe that constraint satisfaction is achieved when the updated multipliers $\bar{\bm{\lambda}}$ remain close to their previous values $\bm{\lambda}$. Moreover, unlike standard penalty methods, the penalty parameters $\bm{\mu}$ do not need to grow indefinitely for the constraints $\bm{\mathcal{C}}(\theta)$ to approach zero. 

From Eq.~\eqref{eq:dual_optimal_cond2}, we can approximate the optimal value $\bar{\bm{\lambda}}$ from the current estimate, which justifies the principled update rule for $\bm{\lambda}$, given by Eq.~\ref{eq:dual_ascent_aug}. 

Overall, this analysis demonstrates the strength of ALM in effectively balancing feasibility and optimality, distinguishing it from pure Lagrange multiplier or pure penalty methods. ALM achieves this balance by combining penalty-based regularization with dual variable updates.


\subsection{Mathematical relationship between point-wise and aggregation formulations}\label{sec:appendixc_comp}
For clarity of exposition, consider a forward problem in which the PDE residual losses are denoted by
$\mathcal{J}_F^{(i)}$, $i \in \{1,2,\dots,N_F\}$, constrained by boundary losses by
$\mathcal{C}_B^{(j)}$, $j \in \{1,2,\dots,N_B\}$.
The point-wise augmented Lagrangian loss is written as
\begin{equation}
    \mathcal{L}_{pw}
    = \sum_{i} \mathcal{J}_F^{(i)}
      + \sum_{j} \lambda_j\,\mathcal{C}_B^{(j)}
      + \sum_{j} \frac{1}{2}\mu_j \mathcal{C}_B^{(j)^{\,2}}.
\end{equation}

To explore the connection with the constraint-aggregation formulation, assume that all Lagrange multipliers take a common value \(\lambda_j = c_\lambda\) and all penalty parameters take a common value
\(\mu_j = c_\mu\).
Introducing the sample means
$
\bar{\mathcal{J}_F} = \frac{1}{N_F}\sum_{i} \mathcal{J}_F^{(i)},  
\quad
\bar{\mathcal{C}_B} = \frac{1}{N_B}\sum_{j} \mathcal{C}_B^{(j)},
$
and using the standard identity relating the mean of squares and the variance,
\begin{equation}
    \frac{1}{N_B}\sum_{j} \mathcal{C}_B^{(j)^{\,2}}
    = \bar{\mathcal{C}_B}^{2} + \operatorname{Var}(\mathcal{C}_B).
\end{equation}
The expression for \(\mathcal{L}_{pw}\) can be rewritten as
\begin{equation}
    \mathcal{L}_{pw}
    = N_F \left[
        \bar{\mathcal{J}_F}
        + \frac{c_\lambda N_B}{N_F}\,\bar{\mathcal{C}_B}
        + \frac{1}{2} \frac{c_\mu N_B}{N_F}\,\bar{\mathcal{C}_B}^{\,2}
        + \frac{1}{2} \frac{c_\mu N_B}{N_F}\,\operatorname{Var}(\mathcal{C}_B)
    \right].
\end{equation}

Considering the constraint-aggregation augmented loss
\begin{equation}
    \mathcal{L}_{avg}
= \bar{\mathcal{J}_F}
+ \lambda_B\,\bar{\mathcal{C}_B}
+ \frac{1}{2}\mu_B\,\bar{\mathcal{C}_B}^{\,2},
\end{equation}
we define the Lagrange multiplier and penalty parameter as
\begin{equation}
\lambda_B = \frac{c_\lambda N_B}{N_F},
\qquad
\mu_B     = \frac{c_\mu N_B}{N_F}.
\label{eq:lambda_mu_avg_from_pntw}
\end{equation}
Then the point-wise augmented loss satisfies
\begin{equation}
    \mathcal{L}_{pw}
    = N_F \left[
        \mathcal{L}_{avg}
        + \frac{1}{2}\mu_B\,\operatorname{Var}(\mathcal{C}_B)
    \right].
\end{equation}
This final expression shows that, under the assumption of constant values for \(\lambda_j\) and \(\mu_j\), the point-wise formulation is precisely \(N_F\) times its constraint-aggregation counterpart, added by an extra penalty term acting on the variance of the boundary constraint residuals.

Apart from the scaling effects induced by the number of sampling points and the relationship in Eq.~\eqref{eq:lambda_mu_avg_from_pntw}, the variance penalty term may introduce undesirable outcomes. As shown in Fig.~\ref{fig:expected_global_constraint}(c,f), the formulation with constraint aggregation achieves accuracy levels that are an order of magnitude superior to (or lower error than) point-wise constraints.

Although this term can help suppress fluctuations in constraint evaluations in certain cases, it may equally degrade convergence toward feasibility. We have encountered this issue specifically in inverse problems where point-wise PDE residuals are incorporated as constraints, as these residuals typically exhibit larger intrinsic variability throughout the optimization process.

\subsection{Boundedness and Convergence of the CAPU Algorithm}
We demonstrate that the CAPU algorithm ensures parameter boundedness and multiplier convergence as constraint violations diminish. This guarantees numerical stability and prevents unbounded parameter growth during the final optimization stages.

\begin{lemma}[Boundedness of CAPU penalty parameters]
\label{lem:capu_mu_boundedness}
Let $\mu_i$ and $\lambda_i$ be the penalty parameter and Lagrange multiplier corresponding to the $i$th constraint $\mathcal{C}_i$ in the PECANN-CAPU algorithm.  
If the moving average of the squared constraint, denoted by $\bar{\nu_i}$, satisfies $\bar{\nu_i} \ll \epsilon$, where $\epsilon$ is a small positive constant introduced for numerical stability in the denominator of the update rule~\eqref{eq:capu_mu}, then $\mu_i$ asymptotically approaches a finite upper bound given by
\begin{equation}
    \mu_i \to \frac{\eta_i}{\sqrt{\epsilon}}.
\end{equation}
Consequently, when the constraint is sufficiently stabilized below $\sqrt{\epsilon}$, the moving average $\bar{\nu_i}$ remains on the order of $\epsilon$, ensuring that $\mu_i$ remains bounded.
\end{lemma}

\begin{proof}
From the CAPU update rule~\eqref{eq:capu_mu}, the denominator involves $\sqrt{\bar{\nu_i} + \epsilon}$.  
When $\bar{\nu_i} \ll \epsilon$, this term becomes approximately $\sqrt{\epsilon}$, leading to
\[
    \mu_i \approx \frac{\eta_i}{\sqrt{\epsilon}}.
\]
Thus, the penalty parameter $\mu_i$ converges to a finite value rather than diverging as required for pure quadratic penalty methods.  
In this regime, the constraint remains stabilized, and $\bar{\nu_i}$ stays near $\epsilon$, preventing unbounded growth of $\mu_i$.  
Hence, $\mu_i$ is bounded as stated, which is also consistently observed in many experiments, for example, shown in Fig.~\ref{fig:vortex_flow_t8_loss_para_evol}.
\end{proof}


\begin{theorem}[Convergence of Lagrange multipliers in CAPU]
\label{thm:capu_lambda_convergence}
Based on Lemma~\ref{lem:capu_mu_boundedness}, suppose that the $i$th constraint $\mathcal{C}_i$ is sufficiently stabilized such that $\mathcal{C}_i \ll \sqrt{\epsilon}$.  
Then, under the CAPU update rule, the corresponding Lagrange multiplier $\lambda_i$ converges to a finite value as the optimization proceeds.

\begin{proof}
When the optimization process reaches a stabilized regime, the conditional check in Algorithm~\ref{alg:capu_adaptive_training_algorithm} is satisfied at each epoch, and the updates of $\lambda_i$ proceed with small increments.  
According to the update rule for the $i$th Lagrange multiplier,
\[
    \lambda_i^{(e)} - \lambda_i^{(e-1)} = \mu_i \mathcal{C}_i,
\]
and by Lemma~\ref{lem:capu_mu_boundedness}, $\mu_i$ remains bounded by $\mu_i < \dfrac{\eta_i}{\sqrt{\epsilon}}$.  
Given that the constraint magnitude satisfies $\mathcal{C}_i \ll \sqrt{\epsilon}$, we have
\begin{equation}
    |\lambda_i^{(e)} - \lambda_i^{(e-1)}| 
    < \frac{\eta_i}{\sqrt{\epsilon}} \, \mathcal{C}_i 
    < \eta_i.
\end{equation}
Therefore, the Lagrange multiplier update becomes progressively smaller as the constraint stabilizes, implying that $\lambda_i$ converges asymptotically to a finite limit.  
\end{proof}
\end{theorem}

\bibliographystyle{model1-num-names}
\bibliography{citations}
\end{document}